\PassOptionsToPackage{table}{xcolor}

\documentclass[final,3p,times,twocolumn,authoryear]{elsarticle}
\usepackage{amssymb}
\usepackage{amsmath}
\usepackage{lineno}
\usepackage{booktabs}
\usepackage{tikz}
\usepackage{pgfplots}\pgfplotsset{compat=1.9}
\usepackage{multirow}
\usepackage[table]{xcolor}
\usepackage{caption}
\usepackage{array}
\usepackage{enumitem}
\usepackage{pifont}% http://ctan.org/pkg/pifont
\newcommand{\cmark}{\ding{51}}%
\newcommand{\xmark}{\ding{55}}% 

\usepackage{hyperref}
\usepackage{subcaption}

\usetikzlibrary{shapes,arrows,positioning}

\journal{Remote Sensing of Environment}

\begin{document}

\begin{frontmatter}

\title{High-resolution Population Maps Derived from Sentinel-1 and Sentinel-2}

\renewcommand{\thefootnote}{\fnsymbol{footnote}} 

\author[label1,label3]{Nando Metzger} 
% \footnotetext[1]{Corresponding Author}

\affiliation[label1]{organization={Photogrammetry and Remote Sensing, ETH Zurich},
            %addressline={},
            city={Zurich},
            %postcode={},
            %state={},
            country={Switzerland}}

\author[label1]{Rodrigo Caye Daudt}

\author[label2]{Devis Tuia}

\affiliation[label2]{organization={Environmental Computational Science and Earth Observation Laboratory, EPFL},%Department and Organization
            %addressline={}, 
            city={Sion},
            %postcode={}, 
            %state={},
            country={Switzerland}}

\affiliation[label3]{organization={Corresponding Author}}

\author[label1]{Konrad Schindler}

\begin{abstract} 
Detailed population maps play an important role in diverse fields ranging from humanitarian action to urban planning. 
Generating such maps in a timely and scalable manner presents a challenge, especially in data-scarce regions.
To address it we have developed \textsc{Popcorn}, a population mapping method whose only inputs are free, globally available satellite images from Sentinel-1 and Sentinel-2; and a small number of aggregate population counts over coarse census districts for calibration.
Despite the minimal data requirements our approach surpasses the mapping accuracy of existing schemes, including several that rely on building footprints derived from high-resolution imagery.
E.g., we were able to produce population maps for Rwanda with 100$\,$m GSD based on less than 400 regional census counts. 
In Kigali, those maps reach an $R^2$ score of 66$\%$ w.r.t.\ a ground truth reference map, with an average error of only $\pm$10 inhabitants/ha.
Conveniently, \textsc{Popcorn} retrieves explicit maps of built-up areas and local building occupancy rates, making the mapping process interpretable and offering additional insights, for instance about the distribution of built-up, but unpopulated areas, e.g., industrial warehouses. 
With our work we aim to democratize access to up-to-date and high-resolution population maps, recognizing that some regions faced with particularly strong population dynamics may lack the resources for costly micro-census campaigns. Project page: \url{https://popcorn-population.github.io/.}
\end{abstract}


\end{frontmatter}

\section{Introduction} \label{sec:intro}

Large-scale high-resolution population maps are an invaluable resource for multiple domains, ranging from humanitarian aid to public health, risk modeling in the reinsurance industry, and urban planning. Fine-grained information about the distribution of the resident population is needed, for instance, to organize the distribution of food and medical supplies, to optimize disaster response, to monitor migration patterns, and as a basis for planning urban development.
Arguably, there is a particularly pressing need for such maps in developing countries that are often confronted with rapid urbanization and internal migration, and associated challenges regarding infrastructure, social development, and pressure on the environment.

The best-performing methods for retrieving fine-grained, spatially explicit population estimates often depend on extensive stacks of off-the-shelf geodata sets~\citep{stevens2015disaggregating,tu2022ensemble,metzger2022fine}.
However, the production of these base data often involves high-resolution (VHR) satellite imagery, whose acquisition, purchase, and/or processing are costly.
Moreover, they quickly become outdated in regions with strong population dynamics and require frequent updating.

Especially \textbf{high-resolution building polygons}, available from organizations like Google~\citep{sirko2021continental,sirko2023high}, Ecopia~\citep{dooley2020gridded}, Microsoft~\citep{microsoft2022worldwide}, and DLR~\citep{esch2022world} offer undeniable benefits for population mapping, but their sustained and comprehensive maintenance presents a significant challenge.
While the datasets themselves are open-source, the models, workflows, and raw imagery behind them are typically proprietary. It remains unclear how frequently they will release updates, which regions will be covered or updated, and what their future cost may be. There is a need for more flexible, scalable, and timely population mapping approaches that do not depend on such high-quality, static, proprietary sources.

% Bottom-Up Population Mapping
\textbf{Bottom-up approaches} use sparse micro-census counts and extrapolate from those counts with the help of densely available auxiliary data to achieve the required spatial coverage.
\cite{hillson2014methods} evaluate the uncertainty of population estimates from satellite imagery and limited survey data, in a case study for Bo City, Sierra Leone.
\cite{weber2018census} combine micro-census counts with high-resolution satellite imagery to create gridded population estimates at 90$\,$m GSD in northern Nigeria, as well as associated uncertainty metrics via Monte Carlo simulation.
\cite{leasure2020national} use a hierarchical Bayesian framework to account for uncertainty in national population maps, with a focus on sparse data.
Similarly, \cite{boo2022high} apply micro-surveys and Bayesian hierarchical models to yield detailed population estimates in the Democratic Republic of Congo.
A limitation of bottom-up approaches is the need for a comprehensive sample of micro-census counts, whose collection involves a significant effort. Consequently, such counts are typically gathered only for specific needs or campaigns, and representative, up-to-date samples are not available for many countries.

% Top-Down Population Mapping
\textbf{Top-down approaches} redistribute coarse census counts to mapping units of about 100$\times$100 to 1000$\times$1000$\,$m\textsuperscript{2}.
This disaggregation process, often termed \emph{dasymetric mapping}, can be understood as estimating \emph{relative} population counts: its goal is to determine what fraction of the overall population is located within each pixel (respectively, mapping unit) in a region. Those fractional weights are derived from various context variables that, contrary to the population counts, are available at the target scale.
Current top-down models mainly use readily available geodata products, including building polygons as mentioned above, OpenStreetMap~\citep{OpenStreetMap}, human settlement layers from land cover maps~\citep{pesaresi2022ghs}, settlement growth models~\citep{nieves2020annually}, and night light composites~\citep{daac2018modis,noaa_ncie_dmsp}.
These covariates are mapped to disaggregation weights either with tabular machine learning algorithms (e.g., random forests, XGBoost), trained in a fully supervised fashion to reproduce more fine-grained census units~\citep{stevens2015disaggregating,tu2022ensemble,sapena2022empiric}; or with neural networks trained in a weakly supervised fashion.
Specifically, our previous work~\citep{metzger2022fine} employs a guided super-resolution technique to directly predict per-pixel counts that add up to the available census units. One key takeaway from that work is that it is more effective to predict local occupancy rates and then multiply them by the number of buildings, rather than directly predict population counts.

When attempting to retrieve \textbf{population maps from raw satellite images}, rather than from geodata layers with a more immediate relation to the number of residents, the challenge becomes to extract features that are predictive of population counts (or densities).
The task is further complicated by the fact that learning the feature extraction in a data-driven manner requires training data, but ground truth population counts are only available either at few, scattered locations or in aggregated form over large areas.
Several studies have explored ways to circumvent that bottleneck.
\cite{islam2017sentinel} proposed to identify built-up regions with a Gaussian maximum-likelihood classifier (a.k.a.\ quadratic discriminant analysis) and then use the resulting built-up area maps to refine population density estimates.
Along the same lines, \cite{CIESIN_Meta_HRSL_2022} detect built-up areas in VHR satellite images and generate a binary built-up area layer with a 30$\times$30$\,$m\textsuperscript{2} grid, which is subsequently used to concentrate the per-region population in the built-up grid cells.
Similarly, \cite{grippa2019improving} also utilize VHR imagery to generate a series of land cover maps, which then serve as a basis for population disaggregation in Dakar, Senegal.
Also using high-resolution imagery, \cite{jacobs2018weakly} combine Planet images (GSD 3$\,$m) with fine-grained U.S.\ census blocks and train a neural network to regress population density maps, using the aggregate density per block as weak supervision signal.
\cite{hafner2023mapping} aim to estimate population growth rather than density, with a Siamese neural network that takes as input a bi-temporal pair of Sentinel-2 images and regresses the population change.
A different, more interactive (and thus less scalable) approach is described by \cite{fibaek2022deep}. Images from Sentinel-1 and Sentinel-2 are segmented into four different classes of settlement structure with a neural network. The class probabilities are then mapped to population densities with the help of hand-crafted formulas. The neural network training involves an active learning loop, during which the user must iteratively select visually well-estimated training regions.

% How do we fill the gap
In summary, bottom-up methods in population mapping offer high accuracy but lack scalability, making them unsuitable for large-scale applications such as public health, risk modeling, and urban planning. Conversely, top-down models based on geodata are scalable, but limited by sporadic updates of the geospatial data they rely on. Previous attempts to incorporate satellite imagery into top-down models mitigate the dependence on derived geodata products but do not scale well, since they require either VHR imagery or manual interventions. 

We argue that high-resolution population maps can be retrieved solely from open-access satellite imagery, using coarse, cumulative census counts as regression targets for offline supervision.  
To that end, we introduce \textsc{Popcorn} (``POPulation from COaRse census Numbers"), a model engineered to overcome the above-mentioned limitations. \textsc{Popcorn}, and its ensemble variant Bag-of-\textsc{Popcorn}, offer a solution for population mapping that

\begin{enumerate}[leftmargin=\labelwidth]
    \item is \textbf{scalable}: Our top-down approach can be trained with as few as 381 population counts over coarse census regions, making it suitable for country-wide mapping. 
    \item allows for \textbf{flexible updates}: The model is based exclusively on data from Sentinel-1 and Sentinel-2, eliminating the reliance on commercial or one-off data products.  
    This ensures cost-effective and sustainable mapping.
    \item provides \textbf{accurate estimates in data-scarce regions}: Despite using satellite imagery with moderate GSD and only weak supervision, the model reaches state-of-the-art mapping accuracy in data-scarce environments.
\end{enumerate}
We hope that the ability to produce population maps in a timely manner, from publicly available data, and at low cost, will particularly benefit developing countries with limited resources, as well as non-governmental organizations operating in such countries.

We start by describing our data in Section~\ref{sec:data}. Section~\ref{sec:meth} presents the \textsc{Popcorn} model in detail. Section~\ref{sec:ex_set} explains the experimental setup and Section~\ref{sec:res} discusses the results. Sections~\ref{sec:con} and~\ref{sec:out} conclude the paper and give an outlook on future research.

\section{Data} \label{sec:data}

\subsection{Census counts and geographic boundaries}

Our experiments use coarse-grained census data from Rwanda, Switzerland, and Puerto Rico for training and fine-grained data for validation, where for this study we limit the experimental settings to residential population. We summarize the datasets in Table~\ref{tab:traindata},
The setup involves maintaining a consistent geographical area while varying the level of detail (or granularity) of the census data. This approach is a standard practice in the field of population disaggregation and super-resolution studies. It allows us to effectively assess the performance of our models in generating detailed population maps from coarser, less detailed census data.

\begin{table*} [tp]
    \centering 
    \resizebox{0.95\textwidth}{!}{%
        \begin{tabular}{@{}lccccccc@{}}
        \toprule
        & year & \# test regions & avg.\ test area &  \multicolumn{1}{c}{\# train regions} & \multicolumn{1}{c}{avg.\ train area} & \multicolumn{1}{c}{upscaling} & \multicolumn{1}{c}{difficulty} \\
        & & & of regions & \multicolumn{1}{c}{\(N\)} & of regions & \multicolumn{1}{c}{\(S\)} & \multicolumn{1}{c}{\(D=\frac{S}{N}\)} \\
        \midrule
        Switzerland      &  2020 & 4$\,$127$\,$837 & 0.01 km\textsuperscript{2} & 2318  & 18  km\textsuperscript{2}                  & \( 42^2\)   & 0.8 \\
        Rwanda           &  2020 & 72$\,$976 & 0.01 km\textsuperscript{2} & 381   & 69  km\textsuperscript{2}                  & \( 83^2\)   & 18  \\
        Puerto Rico      &  2020 & 41$\,$987 & 0.16 km\textsuperscript{2}           & 945   & 14  km\textsuperscript{2}                  & \( 38^2\)   & 1.5 \\
        Uganda           &  2020 & -      & -                                & 1377  & 185 km\textsuperscript{2}                  & \( 136^2 \) & 14 \\
        \bottomrule
        \end{tabular}
    }
    \caption{Summary of training data and dataset difficulty. Upscaling from census regions to a 1$\,$ha raster is more difficult when the upscaling factor \(S\) is large, and when the available number of census regions \(N\) is small.}
    \label{tab:traindata}
\end{table*}

\subsubsection*{Switzerland}

Our training data originates from Switzerland's 2020 register of residence provided by the~\cite{Statpop2020}, which we aggregate to the municipality level. 
While some previous studies~\citep{ciesin_gpqv4_2018,worldpop,schiavina2019ghs,CIESIN_Meta_HRSL_2022} have used an older version of the municipal boundaries with 2'537 municipalities to maintain historical consistency, we opt for the most up-to-date, post-merger boundaries from~\citep{Gemeinden2020} with 2'318 municipalities.
We note that the old boundaries are rasterized to a 100-meter grid, which makes them incompatible with the 10-meter grid of Sentinel imagery. The new boundaries come in vector data, making it possible to work at the images' native resolution.

As reference data for evaluation, we again leverage Switzerland's register of residence~\cite{Statpop2020}, which represents actual population counts. However, we keep it at its original grid resolution of 100$\,$meters. This unique dataset contains 4.1 million individual population counts and provides a robust benchmark to validate the satellite-based predictions of our model. For completeness, we mention that the dataset does not show population counts of one or two people: due to privacy concerns those are rounded up to three. For our purposes this tiny bias is irrelevant.

\subsubsection*{Rwanda}

For training, we utilize an extrapolated census for 2020, which is based on Rwanda's 2012 census. The extrapolation was carried out by the \cite{worldpop} project, with the exponential growth model described in~\cite{stevens2015disaggregating}. The resulting product can be accessed through the~\cite{worldpopDataHub}.
Unfortunately, the 415 census-relevant boundaries are not accurate enough for our purposes: They suffer from discretization patterns to a 100-meter grid, and in some regions exhibit up to 300-meter drift compared to our input imagery. The GSD for our Sentinel-1 \& 2 imagery (described below) is 10$\,$m:
therefore, we match the high-resolution boundary polygons of~\cite{ROSEA_Rwanda_shapefiles} with the original polygons. We transfer the census counts only for regions with at least 70\% intersection-over-union score. This results in a new dataset that can be meaningfully combined with Sentinel images, but has only 381 administrative regions.

For evaluation we utilize detailed, publicly accessible census data for the city of Kigali, made available by \cite{hafner2023mapping}.
That dataset, valid for 2020, offers population counts on a 100$\,$m grid, with a total of $\approx$72$\,$000 grid cells to cover Kigali's 1$\,$132$\,$000 residents.
The product was created by disaggregating census data from 161 city districts with the help of built-up area fractions derived from the Google Open Buildings v1 dataset~\citep{sirko2021continental}.
It provides a high-resolution reference to assess the performance of our model when faced with the conditions of a developing country.
Despite being synthesized from census and building data, the Kigali dataset is sufficiently accurate to serve as validation data. We emphasize that it is an independent reference: our model does not have access to the underlying counts or buildings.
We note that the population counts in the Kigali dataset are not integer numbers, and include a few implausible counts below zero. We conjecture that this is probably the consequence of some interpolation or re-gridding procedure during dataset creation. We have opted to leave those values unchanged, to avoid inadvertently introducing further biases, and to ensure our results on remain comparable {with others}.

\subsubsection*{Puerto Rico}

In Puerto Rico, our training data comes from the second-level administrative tracts of the 2020 U.S.\ Census \citep{us_census_puertorico_2020} and contains  945 individual regions.
The finest census level (so-called \emph{blocks})  serves as a reference for evaluation, with $\approx$42$\,$000 blocks that have an average area of about  400$\times$400$\,$m\textsuperscript{2}.
While not as fine-grained as in Switzerland and Kigali, the granularity is still sufficient for many purposes and can serve as a reference to evaluate our image-based maps (after aggregating them to census blocks).

\subsubsection*{Uganda}
For experiments concerning the model's transferability across national borders, we employ a training dataset from Uganda. For consistency with the Rwanda data, we again opt for 2020 as the reference year. The census numbers as well as administrative boundaries can be accessed through the \cite{worldpopDataHub}.  They contain the extrapolated, sub-national counts from 2014, obtained with the exponential growth model described in~\cite{stevens2015disaggregating}.
For Uganda, we do not have any high-resolution reference data for testing. Therefore, we utilize the fine-grained population counts from Kigali in neighboring Rwanda, described above.

\subsection{Input Data}

% place this figure early in the text to avoid later congestion
\begin{figure*} [tp]
  \includegraphics[width=\textwidth]{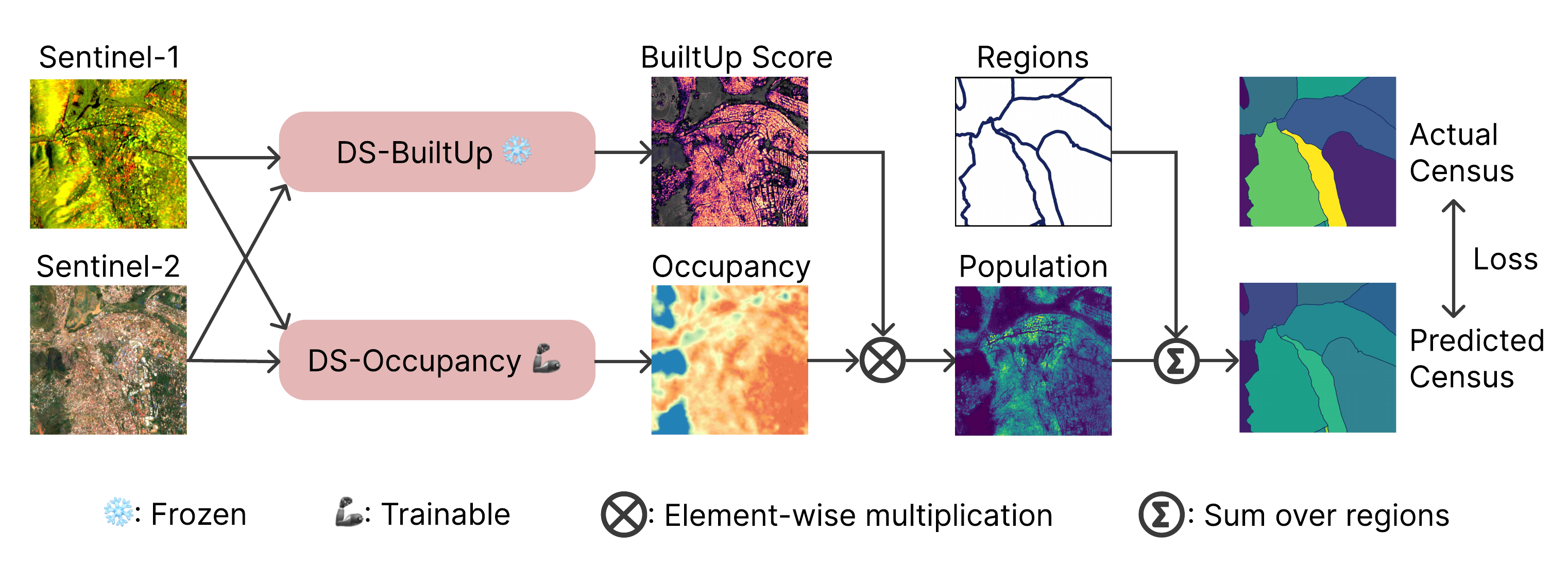}
    \caption{Schematic overview of our approach to population mapping from Sentinel-1 and Sentinel-2  imagery. A pre-trained dual-stream (DS) building detector estimates a per-pixel built-up score. Concurrently, a second, trainable dual-stream block estimates occupancy rates. The population map is derived as the per-pixel product of built-up score and occupancy. To supervise the training of the occupancy branch, the predicted population counts are aggregated within administrative regions and compared to the corresponding census data.}
    \label{fig:graph_abs}
\end{figure*}

\subsubsection*{Sentinel-1}

Sentinel-1 is an active synthetic aperture radar (SAR) system deployed by the European Space Agency (ESA). The mission consists of two satellites, Sentinel-1A and Sentinel-1B, in phase-shifted sun-synchronous orbits%
\footnote{Sentinel-1B failed in December 2021. The launch of its replacement Sentinel-1C is planned for 2024.}.
The constellation provides a revisit time of six days, ensuring regular, frequent coverage.
For our purposes, we source data collected in interferometric wide swath mode from Google Earth Engine. The data comes in the form of ground-range corrected back-scatter (log-)amplitudes for the VV and VH polarisations, resampled to 10$\,$m ground sampling distance.
For each season, we create average composites that reduce the SAR speckle noise. Detailed acquisition dates are given in Appendix \ref{app:season}.

\subsubsection*{Sentinel-2}

Sentinel-2 is an optical satellite mission developed by ESA. It also consists of phase-shifted twin satellites, Sentinel-2A and Sentinel-2B, that operate in sun-synchronous orbits and offer a revisit time of five days for regular, consistent Earth observation and monitoring.
Sentinel-2 captures high-resolution multi-spectral imagery in 13 bands across the visible, near-infrared, and shortwave infrared spectrum. The bands are particularly suited for observing vegetation, land cover, and water bodies.
For our purposes, we source atmospherically corrected surface reflectance images (Level-2A product) for the entire twelve months between March 2020 and February 2021 from Google Earth Engine and produce four seasonal composites to minimize cloud coverage. We use a version of \textit{s2cloudless} that is compatible with Google Earth Engine~\citep{skakun2022cloud, sen2cloudlessEE} to create median composites with both cloud probability threshold and shadow pixel threshold set to 60\% and a mask dilation of 60$\,$m. As commonly done for urban mapping \citep[e.g.,][]{pesaresi2016assessment,vigneshwaran2018extraction}
we only use the four highest resolution (10$\,$m GSD) bands in red (B4), green (B3), blue (B2), and near-infrared (B8).

\section{Methodology} \label{sec:meth}

The core of our method is a neural network model, termed \textsc{Popcorn}. That model has two components: (1) a pre-trained, frozen built-up area extractor; and (2) a building occupancy module that we train through weak supervision with coarse census counts, as illustrated in Figure~\ref{fig:graph_abs} and discussed in the following subsections. 

The model operates at the full Sentinel-1/-2 resolution, i.e., its output has a nominal spatial resolution of 10$\,$m. However, for the final product and evaluation, we aggregate the raw output to a 1$\,$ha (100$\times$100$\,$m) grid. Population counts at 10$\,$m resolution do not have practical advantages, are conceptually questionable, and are impossible to verify, as the inhabitants of a single dwelling unit would in most cases be spread out over multiple pixels.

\subsection{Building Extractor} \label{sec:builtup}

To detect built-up areas we adapt the method developed by \cite{hafner2022unsupervised}. While in principle other building area classifiers could also be considered, we find this classifier a particularly good choice, since its effectiveness and accuracy have been validated across a wide range of geographical contexts globally. Moreover, its training and adaptation strategy aligns well with our project's requirements, notably its independence from the availability of high-resolution geodata sets in the deployment country.

Their work introduces a classification-based neural network architecture that separately extracts features from Sentinel-1 and Sentinel-2 with two parallel U-Net-like streams (called dual-stream architecture), which are then concatenated and mapped to a built-up score, cf.\ Figure~\ref{fig:DDA}.
The model is initially trained with high-resolution building labels from urban areas in North America and Australia, and made globally applicable in a self-supervised fashion, exploiting the fact that predictions based on Sentinel-1 or Sentinel-2 alone should be consistent also in the absence of reference labels.

We make the following modifications to the original architecture: We reduce the channel depth in the U-Net feature extractors to 8 and 16 channels (as opposed to the original  64 and 128 channels), see Figure~\ref{fig:DDA}. This turns out to hardly affect performance (see Appendix \ref{app:builtup} for a quantitative comparison), but shrinks the total parameter count from 1.8 million to a mere 30$\,$000. This brings a drastic reduction in memory consumption which enables training with much larger image patches. Moreover, it makes it possible to reuse the same architecture for the occupancy branch (see below), which needs to be fine-tuned with a much more limited dataset.

As the final layer of our building extraction branch, we employ a 1$\times$1 convolution followed by the sigmoid (a.k.a.\ logistic) activation function. This means that the branch outputs a fractional \emph{BuiltUp score}, i.e., a (pseudo-)probability that a given pixel lies in the built-up area, which can be interpreted as a proxy for the local building density.

The training of our modified version of the building extractor closely follows \cite{hafner2022unsupervised}. We use the same training sites and target labels and also maintain the same split into training, validation, and testing sites. In other words, our model differs from the original one exclusively in terms of architectural modifications.

\subsection{Occupancy and Population Estimation}

In earlier work \citep{metzger2022fine} we showed that, in a scenario where per-pixel building counts have been observed, it is advantageous to leave those counts unchanged and to only estimate a map of the per-building occupancy rate $\text{OccRate}_i$, where $i$ denotes the pixel index.
In much the same way, also for \textsc{Popcorn}, we factor the population number \( {\hat{p}_i} \) into the built-up score and an occupancy rate,
\begin{equation}
{\hat{p}_i} = \text{BuiltUp}_i \times \text{OccRate}_i\;.
\end{equation}
Even though both variables are derived from the same input images, this strategy outperforms a direct estimation of the population counts, as we show in Section~\ref{ssec:abl}. The reason for this is presumably the additional information contributed by the built-up area labels that are used to train the building detector, but for which there are no corresponding population counts to train the direct predictor. 

To estimate building occupancy we instantiate another dual-stream branch with the same U-Net architecture as the building detector. The prediction head for occupancy, after concatenating the Sentinel-1 and Sentinel-2 features, consists of three 1$\times$1 convolution layers, each having a hidden dimensionality of 64, followed by a ReLU activation function. We empirically found that initializing the U-Net with the weights of the pretrained building extractor improves the learning in data-scarce regions, see Section~\ref{ssec:abl}.

\subsection{Weakly Supervised Training}

At the core of our learning approach is a loss function that provides weak supervision from coarse census counts. The function aggregates the fine-grained population estimates from the model into the administrative regions for which the census counts are known. 
Similar training strategies have been used by \cite{jacobs2018weakly,lutio2019guided,metzger2022fine,hafner2023mapping}.
The disagreement between the aggregated population estimates and the ground truth counts serves as the loss to be minimized. Like \cite{metzger2022fine} we use the log-$L1$ distance as a measure of disagreement:
\begin{equation}
    \mathcal{L} = \sum_{j \in \mathcal{N}} \Biggl| \log(1 + c_j) - \log\left(1 + \sum_{k \in A_j} \hat{p}_k \right)  \Biggr| ,
\end{equation}
where \( \mathcal{N} \) denotes the set of administrative regions under consideration.
For each region \(j\) in \(\mathcal{N}\), \(c_j\) is the true census count, and the term \(\sum_{k \in A_j} \hat{p}_k\) sums up the model estimates \(\hat{p}_k\) over all grid cells \(k\) that fall into region \(A_j\). To stabilize the loss, the constant offset of \(1\) bounds the logarithmic counts from below in regions with very few inhabitants (or none at all).
{This loss function allows one to implicitly learn the occupancy rate (\emph{OccRate}) variable from the training data, without explicit information about building functions or types.}

\begin{figure}[t]
    \centering
    \includegraphics[width=0.45\textwidth]{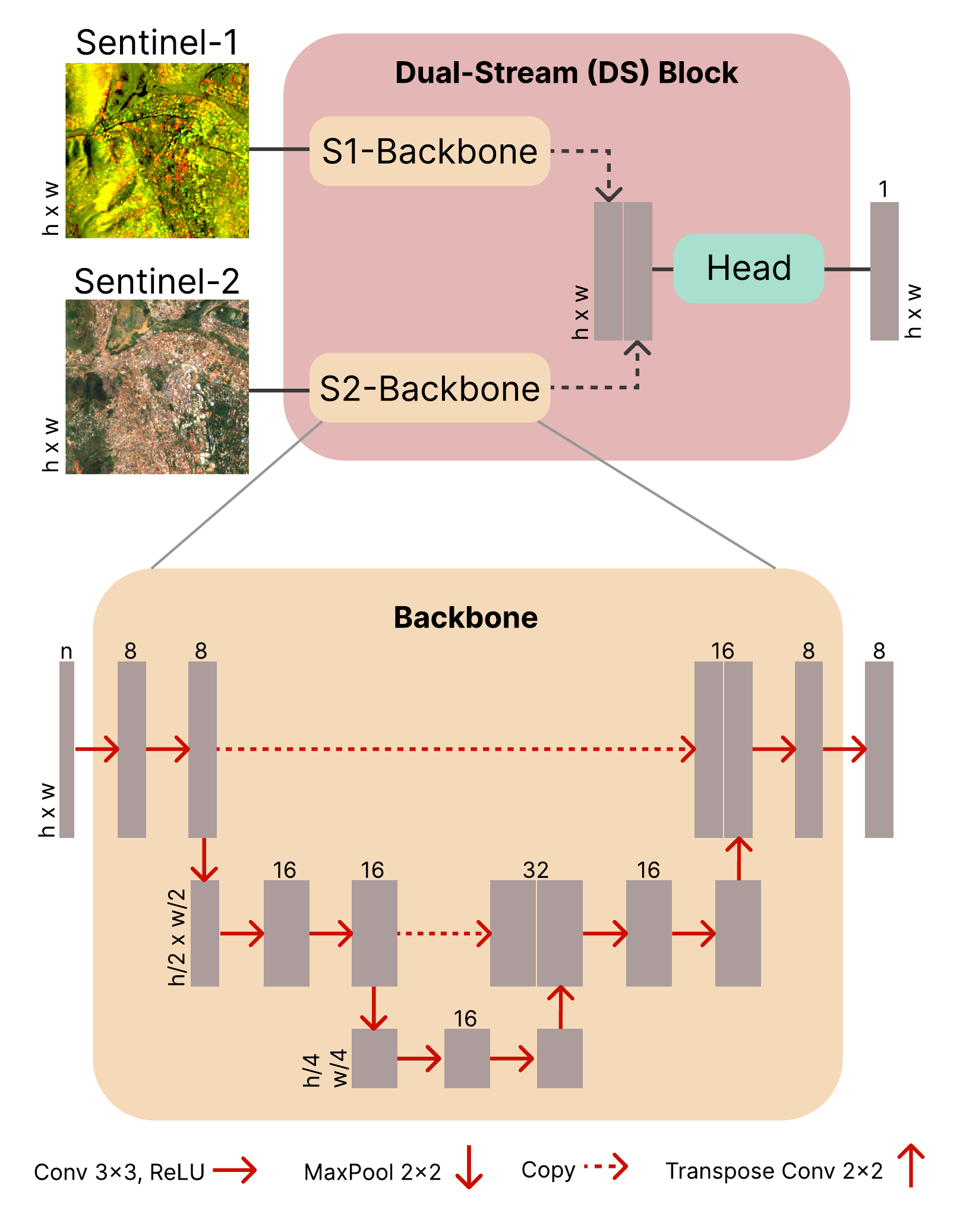}
    \caption{Dual-stream (DS) architecture proposed by~\cite{hafner2022unsupervised}.}
    \label{fig:DDA}
\end{figure}

\subsection{Bagging}

As a standard way of making stochastically learned predictors more reliable, we employ a model ensemble~\citep{dietterich2000ensemble, lakshminarayanan2017simple}, in the following referred to as a "Bag-of-\textsc{Popcorn}".
Five instances of the \textsc{Popcorn} network are trained independently, each time changing the random seed that determines the values of all randomly initialized network weights as well as the composition of the batches during stochastic gradient descent.
Additionally, each of the five models is applied to each of the four seasonal image composites. The 20 resulting predictions are averaged to obtain our final estimate.

\subsection{Dasymetric Mapping}

When estimating high-resolution population maps in the setting where census data for the target region are available, we employ the dasymetric mapping technique~\citep{mccleary1969dasymetric,eicher2001dasymetric}. I.e., in each census region $A_j$ the estimated population numbers \( \hat{p}_i \) of the pixels are understood as \emph{relative} values w.r.t.\ the region aggregate, and re-scaled to adjusted values \( \hat{p}^{adj}_i \), such that their sum matches the actual census count $c_j$ of the region:
\begin{equation}
\hat{p}^{adj}_i = \frac{\hat{p}_i}{\sum_{k \in A_j} \hat{p}_k} \times c_j\;,
\end{equation}
where the first term \( \frac{\hat{p}_i}{\Sigma_{k \in A_j} \hat{p}_k} \) corresponds to the fraction of the region's total population that lives within pixel $i$.
The post-processing step ensures that the estimated numbers add up exactly to the census counts $c_j$. Unless those counts are further from the truth than the satellite-based model estimates, the procedure can thus be expected to improve the population maps, mitigating both over- and under-estimation errors by the model.

We point out that in practice there can be situations where, indeed, the available census counts are not more reliable than the model estimates, e.g., when the census is outdated or relies on projections rather than actual counting. In that case, it may be better to refrain from dasymetric calibration.

\subsection{Implementation Details} \label{ssec:implementation}

Our model is implemented using PyTorch \citep{paszke2019pytorch}. We maintain the same hyper-parameter settings across all datasets and experiments, except for the strength $\lambda_\text{wd}$ of the weight decay regularizer. The latter is derived from the dataset difficulty scores $D$ (c.f.\  Table~\ref{tab:traindata}) via the heuristic relation $5\lambda_\text{wd}=D \times 10^{-6} $.
A regularizer on the model outputs is also included to encourage sparsity via an additional loss term defined as 0.01$\times$ the mean predicted outputs, similarly to \cite{jacobs2018weakly}.

To minimize the loss function we utilize the Adam optimizer \citep{kingma2014adam} with default parameters (i.e., $\beta_1=0.9$ and $\beta_2=0.999$). The batch size is set to 2, such that the model can be trained on a single GPU with 24~GB of onboard memory (in our case an Nvidia GeForce RTX™ 3090 Ti). The base learning rate is progressively decayed by a factor of 0.75 after every fifth epoch.

Considering the small batch size we opt to freeze the pretrained batch normalization layers to avoid discrepancies between the outlier-sensitive batch statistics while training and testing.
The bias term of the final layer in the occupancy branch is initialized with the country-wise disaggregation factor, i.e. the average occupancy rate of the respective dataset. 

The following standard data augmentation techniques are applied to increase the variability of the training data: random adjustments of brightness and linear contrast, random rotations by 0, 90, 180, or 270 degrees, and random mirror reflection (``flipping"). Additionally, we account for seasonal variations by randomly selecting images from all four seasons during training.

\section{Experimental Setup} \label{sec:ex_set}

\subsection{Evaluation Metrics}

To evaluate the performance of our method we compare the predicted population maps to the corresponding, fine-grained reference data in terms of three error metrics: the coefficient of determination ($R^2$), the Mean Absolute Error (MAE), and the Root Mean Squared Error (RMSE).

\paragraph{Coefficient of Determination ($R^2$)} The $R^2$ score measures the proportion of variance explained by our predictions, calculated as::
\begin{equation}
    R^2 = 1 - \frac{\sum_{i=1}^{n}(p_i - \hat{p}_i)^2}{\sum_{i=1}^{n}(p_i - \bar{p})^2}\;,
\end{equation}
with \( p_i \) and \( \hat{p}_i \) the true and predicted population counts for the \( i^{th} \) administrative cell, and \( \bar{p} \) the mean of the true counts.

\paragraph{Mean Absolute Error (MAE)} The average absolute deviation between predicted and actual population counts, in inhabitants per high-resolution unit, quantifies the expected prediction error at a target unit in a robust fashion:
\begin{equation}
    \text{MAE} = \frac{1}{n} \sum_{i=1}^{n} |p_i - \hat{p}_i|
\end{equation}
\paragraph{Root Mean Squared Error (RMSE)} The mean \emph{squared} deviation between predicted and true counts, normalized back to an intuitive scale of inhabitants per unit by taking the square root. Due to the amplification of large deviations, the metric is strongly impacted by rare, extreme prediction errors:
\begin{equation}
\text{RMSE} = \sqrt{\frac{1}{n} \sum_{i=1}^{n} (p_i - \hat{p}_i)^2}
\end{equation}

\subsection{Baselines}

To put the quality of the estimated, fine-grained population maps in context, we benchmark our method against a broad range of competitors. The methods we compare to fall into two categories: self-implemented baselines and established third-party models.

\begin{enumerate}[leftmargin=\labelwidth]
\item \textbf{Self-implemented Baselines:}
  \begin{itemize}[leftmargin=0pt]
    \item[] \textit{Plain Building Disaggregation:} This elementary baseline simply apportionates the population within a census region among its constituting 1$\,$ha pixels according to their building count. For Switzerland, the buildings are extracted from the national mapping agency's Topographic Landscape Model (TLM)~\citep{swisstopo_tlm2023}, for Puerto Rico and Rwanda we use Google Open Buildings~\citep{sirko2021continental}. {As building type labels are not available for the Google Open Buildings, we treat residential and non-residential buildings equally.}
    \item[] \textit{BuiltUp Disaggregation:} This baseline is closer to our work in that it does not require external building counts, but instead redistributes the aggregate numbers per census region to pixels based on the built-up scores derived from {the same Sentinel composites of 2020 as described in the Data Section}. We use the BuiltUp scores obtained with the extractor of Section~\ref{sec:builtup}
    \item[] \textit{Bag-of-\textsc{Popcorn}+count:} This variant combines the proposed \textsc{Popcorn} model with external building counts. The image-based built-up scores are replaced by high-resolution building count datasets based on TLM~\citep{swisstopo_tlm2023} from 2020, respectively Google Open Buildings {version 3}~\citep{sirko2021continental}, whereas the per-pixel occupancy rates are predicted from satellite images as in \textsc{Popcorn}.
  \end{itemize}

\item \textbf{Third-Party Models:}
    For Switzerland and Rwanda, we compare to several other population mapping schemes. A fair comparison for Puerto Rico is unfortunately not possible, because most of those schemes use the highest-resolution census blocks to generate their maps (whereas we intentionally build on a coarser census level, so that the blocks can serve as ground truth for evaluation).
  \begin{itemize}[leftmargin=0pt]
    \item[] \textit{WorldPop~\citep{stevens2015disaggregating}:} WorldPop employs dasymetric mapping with a random forest model based on various off-the-shelf geo-datasets. 
    {For evaluation, we use their 2020 high-resolution product with 100$\,$m resolution, which also has been adjusted to match the United Nations national population estimates. There are two markedly different versions of the product: The first one distributes the population across all pixels in a census region using conventional geodata products (later denoted as "WorldPop"). The second one additionally uses built-up area maps} at the higher target resolution and constrains the disaggregation to the built-up pixels {(later denoted as "WorldPop-Builtup")}.
    \item[] \textit{GPWv4 ~\citep{ciesin_gpqv4_2018}:} The ``Gridded Population of the World Version 4" dataset offers global population estimates derived from the 2010 census round, and extrapolated for the years 2000, 2005, 2010, 2015, and 2020 using UN World Population Prospects. {For our comparison, we use the 2020 version}. The population, according to the extrapolated numbers, is uniformly dispersed across each census region and stored in the form of counts on a 30 arc-second grid ($\approx1$ km at the equator), taking into account the latitude-dependent area of the grid cells.
    \item[] \textit{GHS-Pop~\citep{freire2016development,schiavina2019ghs}:} Fuses sub-national census data with the GHS-Built layer~\citep{pesaresi2009methodology} and a gridded built-up area map derived from Landsat. {The maps are provided at 5-year intervals since 1975, we used the 2020 version. The data can be accessed via the website of the Global Human Settlement Layer~\citep{GHSLwebsite} and comes in 100x100$\,$m gridded estimates}
    \item[] \textit{HRPDM~\citep{CIESIN_Meta_HRSL_2022}:} The ``High-Resolution Population Density Maps" were produced by extracting a binary built-up layer from high-resolution satellite imagery and redistributing the available census counts to only the built-up pixels. {The maps for Rwanda and Switzerland are valid for 2020. All rasters have a resolution of 30$\,$m and are available at the Humanitarian Data Exchange~\citep{MetaHDX}, from which we take the map displaying the ``overall population density".
    }
    \item[] \textit{{LandScan Global and USA~\citep{landscan}:}} The LandScan dataset employs geodata products and high-resolution imagery, which are combined via multi-variable dasymetric mapping to 1$\,$km resolution maps. We evaluate the 2020 ``LandScan Global" product on all our evaluation sets. Moreover, for Puerto Rico we also evaluate the 2020 ``LandScan USA", which additionally uses block-level information to improve the quality of the global maps and comes at 100$\,$m resolution.
    The LandScan HD product is currently not available for any of our study regions, so it was not considered for the present work. 
    \item[] \textit{Kontur~\citep{konturwebsite}:} The Kontur population maps are created using the Global Human Settlement Layer~\citep{pesaresi2022ghs}, Meta's HRPDM population data~\citep{CIESIN_Meta_HRSL_2022} and building polygons from various sources -- among them, the Microsoft Building Footprints~\citep{microsoft2022worldwide}. The dataset is provided on hexagonal grids with resolutions of 400$\,$m, 3$\,$km and 22$\,$km. 
    We utilize the 400$\,$m version, released on the September 28, 2020 via the Humanitarian Data Exchange~\citep{konturHDX}.
    \item[] \textit{\textsc{Pomelo}~\citep{metzger2022fine}:}  This method predicts building occupancy from a collection of geodata layers (nightlights, terrain height, distance to roads and waterways, etc.) and combines it with building counts derived from high-resolution footprints (Google Open Buildings, Maxar Ecopia Maps). Optionally, it offers dasymetric disaggregation of available census counts. The 100$\,$m resolution maps are not explicitly dated but have been produced with covariates collected between 2015 and 2020.
  \end{itemize}
\end{enumerate}
    
\section{Results} \label{sec:res}

In the following, we present the prediction quality of the proposed model and compare it with the alternative population mapping techniques described above.
The evaluation encompasses three very different regions with diverse geographical conditions, chosen also to have sufficiently high-resolution reference data: Switzerland, Kigali (Rwanda), and Puerto Rico.
Quantitative results are shown in Tables \ref{tab:che}, \ref{tab:rwa}, and \ref{tab:pri}, respectively, where bold numbers indicate the best-performing map per category.
    
\subsection{Evaluation of Population Maps}

\begin{table}[tp]
    \centering
    \caption{Quantitative evaluation for Switzerland.}
    \label{tab:che}
    \resizebox{0.48\textwidth}{!}{%
    \begin{tabular}{@{}m{2em}lccc@{}}
        \toprule
        & & $R^2$ & MAE & RMSE \\ 
        \midrule
        \multirow{7}{*}{\rotatebox[origin=c]{90}{\parbox[c]{2cm}{\centering High Resolution}}} 
        & TLM Disaggregation & 0.40 & 1.60 & 10.2 \\
        & WorldPop-Builtup & 0.37 & 2.11 & 10.5  \\
        & HRPDM (Meta) & 0.41 & 1.77 & 10.2 \\
        & Kontur & 0.31 & 2.9 & 11.1 \\
        & LandScan Global & 0.29 & 2.6 & 11.2 \\
        & \textsc{Pomelo} & \textbf{0.53} & \textbf{1.45} & \textbf{9.1}  \\ 
        & Bag-of-\textsc{Popcorn}+count & 0.39 & \textbf{1.45} & 10.4\\
        \midrule
        \multirow{5}{*}{\rotatebox[origin=c]{90}{\parbox[c]{2cm}{\centering Medium Resolution}}} 
        & BuiltUp Disaggregation & 0.53 & 1.68 & 9.1    \\
        & GPWv4 & 0.06  & 4.3  & 12.9  \\
        & GHS-Pop & 0.45 & 1.88 & 9.8 \\
        & WorldPop & 0.38 & 2.4  & 10.4  \\ 
        & Bag-of-\textsc{Popcorn} & \textbf{0.60}  & \textbf{1.35}  & \textbf{8.4} \\ 
        \bottomrule
\end{tabular}%
}
\end{table}

\begin{table}[tp]
    \centering
    \caption{Quantitative evaluation for Kigali, Rwanda.}
    \label{tab:rwa}
    \resizebox{0.48\textwidth}{!}{%
    \begin{tabular}{@{}m{2em}lccc@{}}
        \toprule
        & & $R^2$ & MAE  & RMSE \\ 
        \midrule
        \multirow{7}{*}{\rotatebox[origin=c]{90}{\parbox[c]{2cm}{\centering High\\Resolution}}} 
        & Google Buildings Disag. & 0.43 & 10.5 & 26.2 \\
        & WorldPop-Builtup & 0.58 & 10.2 & 22.6 \\
        & HRPDM (Meta) & 0.42 & 12.2 & 26.5 \\
        & Kontur & 0.35 & 15.9 & 28.0 \\
        & LandScan Global & 0.25 & 15.6 & 30.2 \\
        & \textsc{Pomelo} & \textbf{0.69} & \textbf{9.8} & \textbf{19.5} \\ 
        & Bag-of-\textsc{Popcorn}+count & 0.61 & 10.2 & 21.6 \\
        \midrule
        \multirow{5}{*}{\rotatebox[origin=c]{90}{\parbox[c]{2cm}{\centering Medium\\Resolution}}} 
        & BuiltUp Disaggregation & 0.34 & 12.0 & 28.1 \\
        & GPWv4 & 0.12 & 20.4 & 32.6 \\
        & GHS-Pop & 0.20 & 14.6 & 31.1 \\
        & WorldPop & 0.33 & 16.7 & 28.4 \\ 
        & Bag-of-\textsc{Popcorn}  & \textbf{0.66} & \textbf{10.1} & \textbf{20.2} \\
        \bottomrule
    \end{tabular}%
    }
\end{table}

\begin{table}[tp]
\centering
\caption{Quantitative evaluation for Puerto Rico.}
\label{tab:pri}
\resizebox{0.48\textwidth}{!}{%
\begin{tabular}{@{}m{2em}lccc@{}}
    \toprule
    & & $R^2$ & MAE  & RMSE \\ 
    \midrule
    \multirow{5}{*}{\rotatebox[origin=c]{90}{\parbox[c]{2cm}{\centering High\\Resolution}}} 
    & Google Buildings Disag. & 0.75 & 26.8 & 65 \\ 
    & Kontur & 0.28 & 50.5 & 109\\ 
    & LandScan Global & -0.38 & 59.4 & 151\\
    & LandScan USA & 0.54 & 31.4 & 88\\ 
    & Bag-of-\textsc{Popcorn}+count & \textbf{0.78} & \textbf{24.3} & \textbf{61} \\
    \midrule
    \multirow{3}{*}{\rotatebox[origin=c]{90}{\parbox[c]{1.25cm}{\centering Medium\\Resolut.}}} 
    & BuiltUp Disaggregation & 0.74 & 28.7 & 65  \\
    & GPWv4 & -0.77 & 67.3  & 171  \\ 
    & Bag-of-\textsc{Popcorn} & \textbf{0.82} & \textbf{23.6} & \textbf{55} \\
    \bottomrule

\end{tabular}%
}
\end{table}

\begin{figure*}[tp]
    \centering
    \resizebox{1.0\textwidth}{!}{%
    \begin{tabular}{c@{}c@{\hspace{10pt}}c@{}c@{\hspace{10pt}}c@{}c}
        \textbf{Switzerland} & & \textbf{Rwanda} & & \textbf{Puerto Rico} & \\
        \includegraphics[height=0.30\textwidth, trim={0 0 8pt 0}, clip]{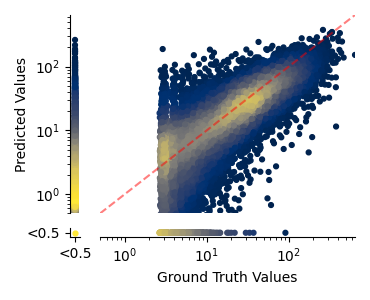} &
        \raisebox{.3\height}{\includegraphics[height=0.2\textwidth]{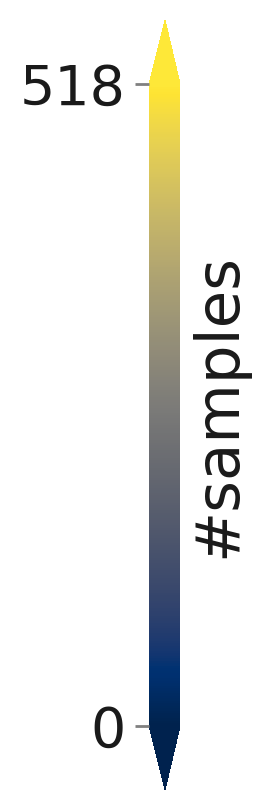}} &
        \includegraphics[height=0.30\textwidth, trim={19pt 0 8pt 0}, clip]{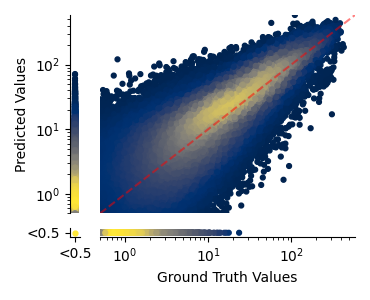} &
        \raisebox{.3\height}{\includegraphics[height=0.2\textwidth]{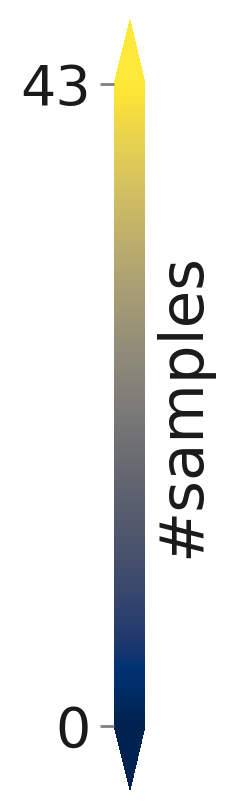}} &
        \includegraphics[height=0.30\textwidth, trim={19pt 0 8pt 0}, clip]{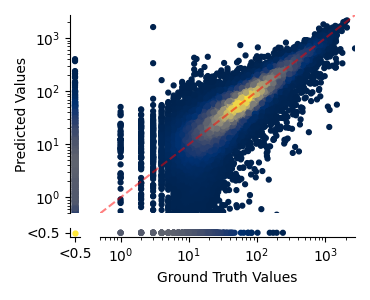} &
        \raisebox{.3\height}{\includegraphics[height=0.2\textwidth]{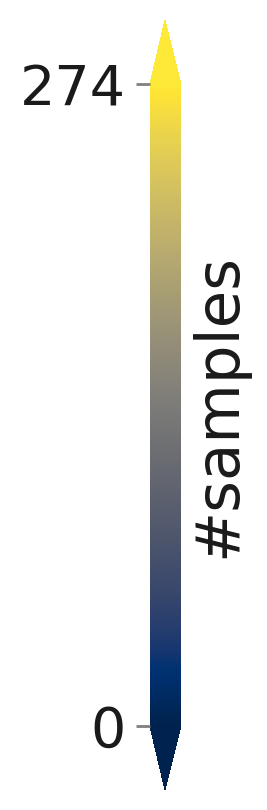}} \\
    \end{tabular}
    }
    \caption{Scatter plots for Switzerland, Rwanda, and Puerto Rico. Note the logarithmic scale of the axes. Values close to zero (below 0.5) have been grouped into a single bin.}
    \label{tab:scatter-graphics}
\end{figure*}

Tables \ref{tab:che}, \ref{tab:rwa}, and \ref{tab:pri} summarize the evaluation metrics for the three tested regions.
We note in particular that the Bag-of-\textsc{Popcorn} model, based exclusively on medium-resolution Sentinel imagery, is competitive even with methods that have access to additional, higher-resolution data products such as building footprints extracted from VHR images or topographic -- in most cases even overtaking them.
On the Swiss dataset it achieves the best results for all three error metrics, i.e., the highest correlation with the ground truth counts as well as the smallest (absolute and squared) deviations from the true counts.
In Rwanda, arguably the most challenging environment for population mapping, Bag-of-\textsc{Popcorn} surpasses all other medium-resolution methods by large margins, boosting the $R^2$ value from 34\% (for the strongest competitor) to an impressive 66\%.
The good mapping quality underlines the robustness of our approach under conditions that are challenging for map down-scaling and for machine learning: to map Kigali at 1$\,$ha resolution one must disaggregate the census counts by a factor \textgreater 6800, while only 381 regions are available to train the method.
Even among methods that make use of high-resolution inputs, only \textsc{Pomelo} narrowly outperforms our proposed method, but to do so requires up-to-date building footprints. We note that one of the data sources to build the Kigali dataset was the Google Open Buildings version 1 -- the same dataset used also by \textsc{Pomelo}. Hence, the performance metrics that we obtain for the latter may be slightly too optimistic.
The mapping performance in Puerto Rico had to be evaluated at the level of census blocks that are on average $\approx$400$\times$400$\,$m\textsuperscript{2} in size, as no ha-level reference data is available.
The corresponding results are consistent with those for Switzerland and Rwanda. Bag-of-\textsc{Popcorn} has a clear edge over simpler disaggregation schemes driven only by building counts. Notably, dropping the retrieval of built-up scores from Sentinel and instead using building counts derived from Google Open Buildings (Bag-of-\textsc{Popcorn}+count) does improve the mapping performance compared to simple disaggregation, i.e., the estimated occupancy values bring an added value. Interestingly, our default Bag-of-\textsc{Popcorn} is still better than this variant, in other words, the counts derived from high-resolution building outlines do not bring an advantage over built-up scores estimated from Sentinel images.
Generally speaking, we find that methods that allow for varying building occupancy work best, while methods that assume a constant occupancy and disaggregate building information are less accurate, but still fairly robust.
Finally, the Kontur, LandScan, and GPWv4 maps with resolutions of 400$\,$m 1$\,$km, and 1$\,$km perform worst, with very low $R^2$ values, and even strongly negative for the Puerto Rico case. {This is not surprising, on the one hand, due to the lower resolution of the maps and on the other hand because the aim of these products is a globally consistent map product rather than a fine-grained one.}
Our results support the hypothesis that down-scaling to ha-level resolutions is indeed meaningful, and that readily available covariates like satellite images or building counts do contain the necessary information to resolve population maps to that resolution. 
We also observe that among the maps derived from low- and medium-resolution inputs, Bag-of-\textsc{Popcorn} exhibits the largest performance gains for the challenging setting of Rwanda. This is further evidence that \textsc{Popcorn}'s modeling design choices are justified. 
% implements favorable inductive biases that make it particularly data-efficient.
On the datasets from Switzerland and Puerto Rico, the BuiltUp Disaggregation method performs rather well ($R^2$ scores only 7 percentage points lower than \textsc{Popcorn}), likely because the built-up area classifier was trained on data from developed countries. In contrast, Rwanda represents a domain shift that degrades the classifier's performance. The \textsc{Popcorn} architecture is able to compensate for that issue and boosts the $R^2$ score from 34\% to 66\%.

In Figure~\ref{tab:scatter-graphics}, we show scatter plots between Bag-of-\textsc{Popcorn} predictions and true population counts for the three different geographic regions. While, naturally, there are some large relative discrepancies (especially for low counts), the values are clustered near the identity line across the entire value range, and the distribution of the prediction errors is symmetric. I.e., most predictions deviate only little from the true values, and the overall fit appears largely unbiased.

\newcommand{\heightcomp}{0.17\textwidth}
\begin{figure*}[htbp]
\centering
\resizebox{0.82\textwidth}{!}{%
\begin{tabular}{cccccccc}
            &
            & \parbox{21mm}{\centering VHR} 
            % & \parbox{21mm}{\centering Sentinel-2}
            & \parbox{21mm}{\centering WorldPop \\ Builtup.}
            & \parbox{21mm}{\centering HRPDM \\ Meta}
            & \parbox{21mm}{\centering Kontur} 
            & \parbox{21mm}{\centering Landscan Global}  
            & \parbox{21mm}{\centering \textsc{Pomelo}} \\
            
% \rotatebox{90}{\textbf{\parbox{35mm}{\centering Mahama Refugee \\ Camp, Rwanda}}}
\multirow{3}{*}[6em]{\rotatebox{90}{\textbf{\parbox{55mm}{  \centering Mahama Refugee Camp, Rwanda}}}}
            & \rotatebox{90}{\textbf{\parbox{35mm}{\centering High Resolution}}}
            & \rotatebox{90}{\includegraphics[height=\heightcomp]{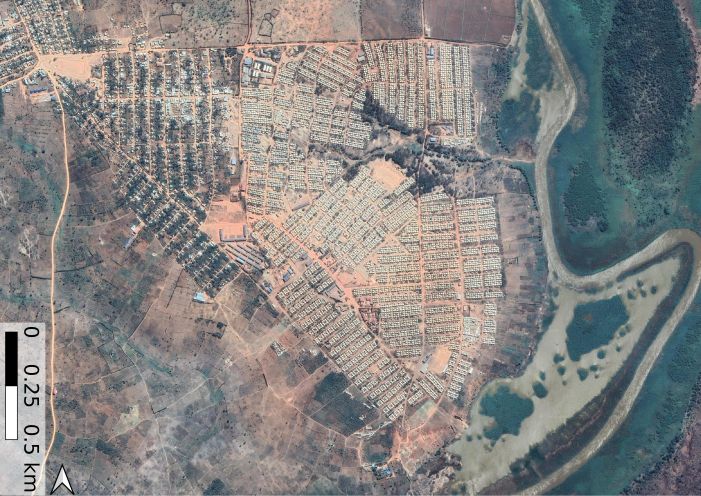}}
            % & \rotatebox{90}{\includegraphics[height=\heightcomp]{rwa1_sentinel2.png}}
            & \rotatebox{90}{\includegraphics[height=\heightcomp]{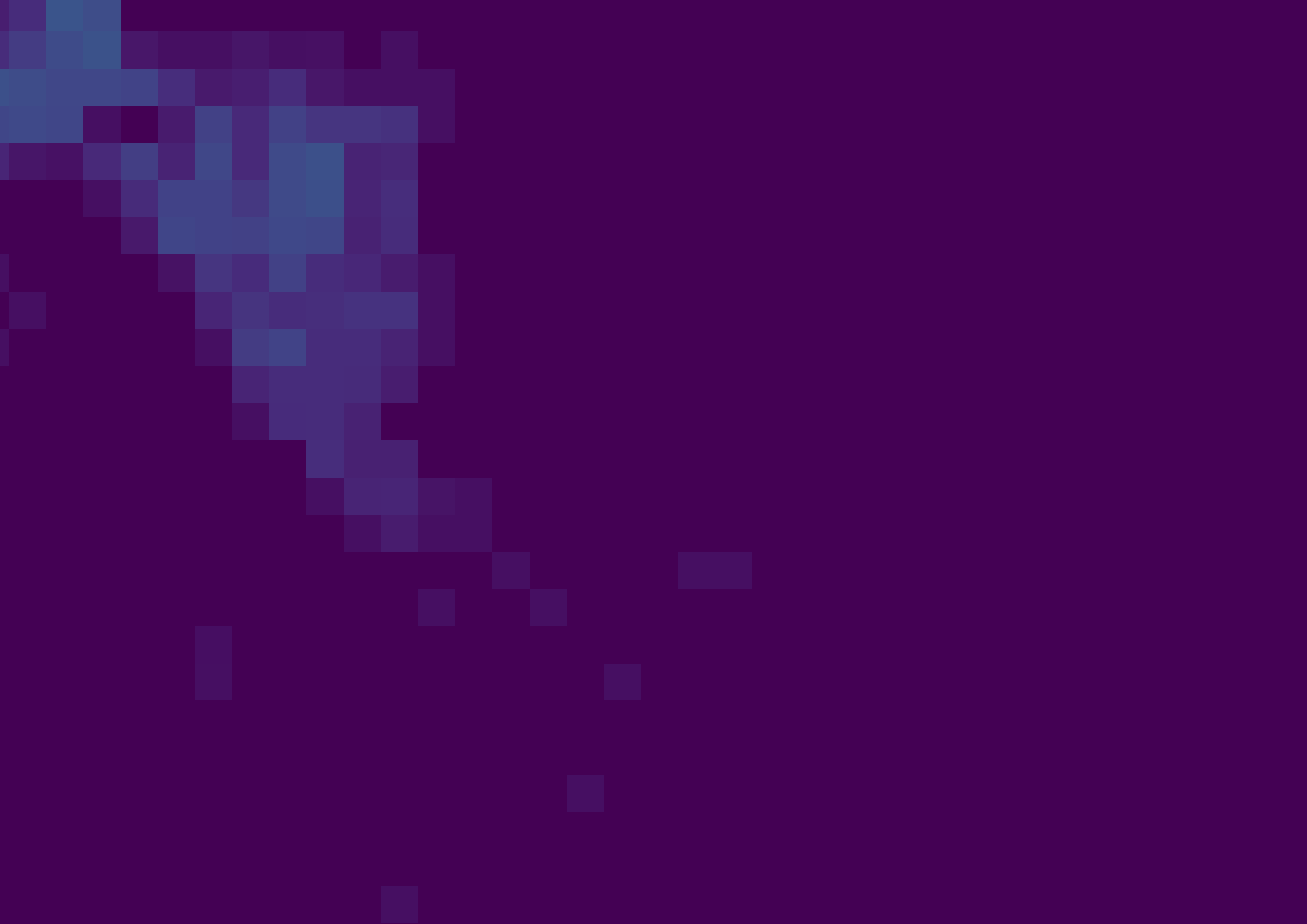}}
            & \rotatebox{90}{\includegraphics[height=\heightcomp]{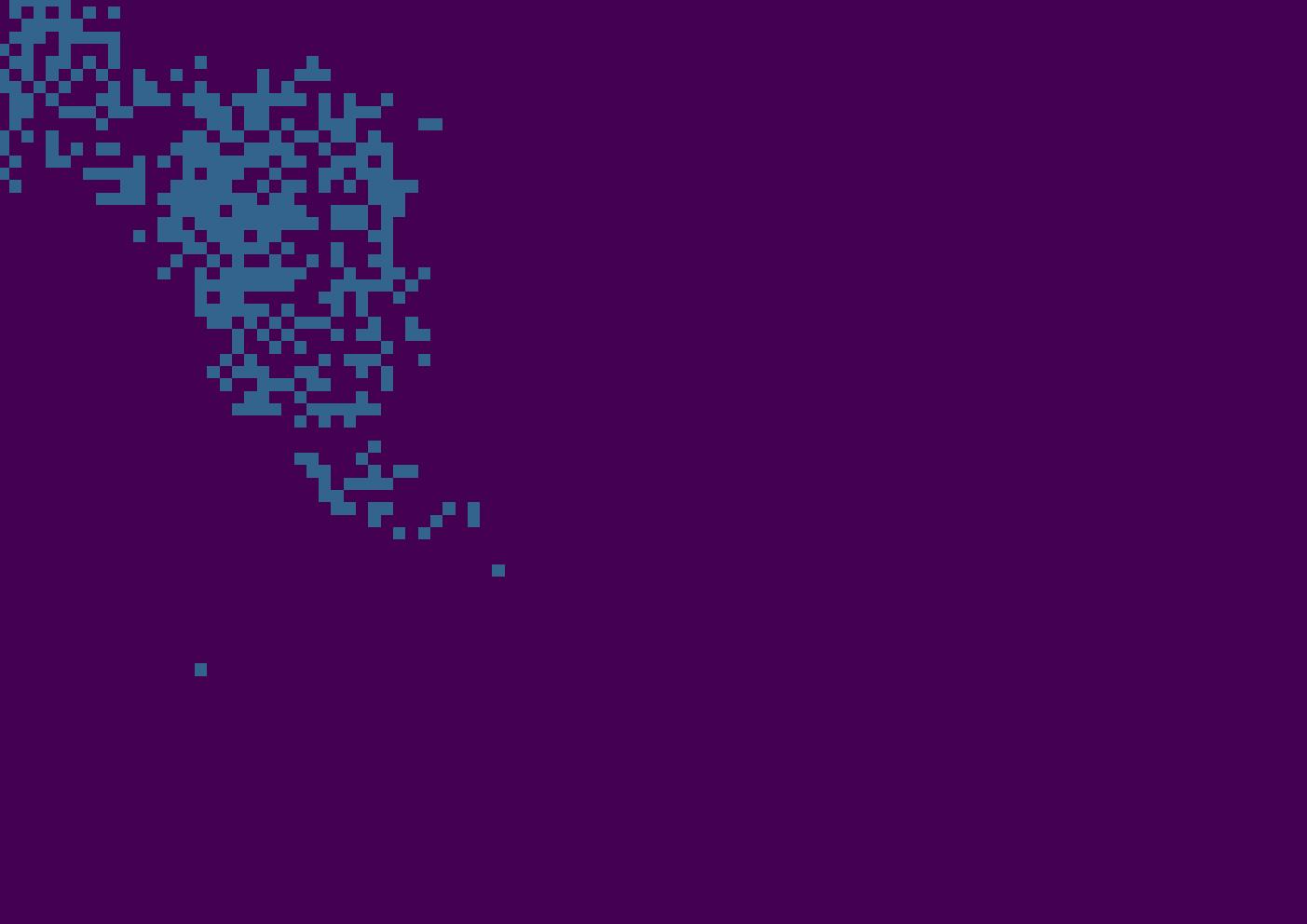}}
            & \rotatebox{90}{\includegraphics[height=\heightcomp]{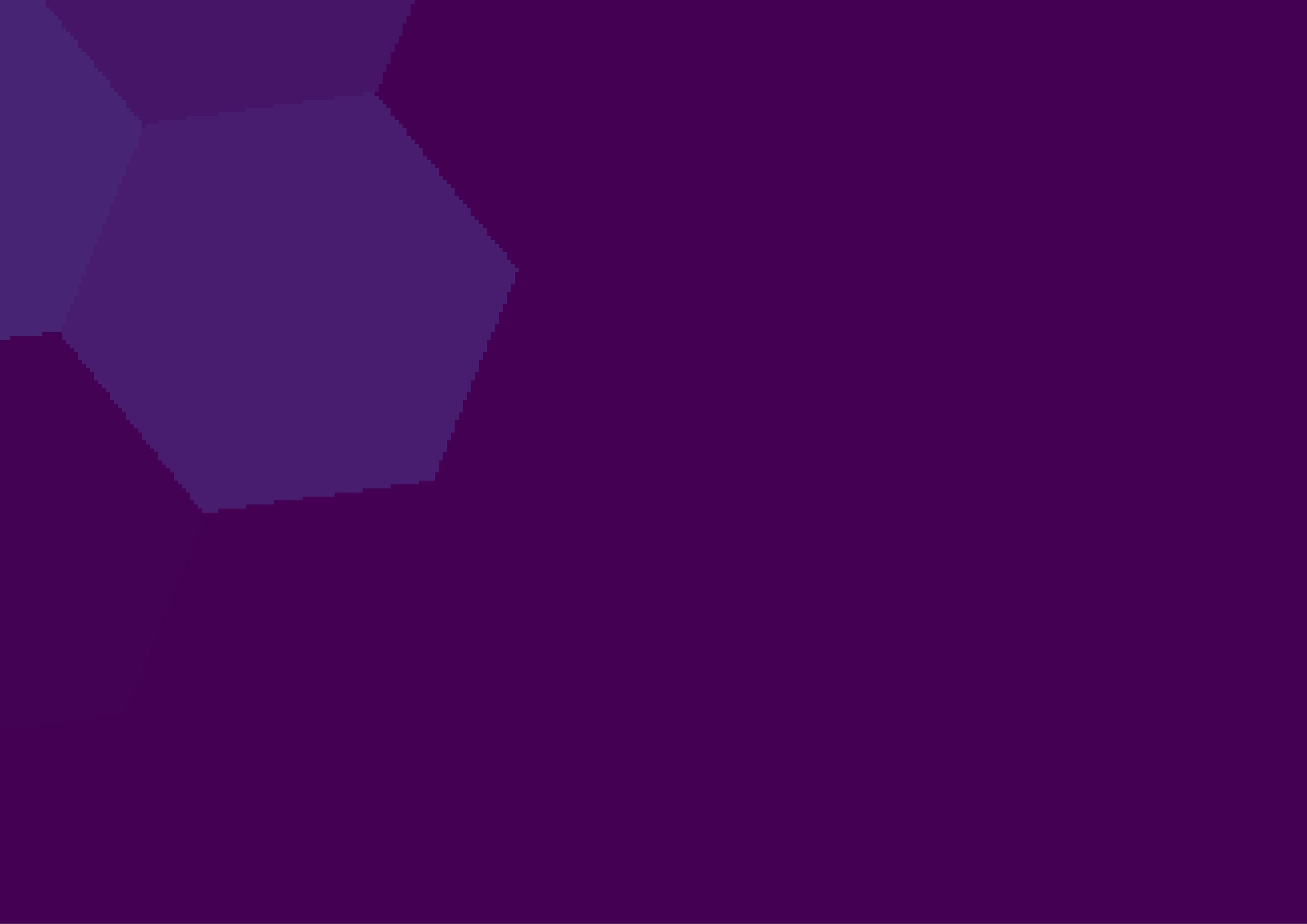}}
            & \rotatebox{90}{\includegraphics[height=\heightcomp]{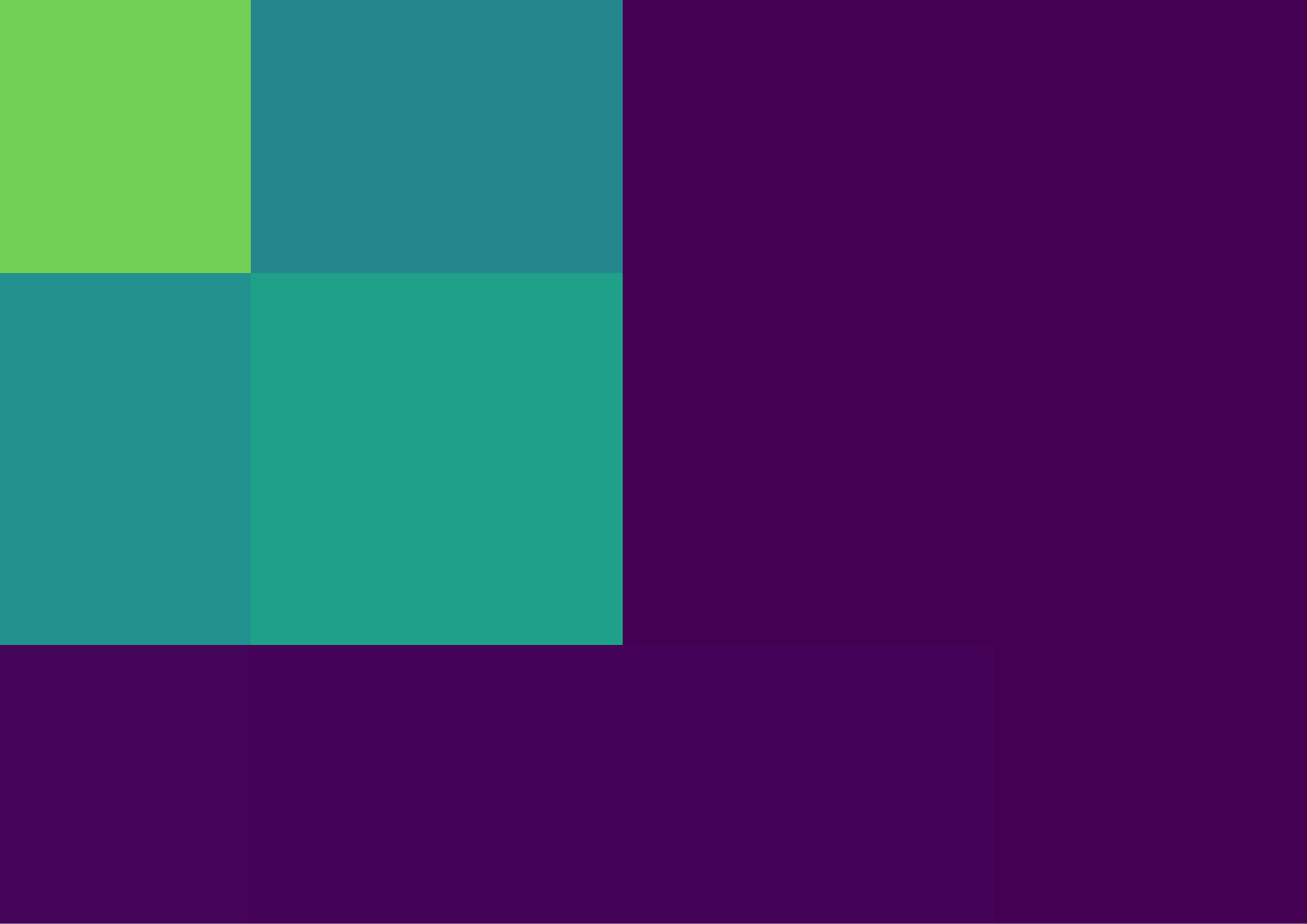}}
            & \rotatebox{90}{\includegraphics[height=\heightcomp]{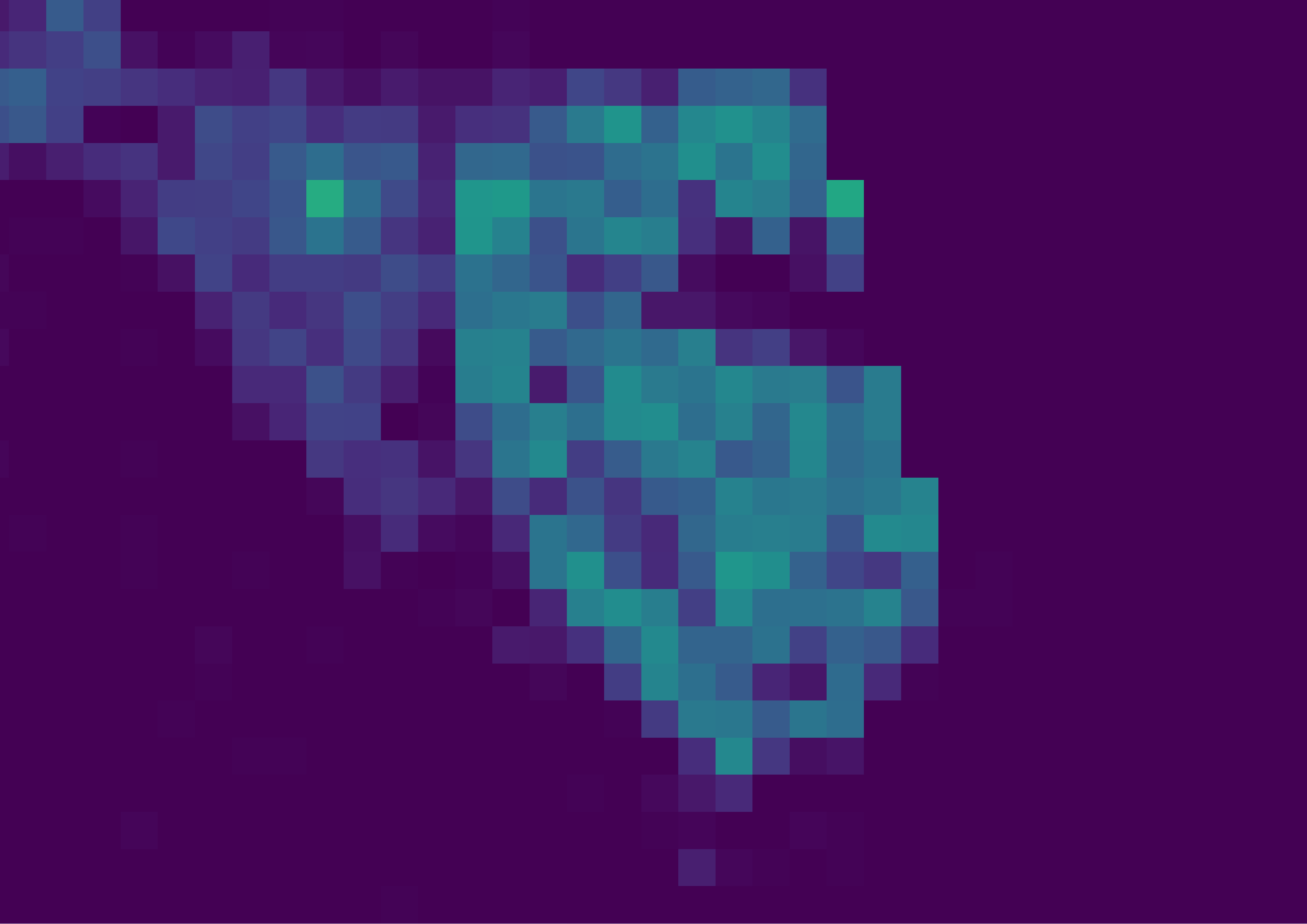}} \\ 

            &
            & \parbox{21mm}{\centering Sentinel-2}
            & \parbox{21mm}{\centering GPWv4} 
            & \parbox{21mm}{\centering GHS-Pop} 
            & \parbox{21mm}{\centering WorldPop}
            & \parbox{21mm}{\centering Bag-of-\textsc{Popcorn}} \\
            
            & \rotatebox{90}{\textbf{\parbox{35mm}{\centering Medium Resolution}}}
            & \rotatebox{90}{\includegraphics[height=\heightcomp]{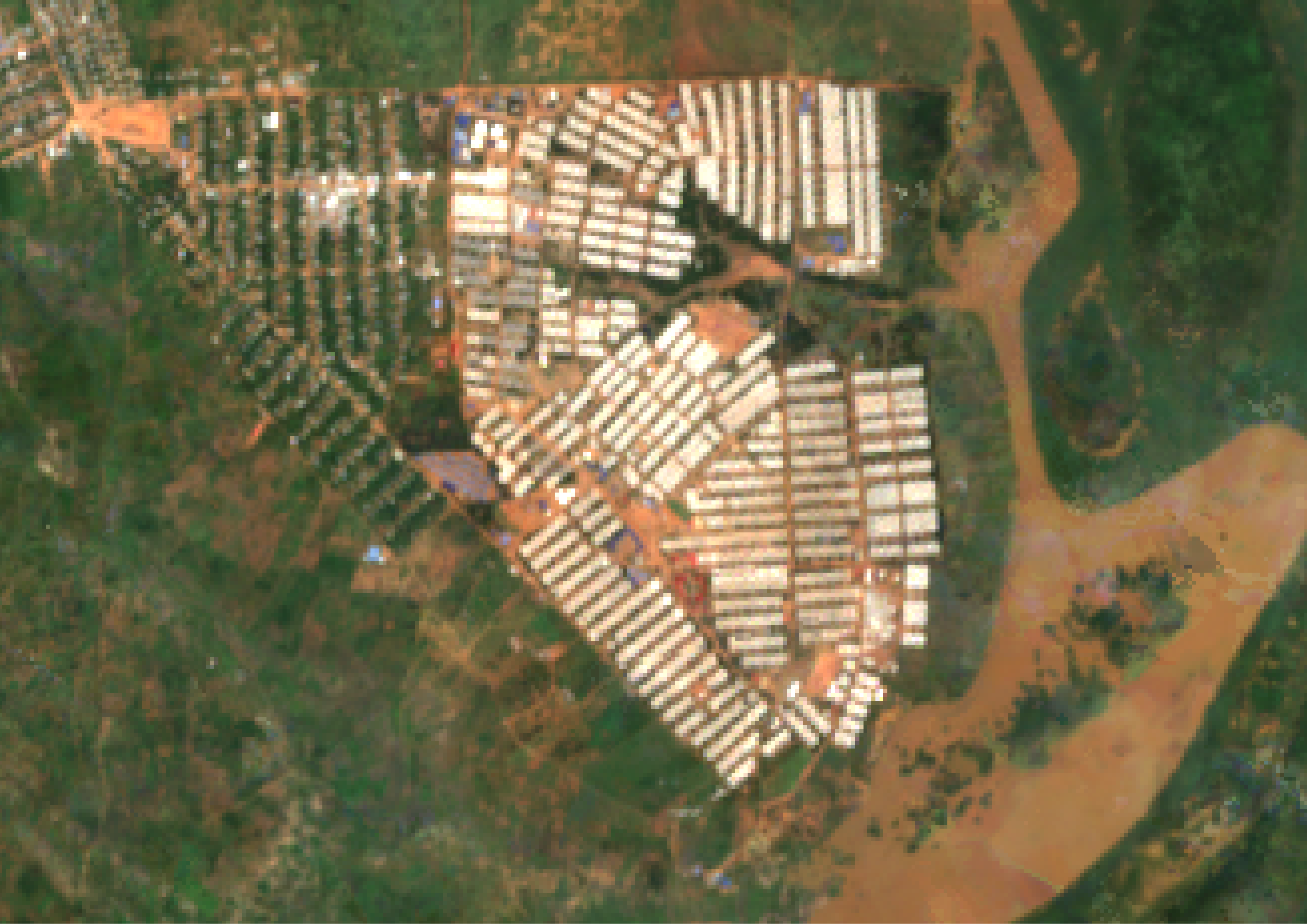}}
            & \rotatebox{90}{\includegraphics[height=\heightcomp]{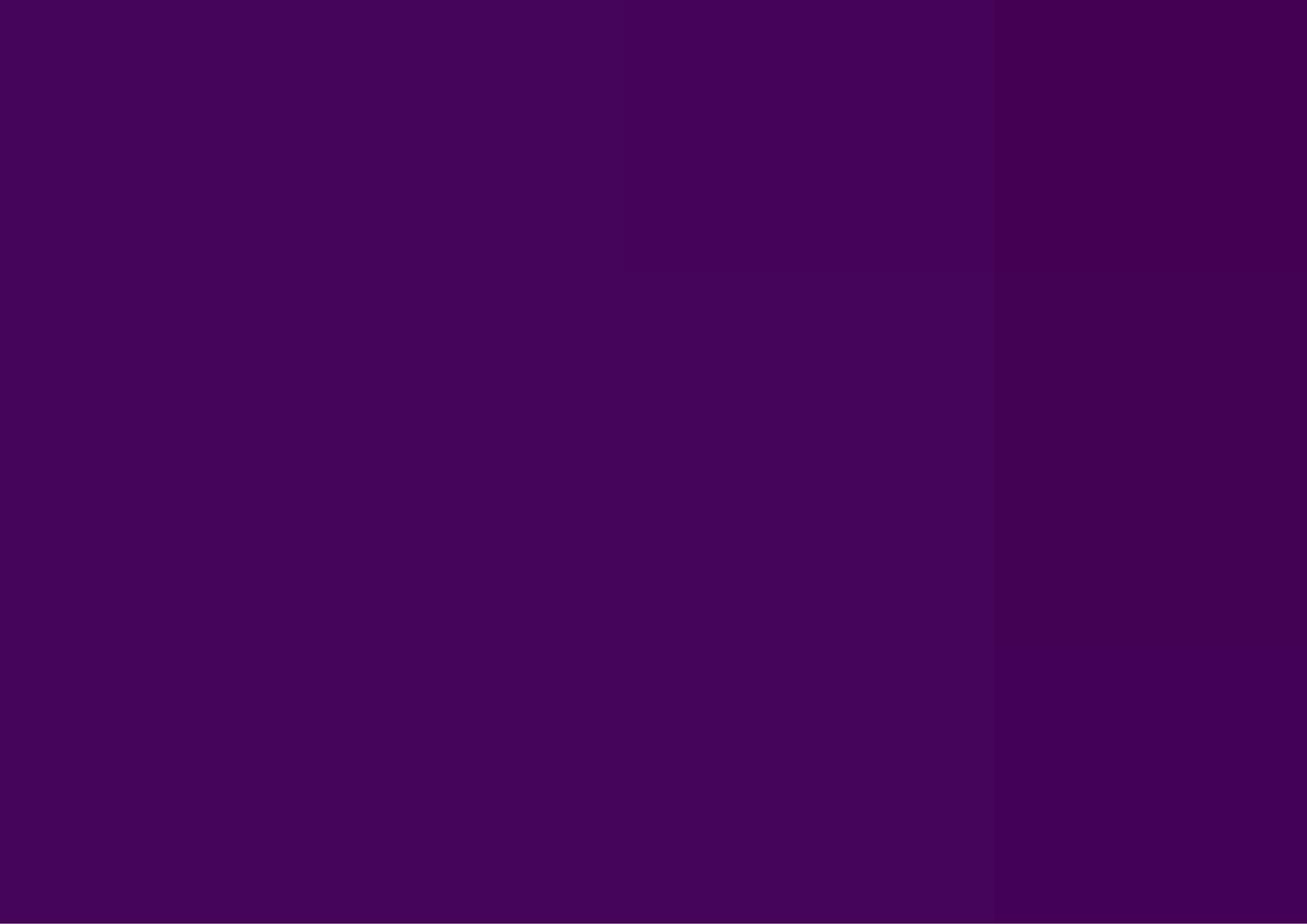}}
            & \rotatebox{90}{\includegraphics[height=\heightcomp]{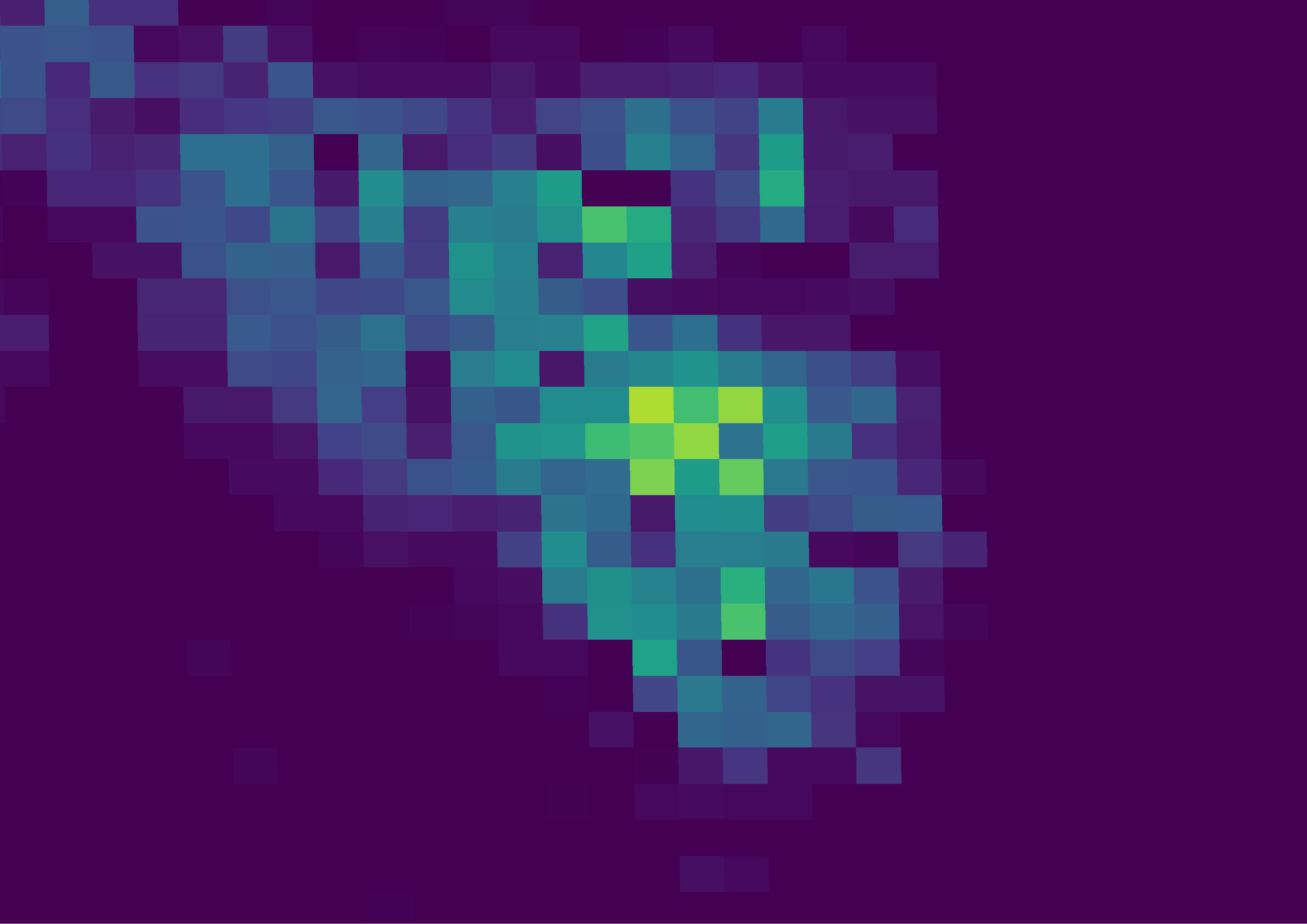}}
            & \rotatebox{90}{\includegraphics[height=\heightcomp]{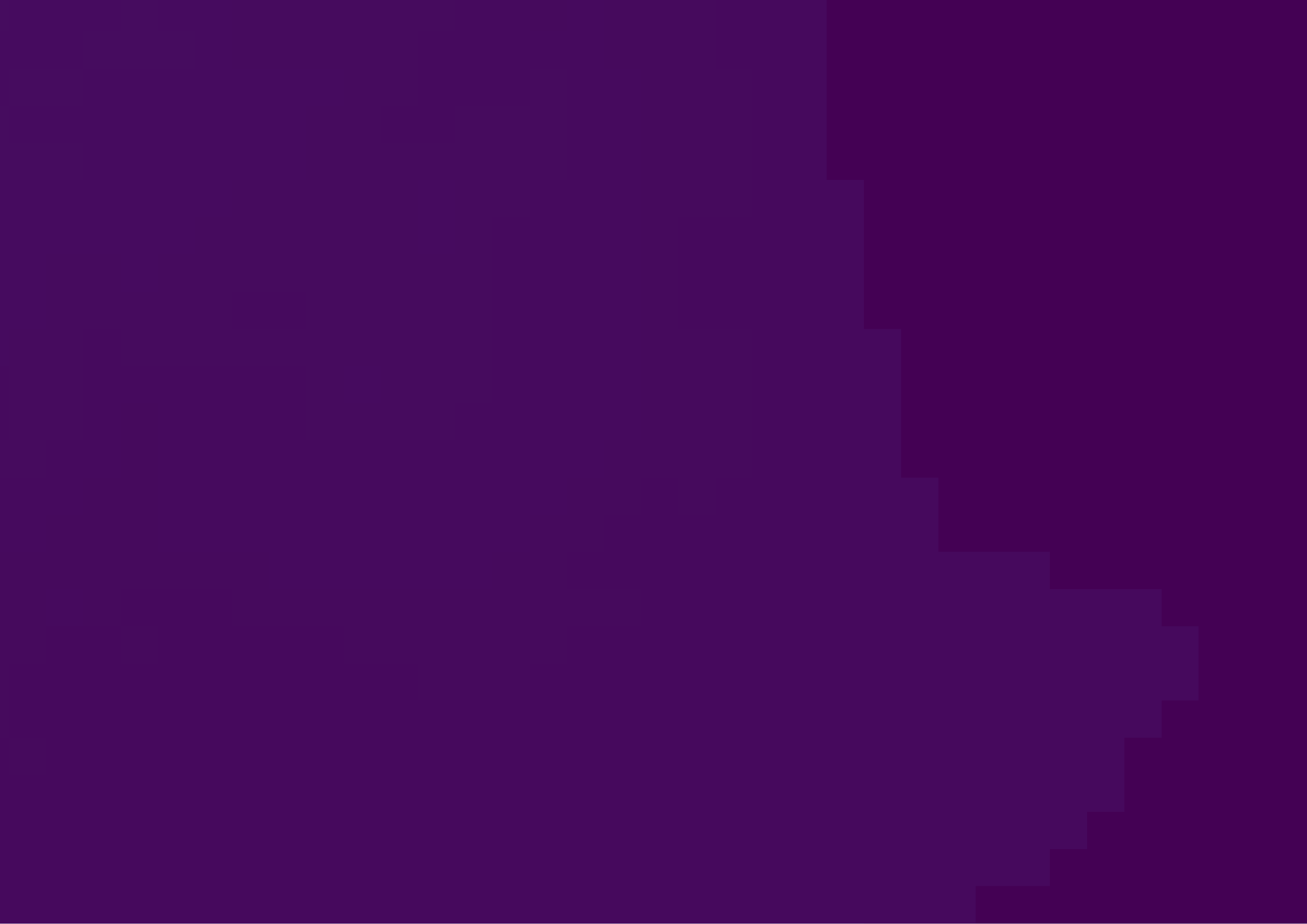}}
            & \rotatebox{90}{\includegraphics[height=\heightcomp]{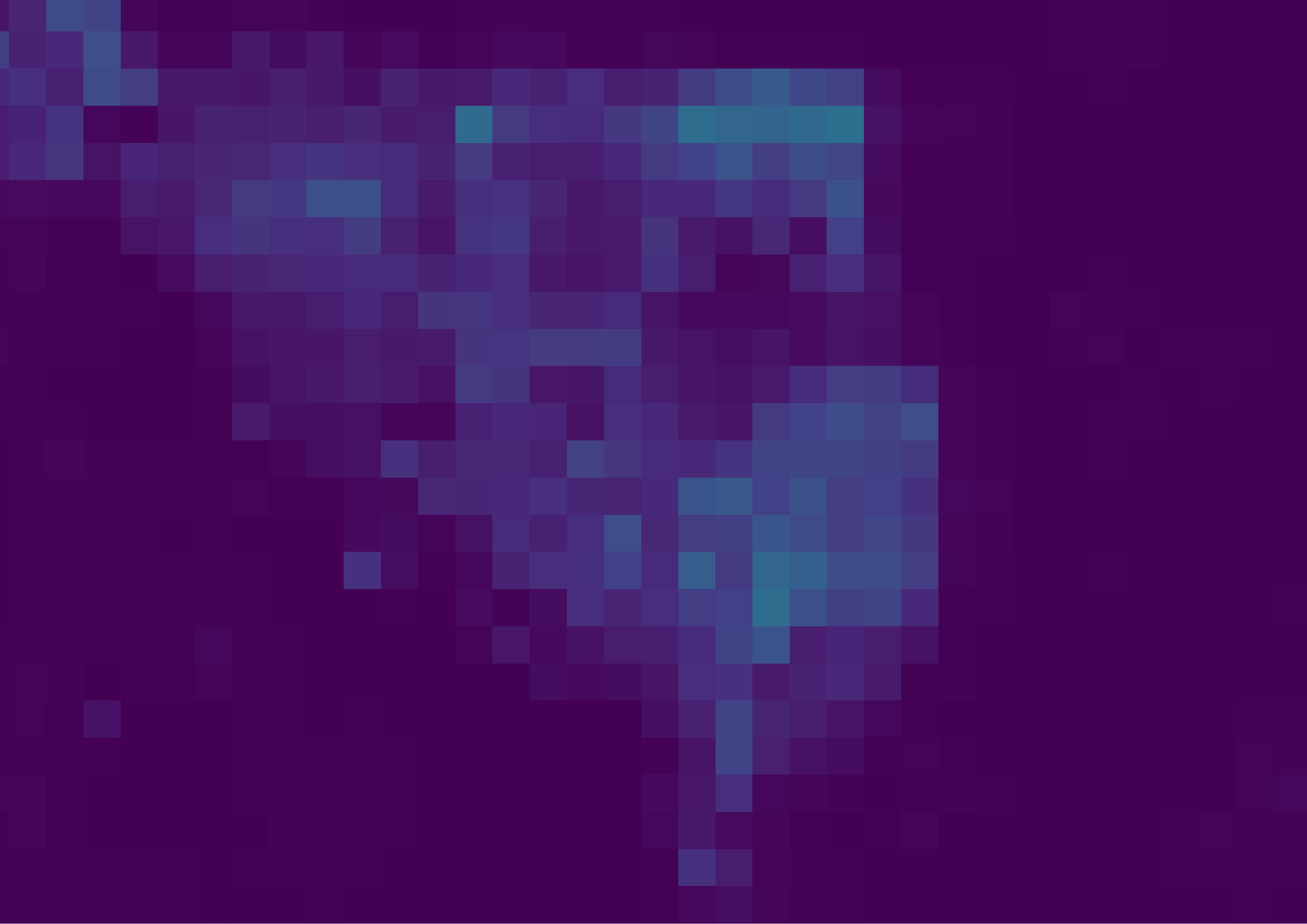}} \\

\midrule
            &
            & \parbox{21mm}{\centering VHR} 
            % & \parbox{21mm}{\centering Sentinel-2}
            & \parbox{21mm}{\centering WorldPop \\ Builtup.}
            & \parbox{21mm}{\centering HRPDM \\ Meta}
            & \parbox{21mm}{\centering Kontur} 
            & \parbox{21mm}{\centering Landscan Global}  
            & \parbox{21mm}{\centering \textsc{Pomelo}} \\
            
\multirow{3}{*}[3em]{\rotatebox{90}{\textbf{\parbox{35mm}{\centering Kingi, Rwanda}}}}
            & \rotatebox{90}{\textbf{\parbox{35mm}{\centering High Resolution}}}
            & \rotatebox{90}{\includegraphics[height=\heightcomp]{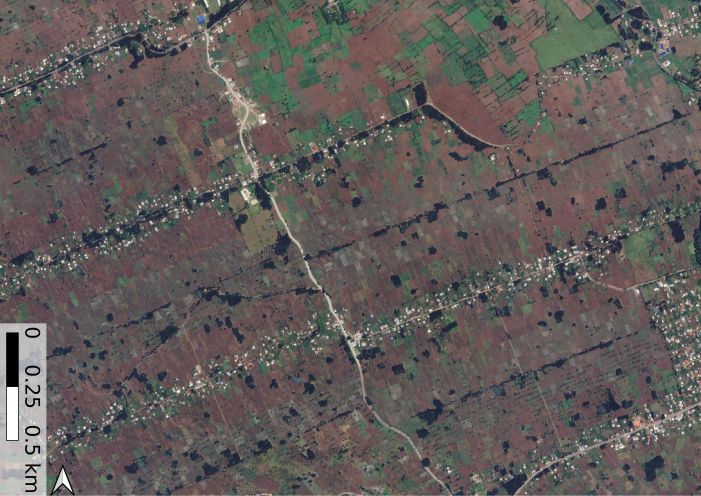}}
            % & \rotatebox{90}{\includegraphics[height=\heightcomp]{rwa2_sentinel2.png}}
            & \rotatebox{90}{\includegraphics[height=\heightcomp]{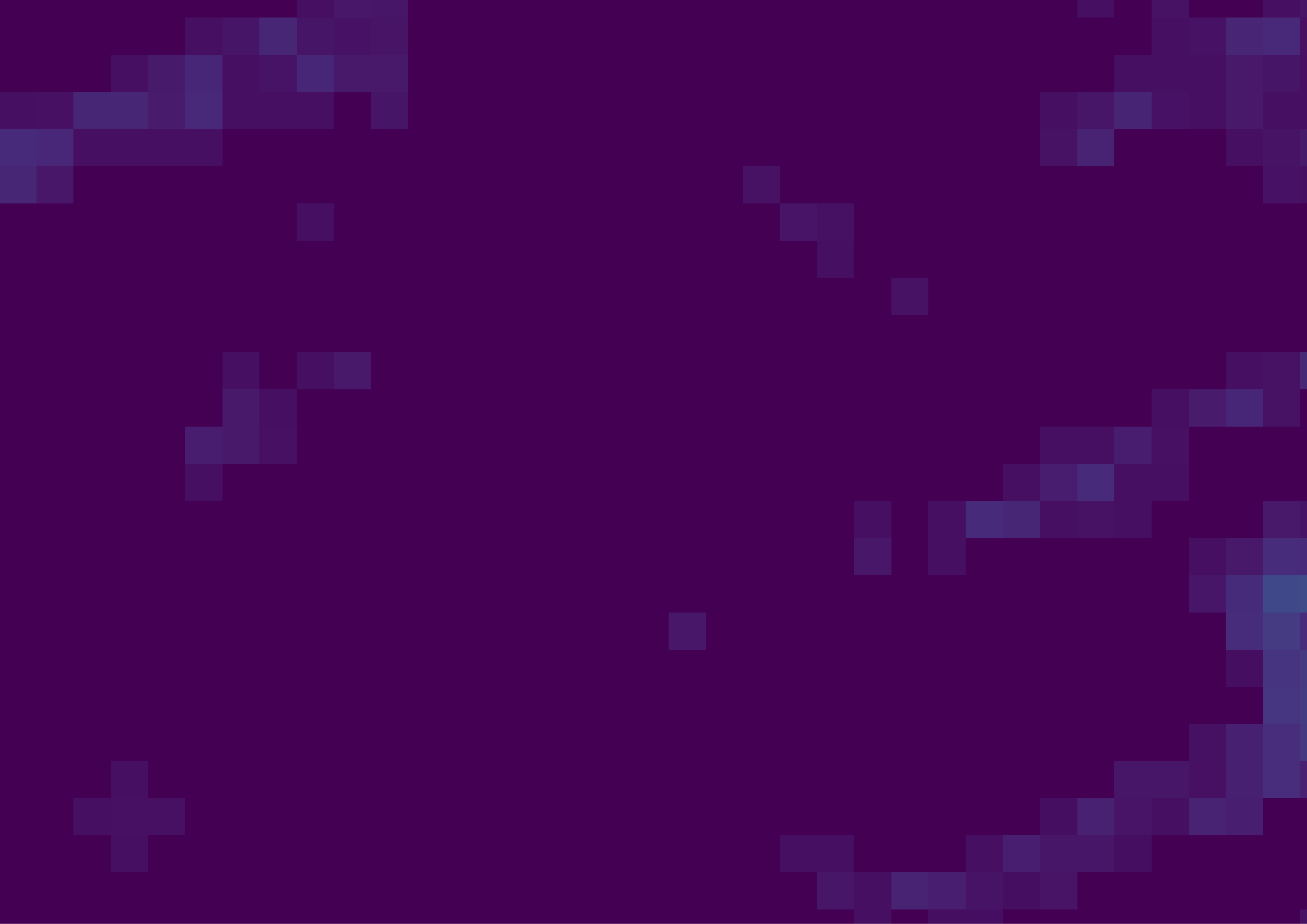}}
            & \rotatebox{90}{\includegraphics[height=\heightcomp]{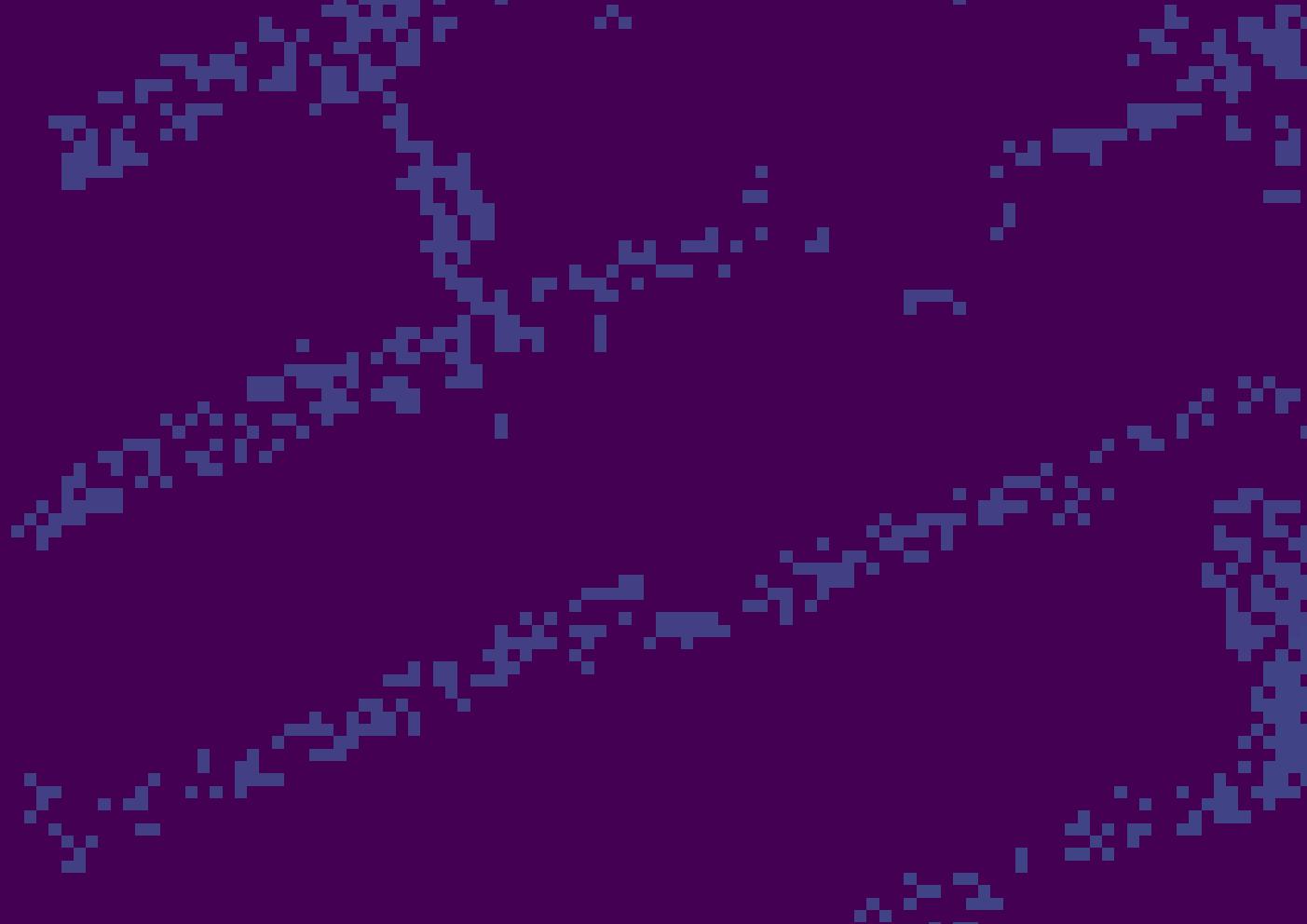}}
            & \rotatebox{90}{\includegraphics[height=\heightcomp]{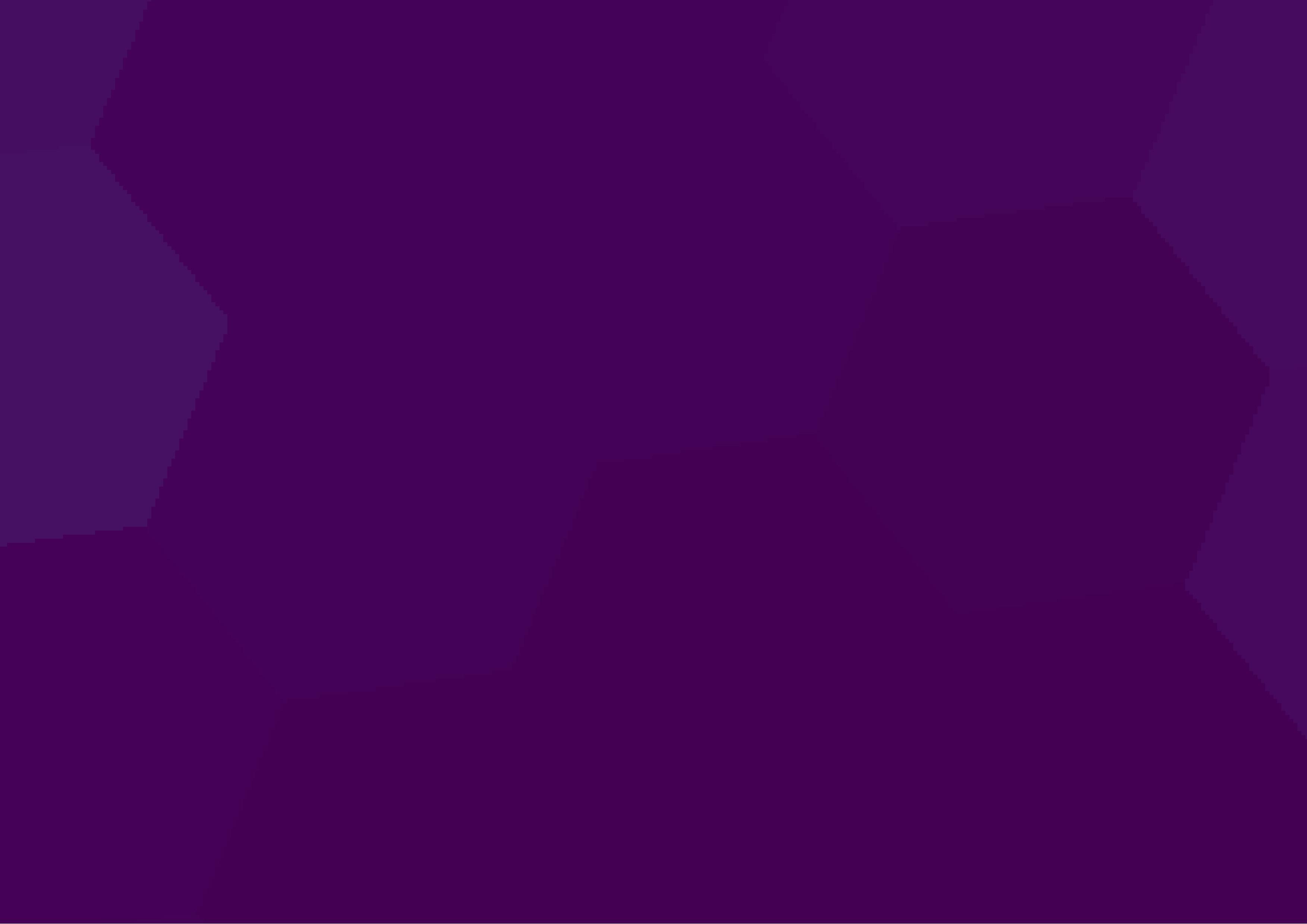}}
            & \rotatebox{90}{\includegraphics[height=\heightcomp]{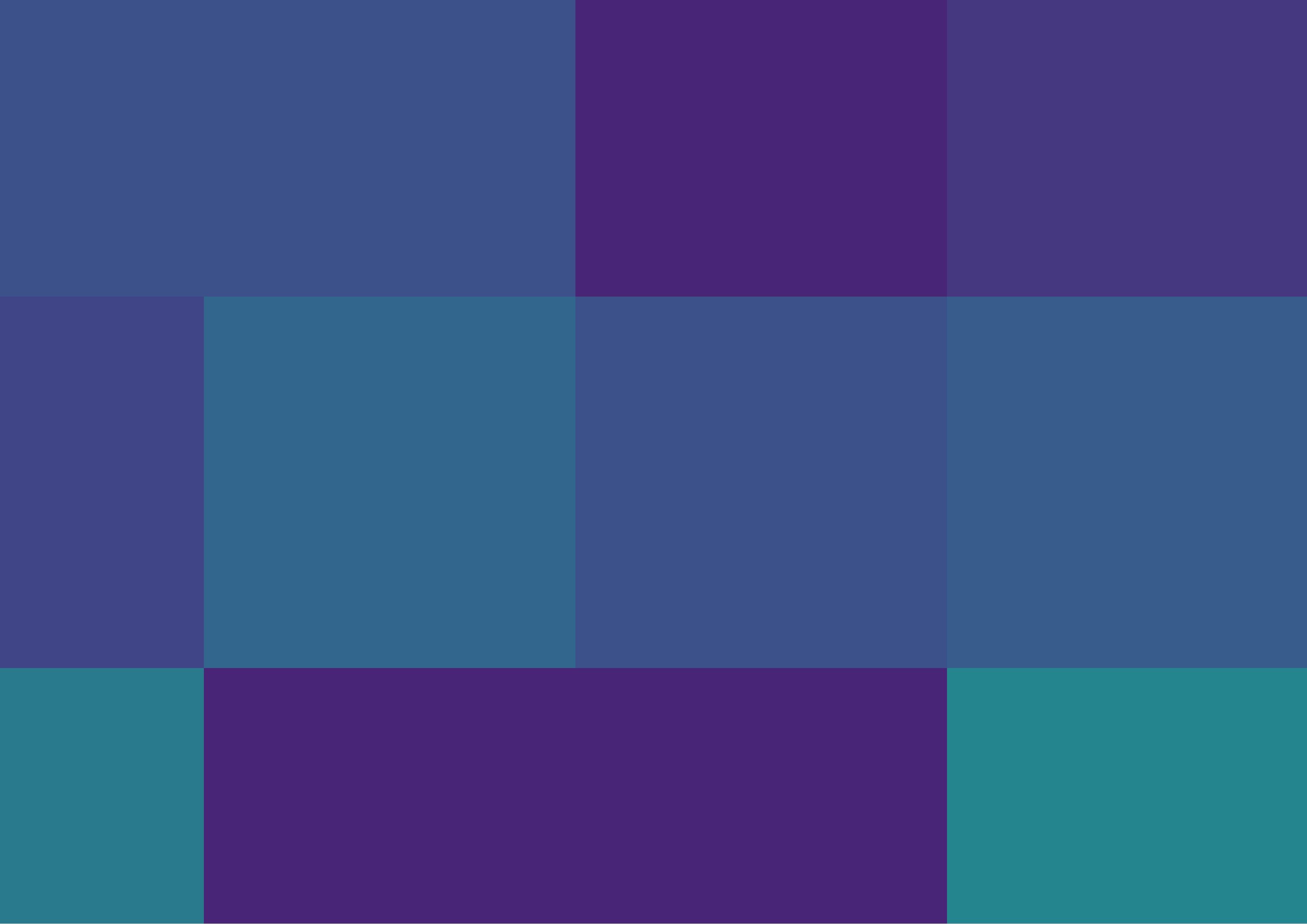}}
            & \rotatebox{90}{\includegraphics[height=\heightcomp]{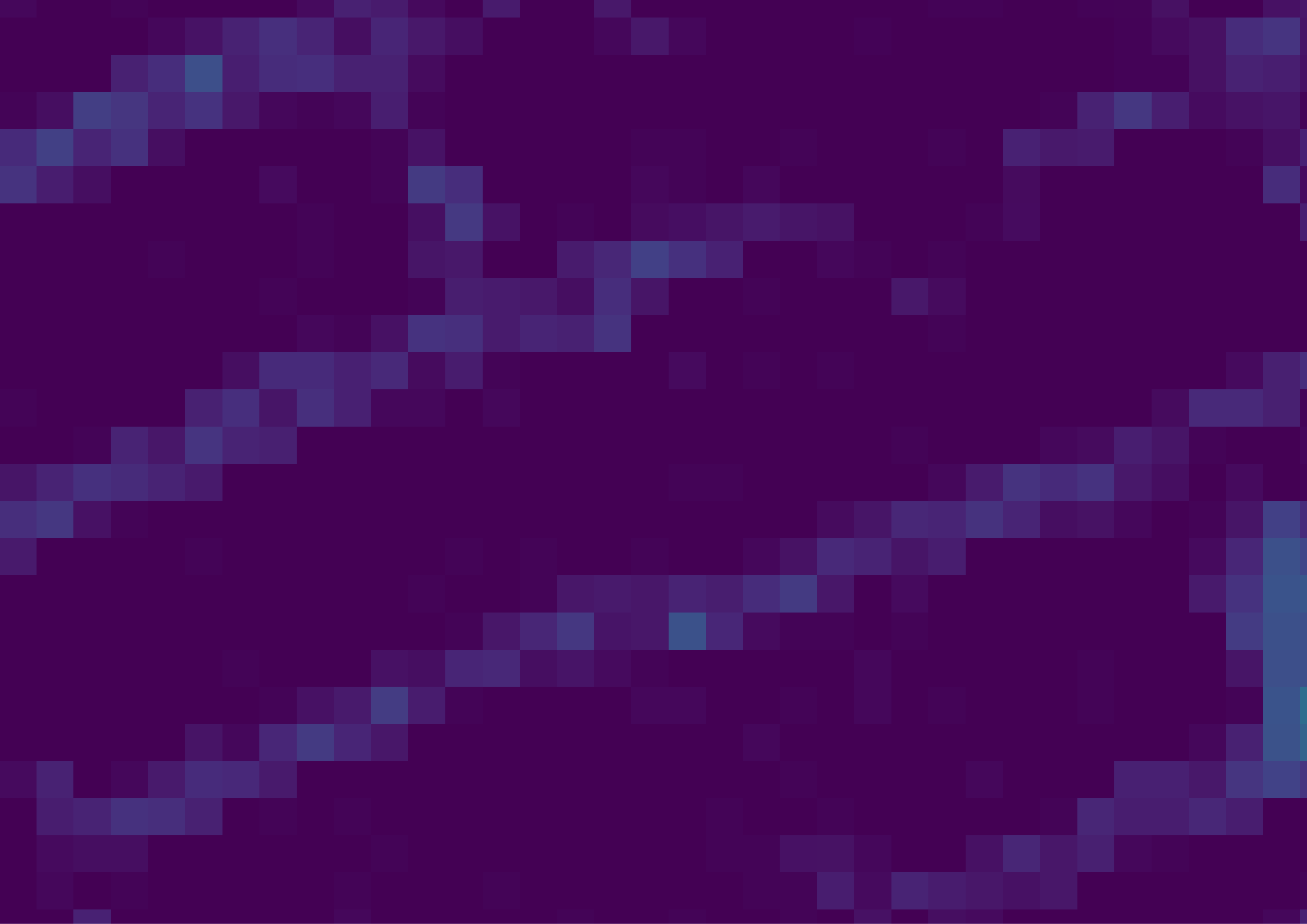}} \\

            &
            & \parbox{21mm}{\centering Sentinel-2}
            & \parbox{21mm}{\centering GPWv4} 
            & \parbox{21mm}{\centering GHS-Pop} 
            & \parbox{21mm}{\centering WorldPop}
            & \parbox{21mm}{\centering Bag-of-\textsc{Popcorn}} \\

            & \rotatebox{90}{\textbf{\parbox{35mm}{\centering Medium Resolution}}}
            & \rotatebox{90}{\includegraphics[height=\heightcomp]{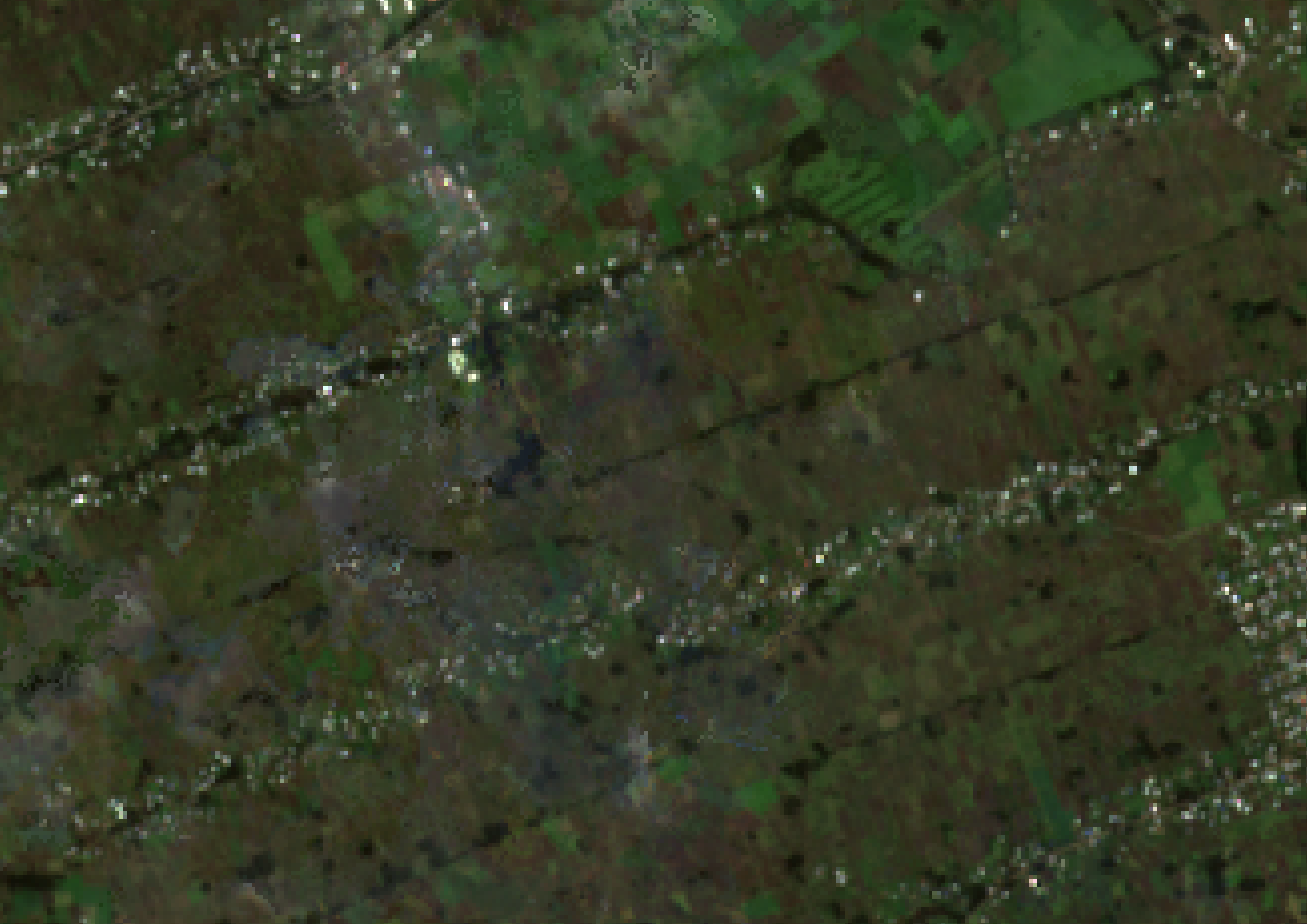}}
            & \rotatebox{90}{\includegraphics[height=\heightcomp]{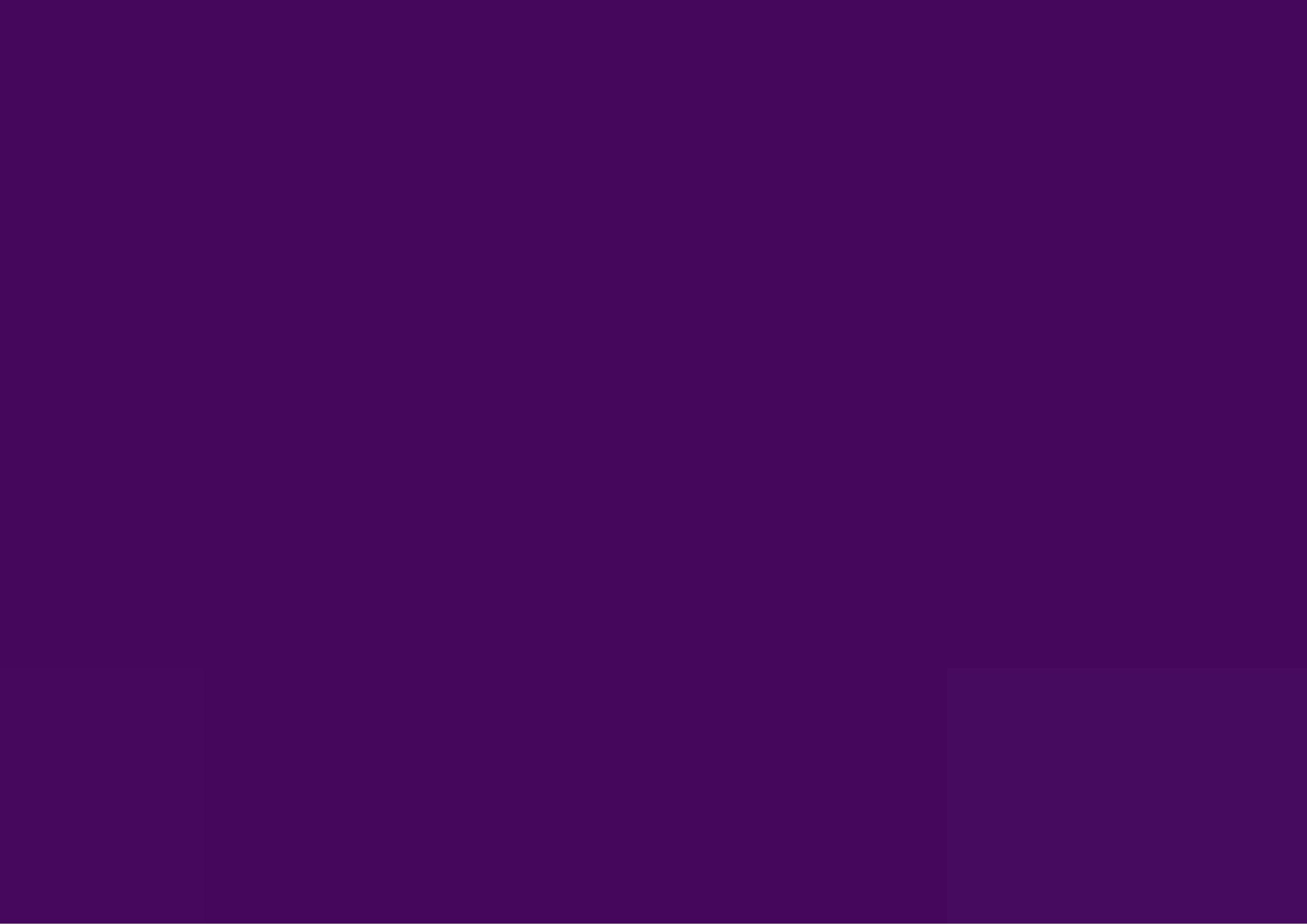}}
            & \rotatebox{90}{\includegraphics[height=\heightcomp]{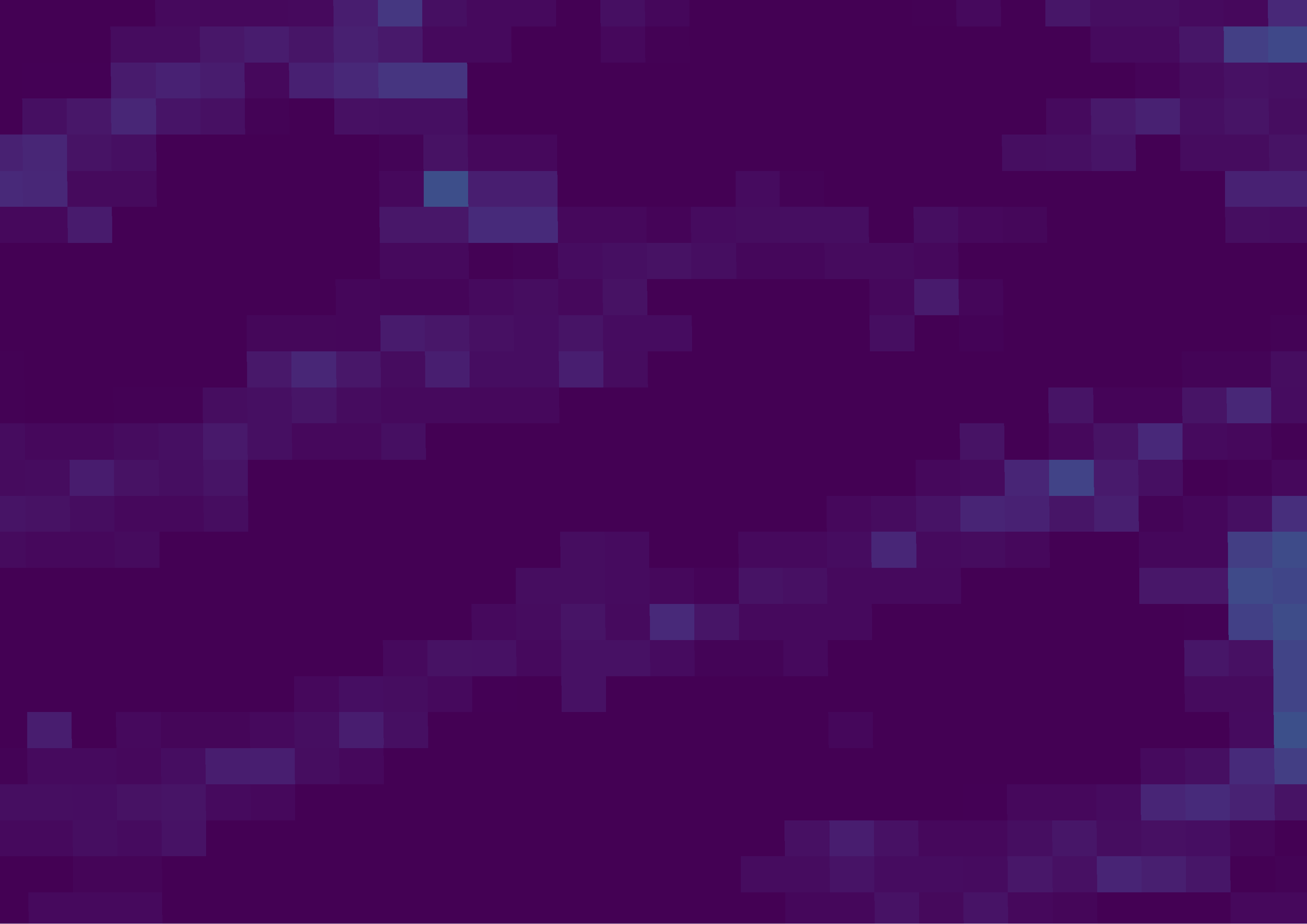}}
            & \rotatebox{90}{\includegraphics[height=\heightcomp]{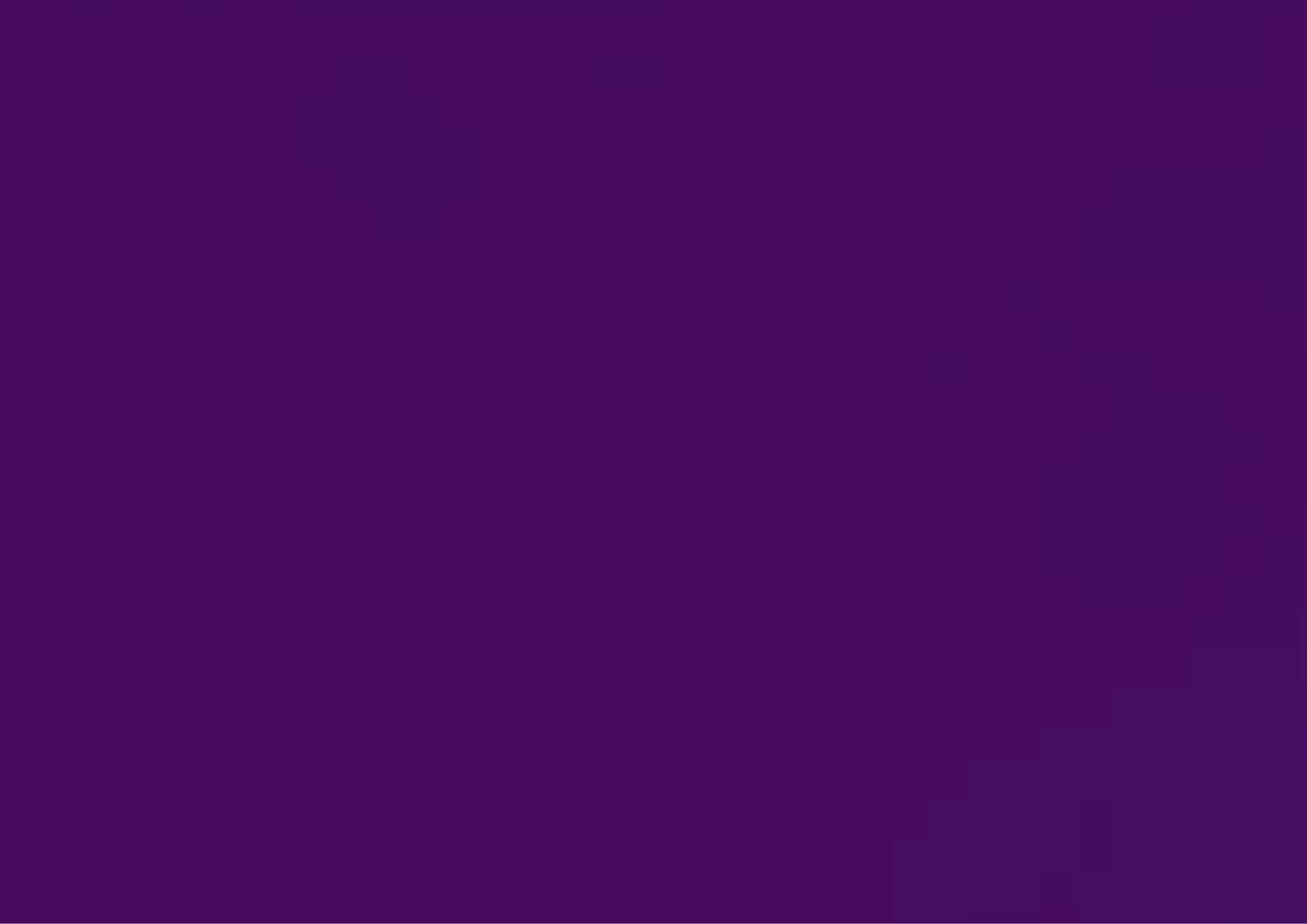}}
            & \rotatebox{90}{\includegraphics[height=\heightcomp]{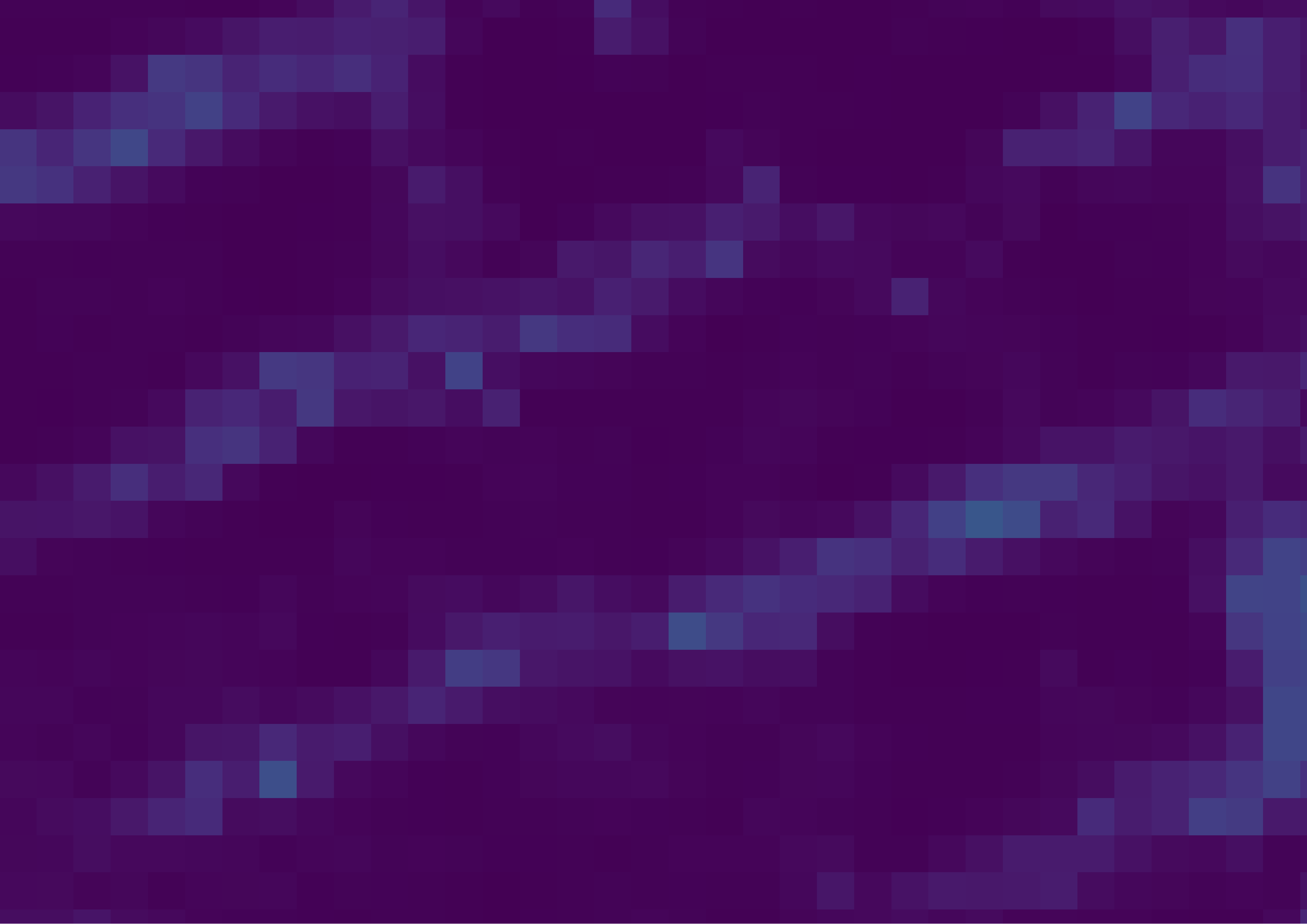}}\\
\midrule

            &
            & \parbox{21mm}{\centering VHR} 
            % & \parbox{21mm}{\centering Sentinel-2} 
            & \parbox{21mm}{\centering WorldPop \\ Builtup.}
            & \parbox{21mm}{\centering HRPDM \\ Meta}
            & \parbox{21mm}{\centering Kontur} 
            & \parbox{21mm}{\centering Landscan Global}  
            & \parbox{21mm}{\centering \textsc{Pomelo}} \\
            
\multirow{3}{*}[3em]{\rotatebox{90}{\textbf{\parbox{35mm}{\centering Zurich, Switzerland}}}}
            & \rotatebox{90}{\textbf{\parbox{35mm}{\centering High Resolution}}}
            & \rotatebox{90}{\includegraphics[height=\heightcomp]{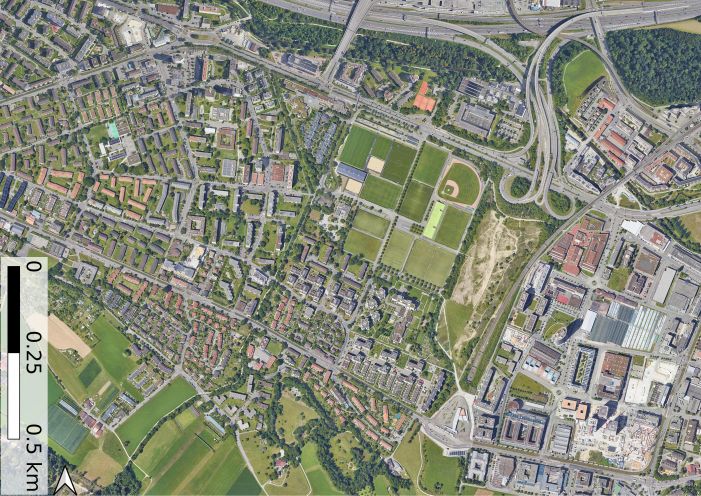}}
            % & \rotatebox{90}{\includegraphics[height=\heightcomp]{zh1_sentinel2.png}}
            & \rotatebox{90}{\includegraphics[height=\heightcomp]{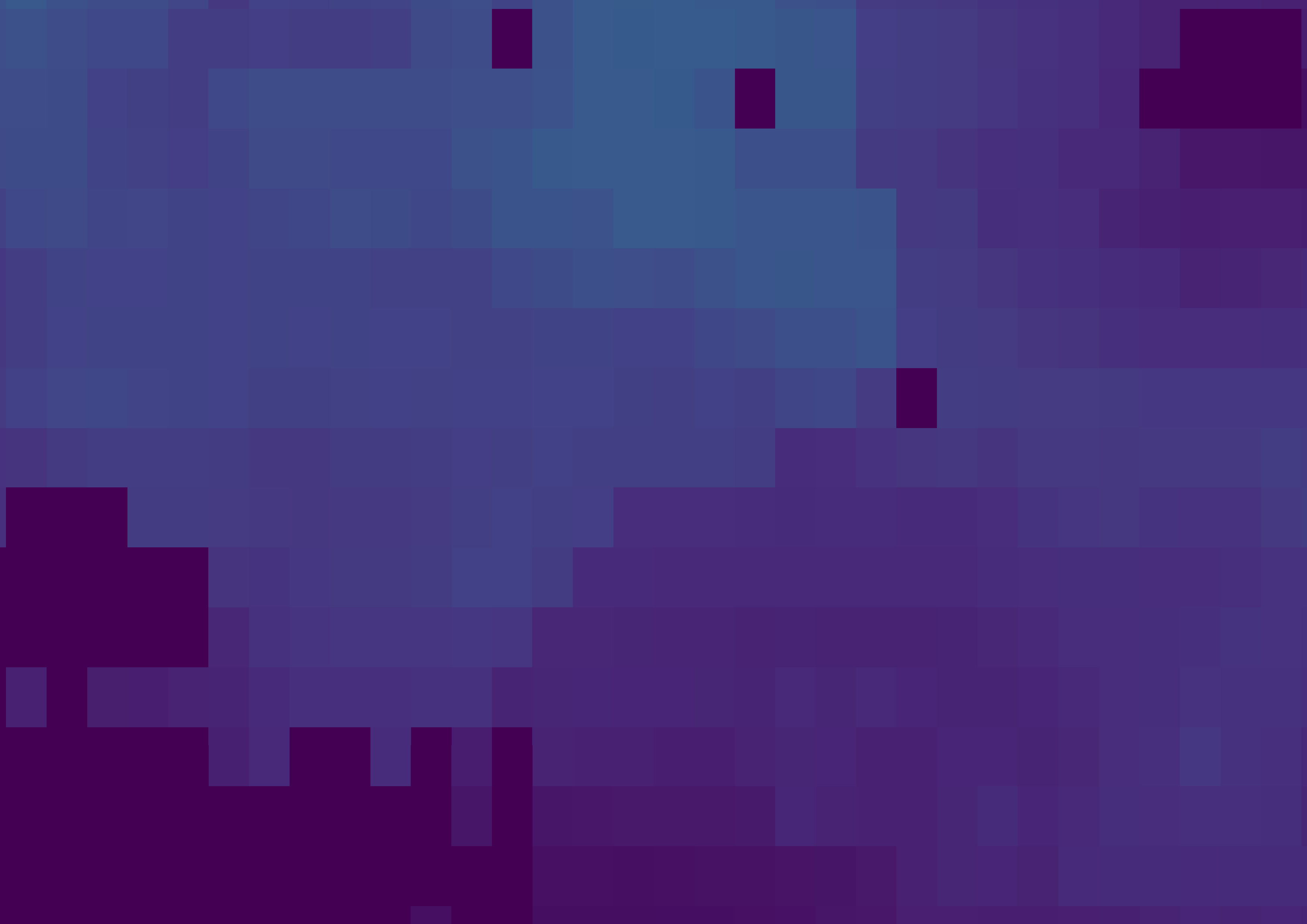}}
            & \rotatebox{90}{\includegraphics[height=\heightcomp]{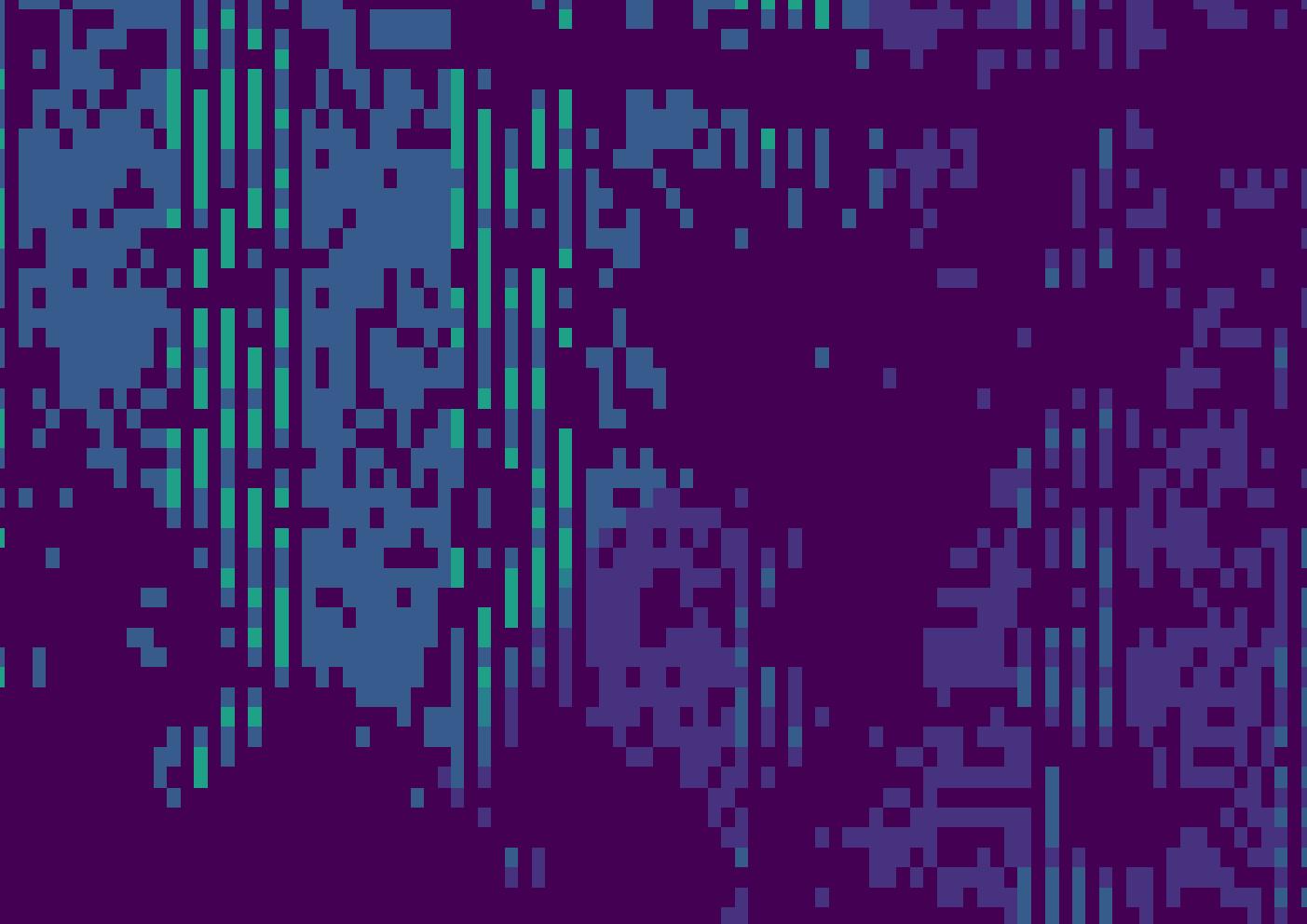}}
            & \rotatebox{90}{\includegraphics[height=\heightcomp]{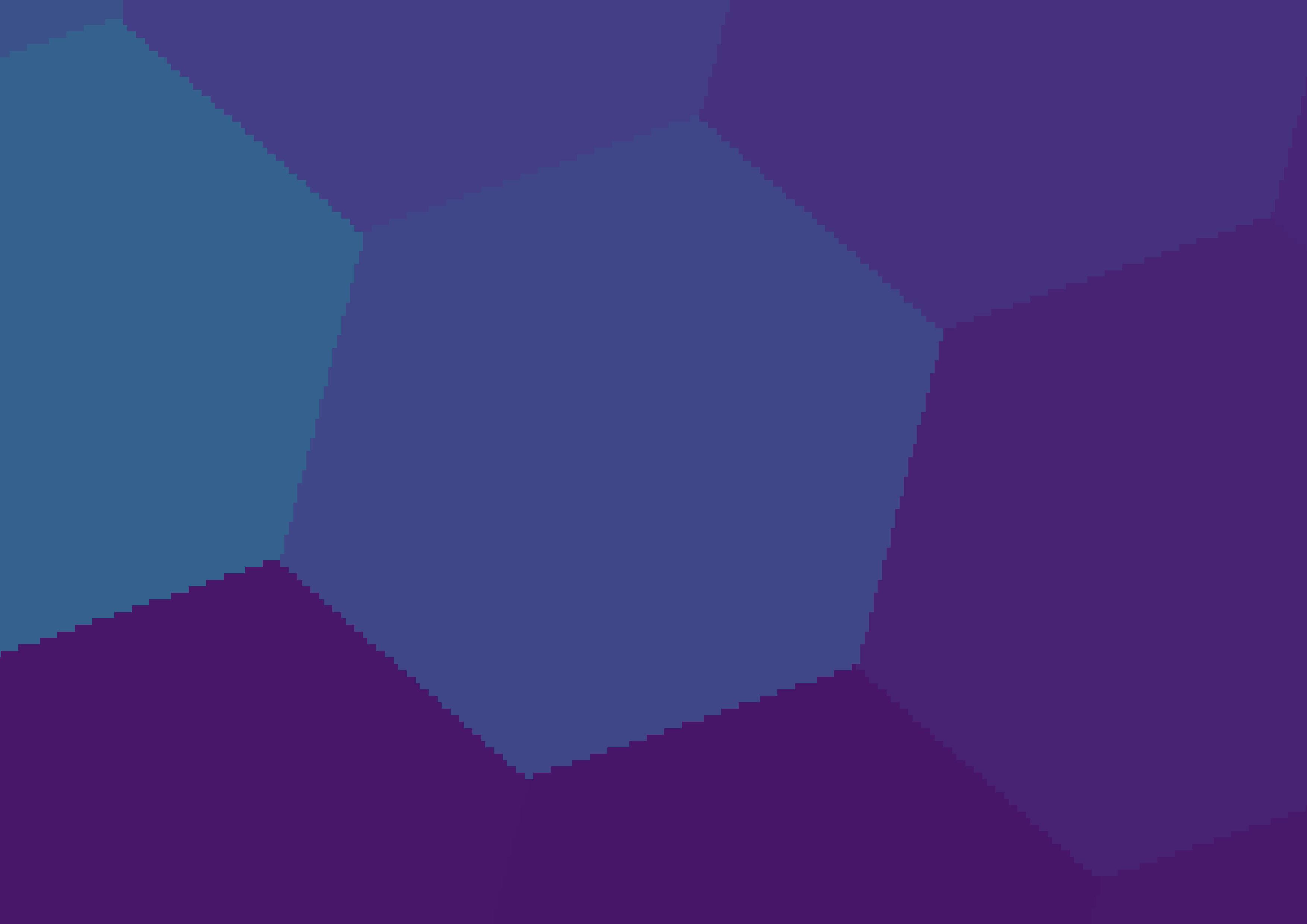}}
            & \rotatebox{90}{\includegraphics[height=\heightcomp]{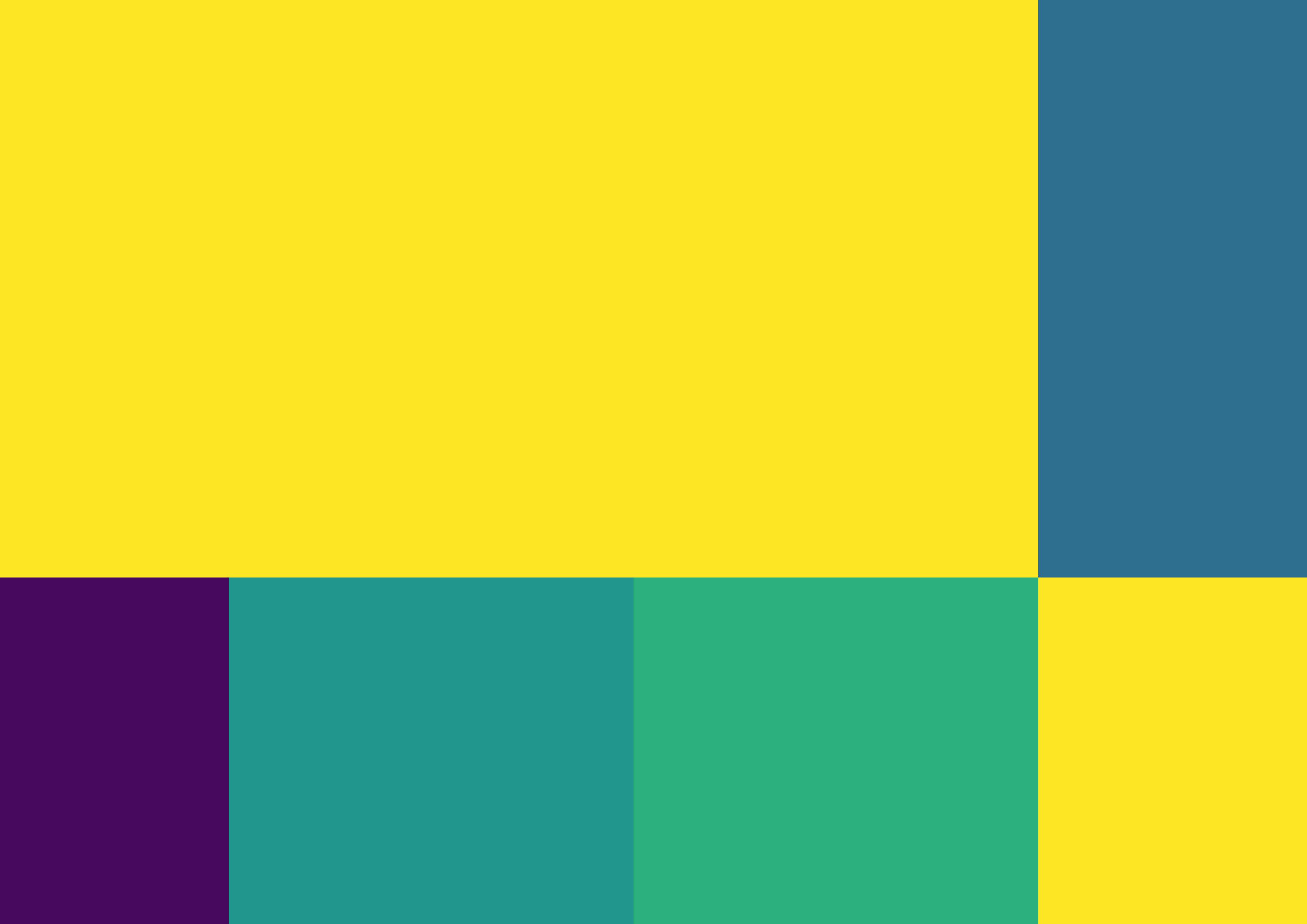}}
            & \rotatebox{90}{\includegraphics[height=\heightcomp]{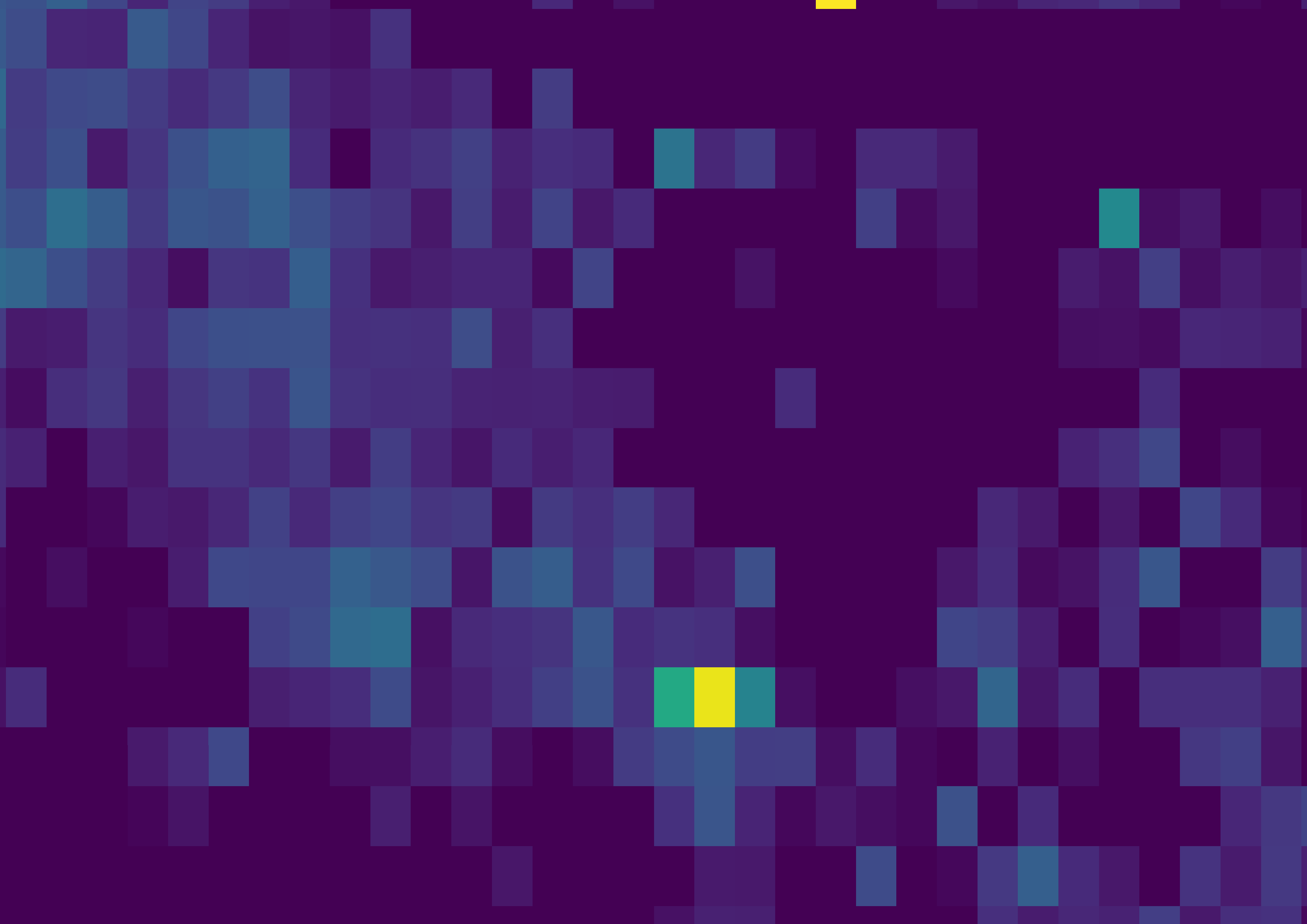}} \\

            &
            & \parbox{21mm}{\centering Sentinel-2}
            & \parbox{21mm}{\centering GPWv4} 
            & \parbox{21mm}{\centering GHS-Pop} 
            & \parbox{21mm}{\centering WorldPop}
            & \parbox{21mm}{\centering Bag-of-\textsc{Popcorn}} 
            & \parbox{21mm}{\centering Ground Truth} \\
            
            & \rotatebox{90}{\textbf{\parbox{35mm}{\centering Medium Resolution}}}
            & \rotatebox{90}{\includegraphics[height=\heightcomp]{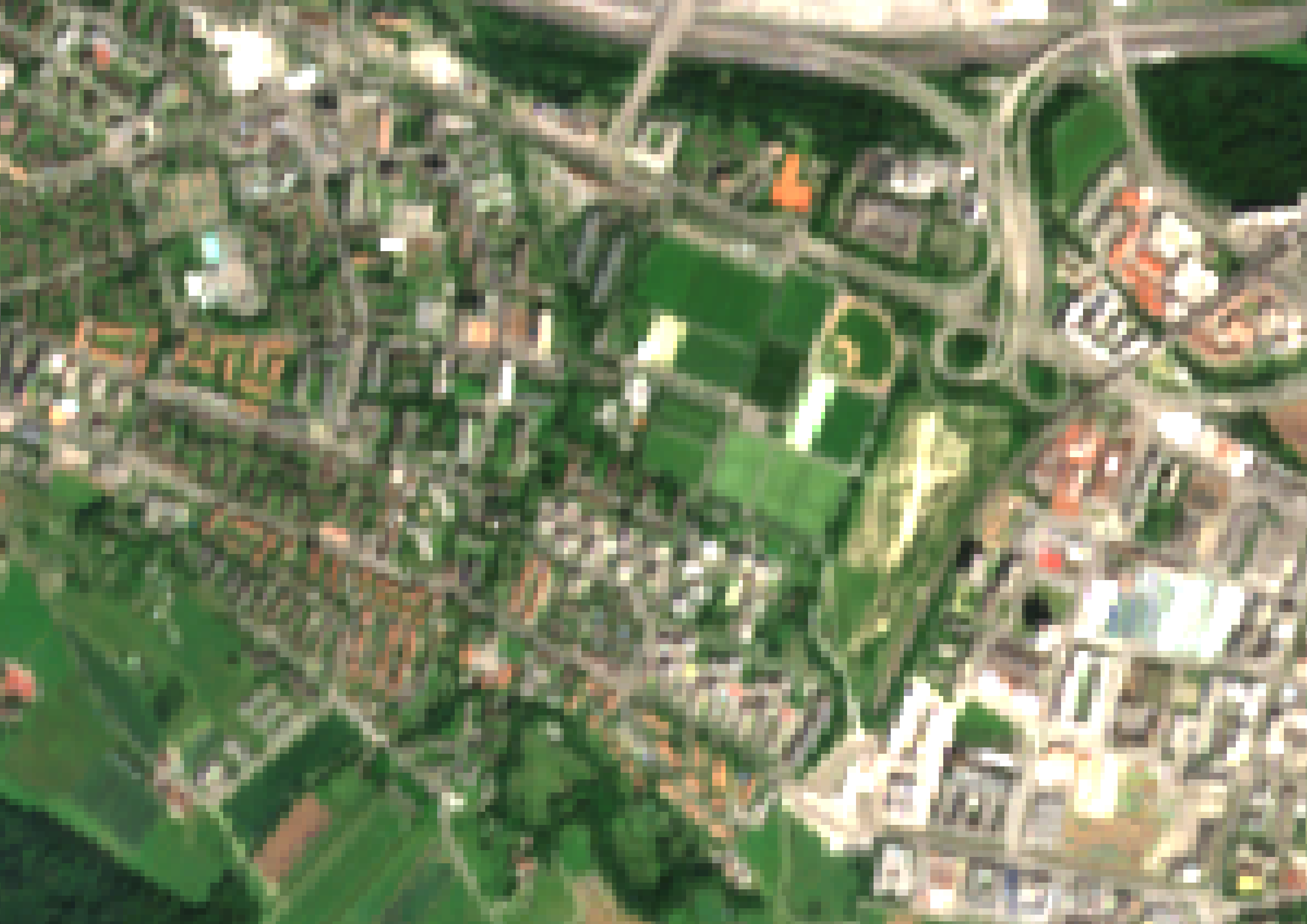}}
            & \rotatebox{90}{\includegraphics[height=\heightcomp]{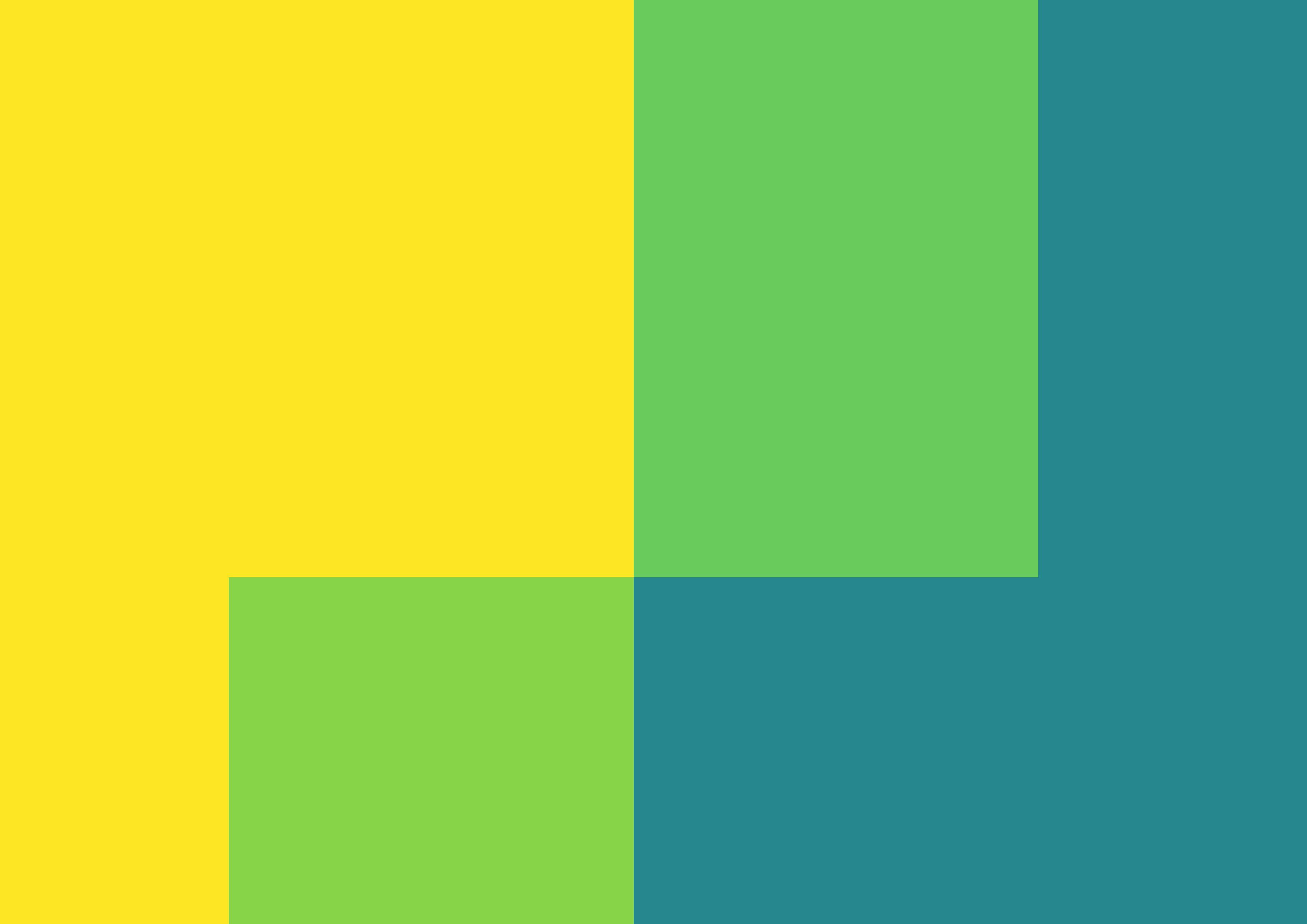}}
            & \rotatebox{90}{\includegraphics[height=\heightcomp]{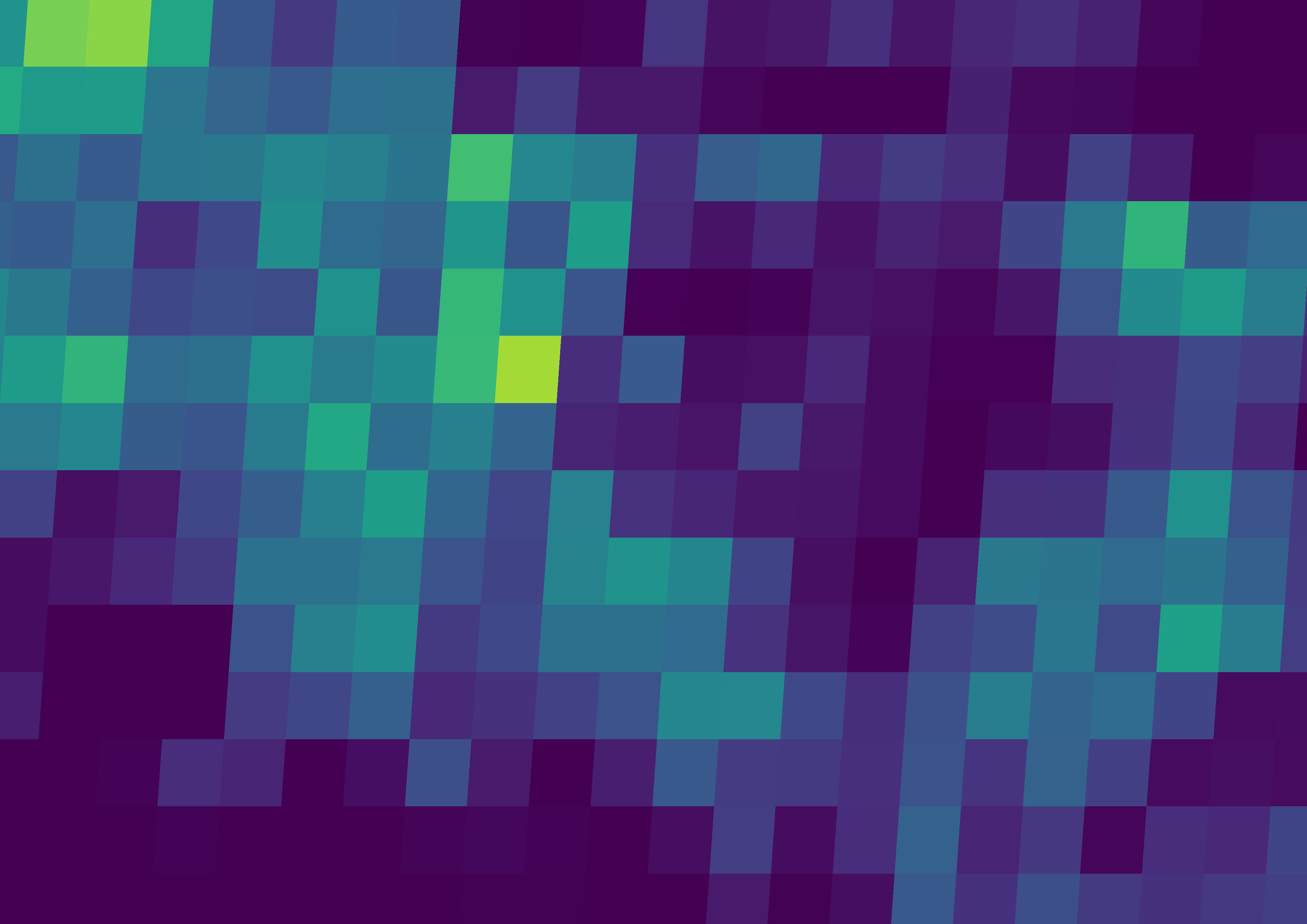}}
            & \rotatebox{90}{\includegraphics[height=\heightcomp]{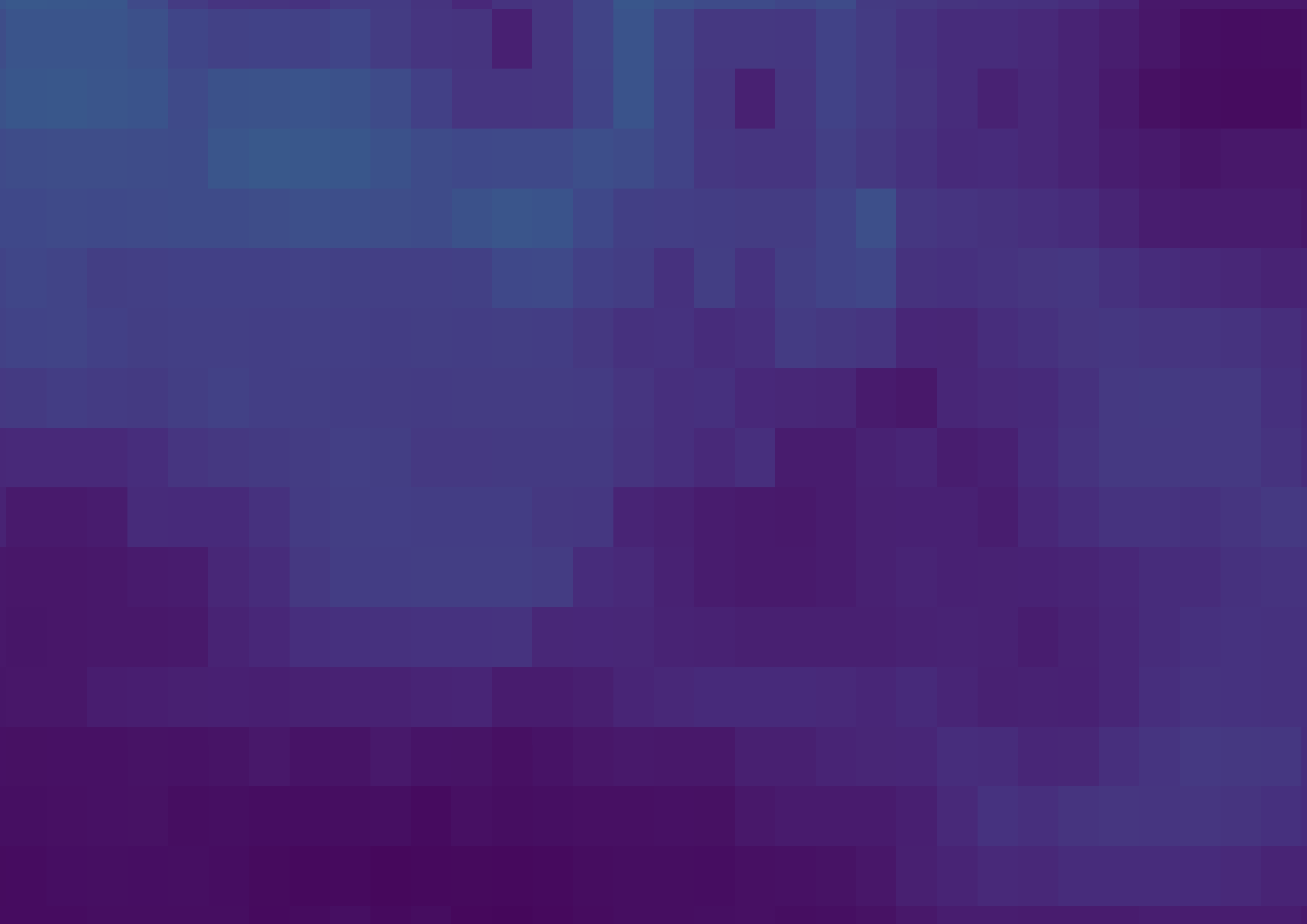}}
            & \rotatebox{90}{\includegraphics[height=\heightcomp]{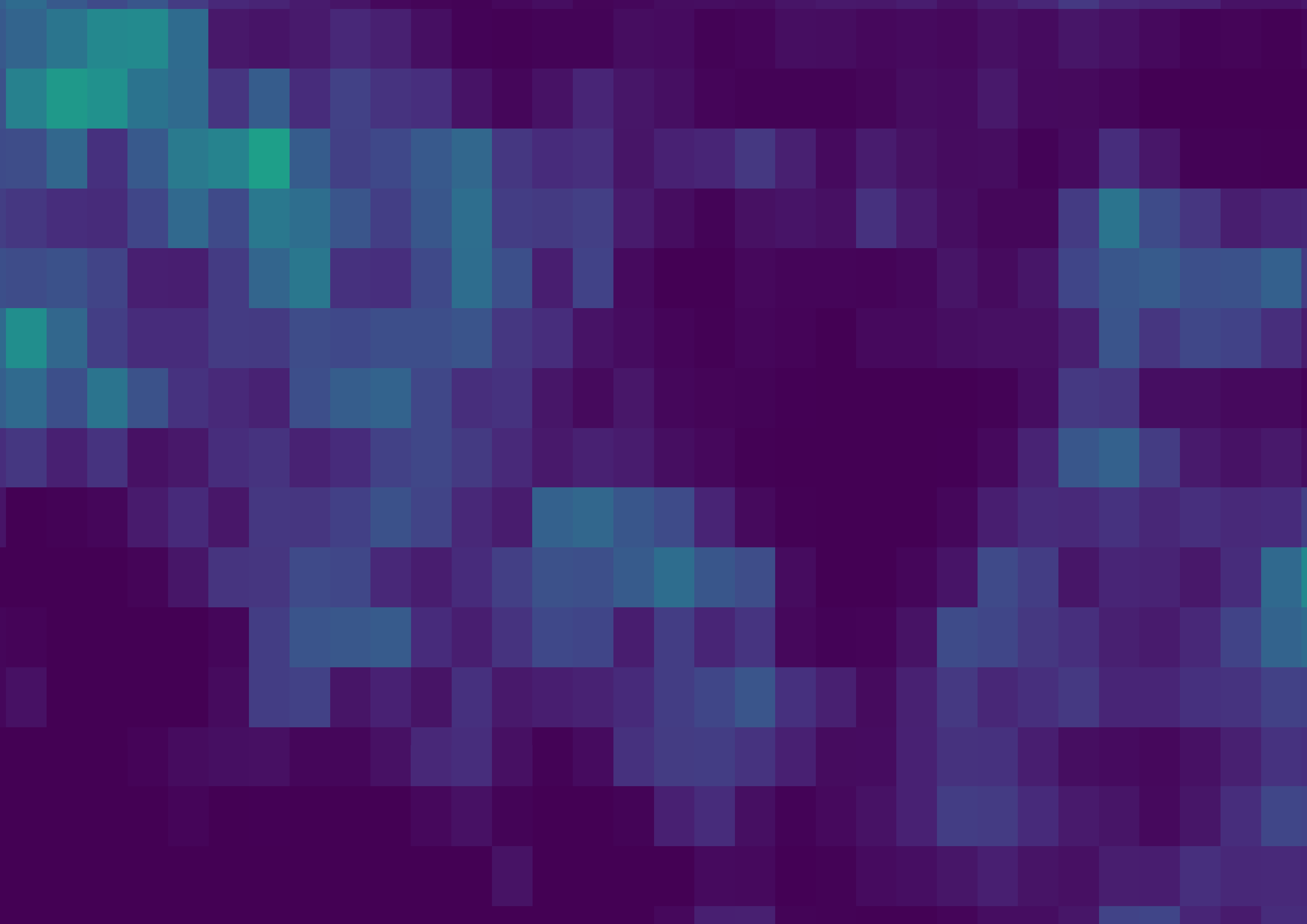}} 
            & \rotatebox{90}{\includegraphics[height=\heightcomp]{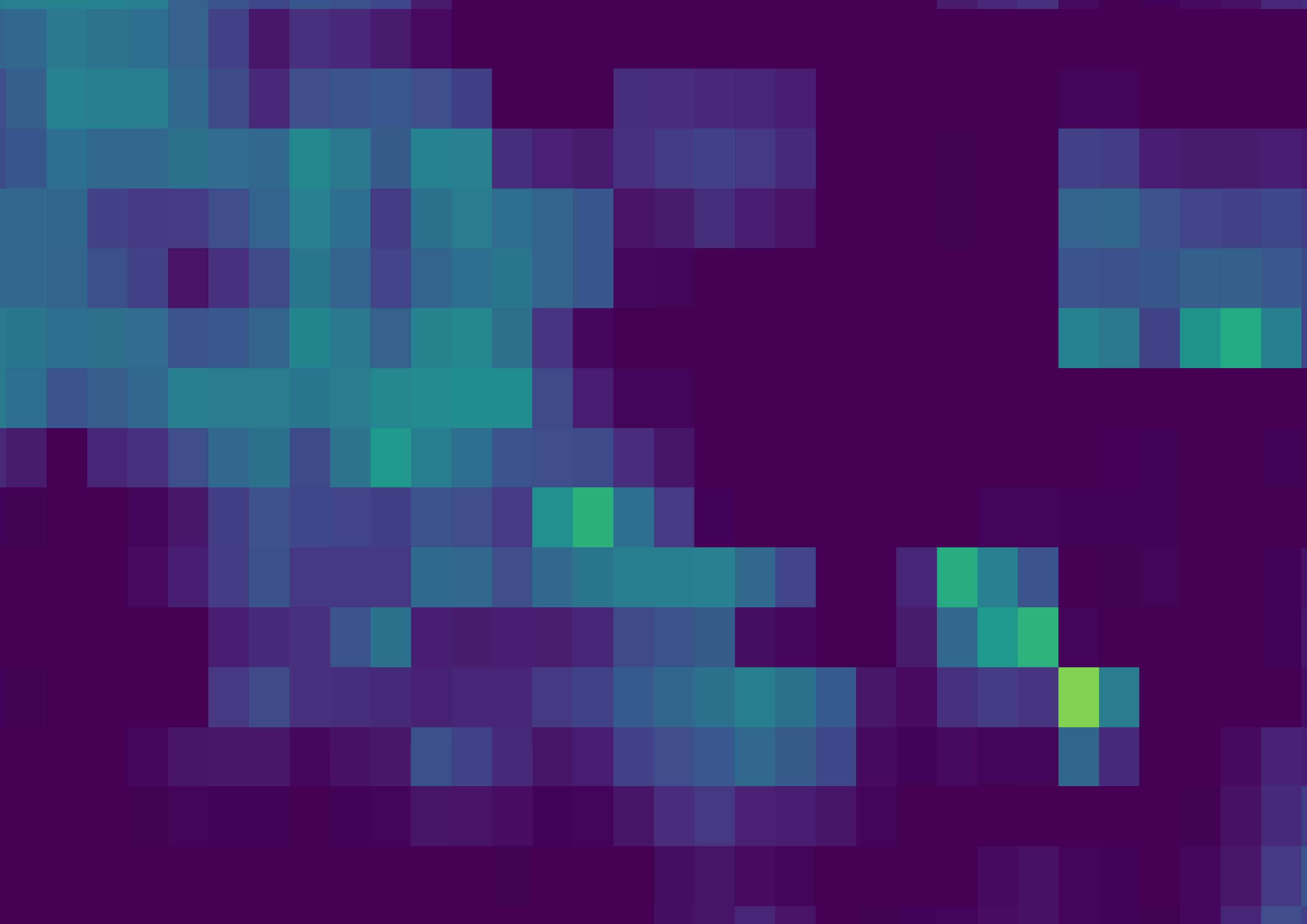}} \\
            
 & & \multicolumn{6}{c}{\includegraphics[width=120mm]{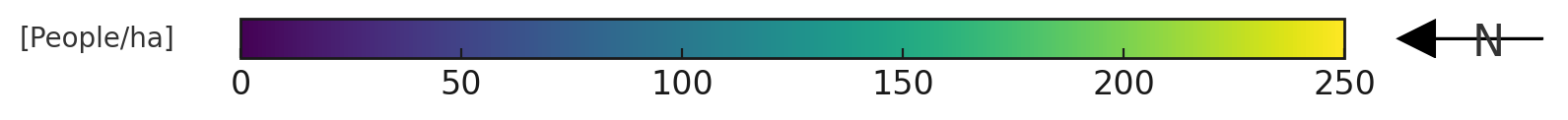}}
\end{tabular}
}
\caption{Comparison of population maps for the Mahama refugee camp (Rwanda), Kingi (Rwanda), and Zurich, Switzerland. The first column shows very high-resolution images~\citep{googlemaps2023}, while the second column shows the Sentinel-2 RGB composites~\citep{sentinel2_composite} for reference. Bag-of-\textsc{Popcorn} maps were resampled to the same grid as WorldPop to ease visual comparison. {All maps claim validity for 2020}. Best viewed on screen. }
\label{tab:map_comparison}
\end{figure*}

Figure~\ref{tab:map_comparison} provides a visual comparison, contrasting population maps created by the Bag-of-\textsc{Popcorn} model with those of WorldPop, \textsc{Pomelo}, and GHS-Pop. For the examples from Rwanda, no high-resolution ground truth is available to verify the estimates.
Bag-of-\textsc{Popcorn}'s map prediction for the Mahama Refugee Camp indicates that the model has identified the human-made structures of the camp. At the same time, it assigns them comparably lower densities than \textsc{Pomelo} and GHS-Pop. By contrast, both WorldPop maps completely missed the camp, possibly owing to the use of outdated base data.
Turning to Kingi in Rwanda, the Bag-of-\textsc{Popcorn} estimates are very similar to those of its most accurate competitor \textsc{Pomelo}. In contrast, WorldPop seems to fall short when it comes to detecting thin rows of buildings along the roads, a rather frequent building pattern in low-density, rural regions that have basic road infrastructure.
In Zurich, Switzerland, Bag-of-\textsc{Popcorn} and \textsc{Pomelo} again predict the most credible overall density pattern, although the \textsc{Pomelo} map contains one implausible spike. Those two methods also stand out as the only ones that correctly handle zero-density regions within the urban fabric such as the green fields in the image center. The WorldPop maps exhibit excessive smoothing, possibly due to inaccuracies of the underlying built-up area layer.
    
\subsection{Further Observations}

\paragraph{Built-up scores vs.\ building footprints} Somewhat surprisingly, our results show that, when it comes to population disaggregation, high-resolution building counts do not necessarily provide superior guidance compared to low-resolution built-up area scores. We hypothesize that the continuous, soft built-up scores contain implicit information about (aggregate) building size.
We note that the effect is more apparent in developed regions (Switzerland and Puerto Rico), possibly because the built-up area detector was originally trained with data from the US and Australia and is less familiar with the building structures of Rwanda.

\paragraph{Resolution and Synchronization} Another unexpected result was that the variant of \textsc{Popcorn} that has access to high-resolution building counts tends to perform worse than its purely Sentinel-based counterpart. On the one hand, this could be one more instance of the issue discussed in the previous paragraph.
But other factors might also be at play. The high-resolution building footprints, derived from VHR satellite data or airborne imagery, are not temporally aligned with the Sentinel data used to derive occupancy, so in regions with substantial construction (or demolition) activity the observed state might differ, leading to inconsistencies between the two branches.
The Google Open Buildings, in particular, do not specify the period for which a building footprint is valid.

\section{Discussion}

In the following, we discuss the resulting model, by providing a visual analysis of the results, a discussion about the scalability of our method, and an ablation study considering the most important components of our methodology.

\subsection{Visual Examples}

\newcommand{\imgWidthParam}{0.45\textwidth}
\begin{table}[tp]
    \centering
    \resizebox{0.48\textwidth}{!}{%
    \begin{tabular}{c}
    \rotatebox{90}{\textbf{VHR}} \includegraphics[width=\imgWidthParam]{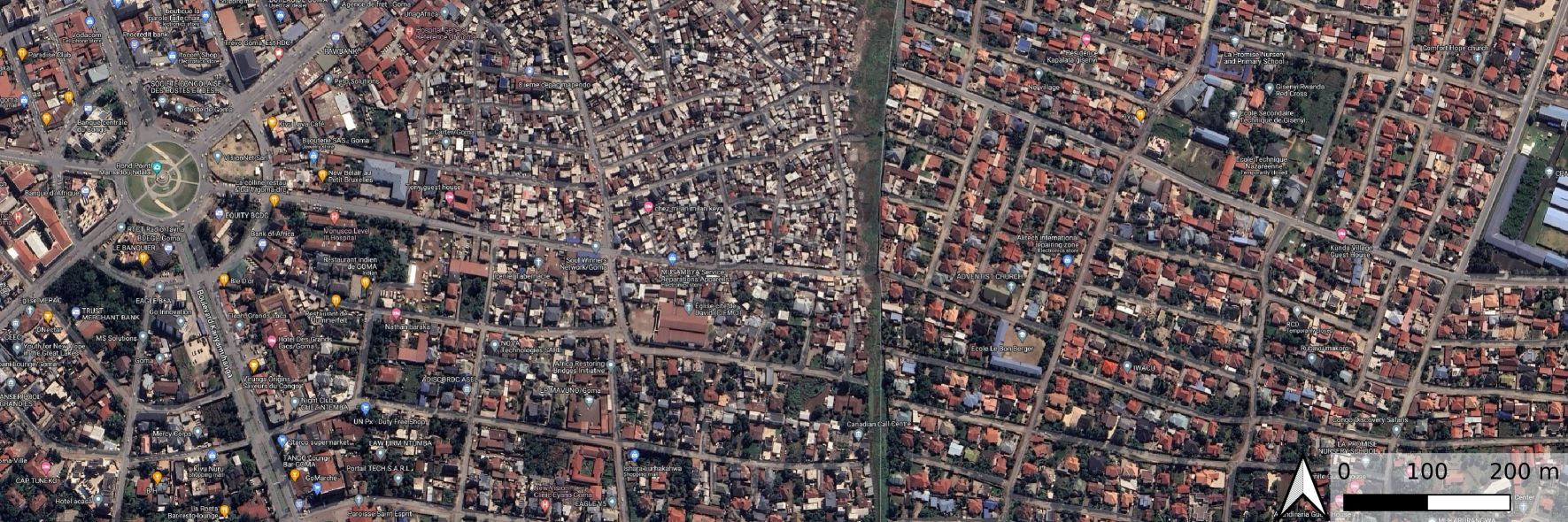} \\
    \rotatebox{90}{\textbf{Sentinel-2}} \includegraphics[width=\imgWidthParam]{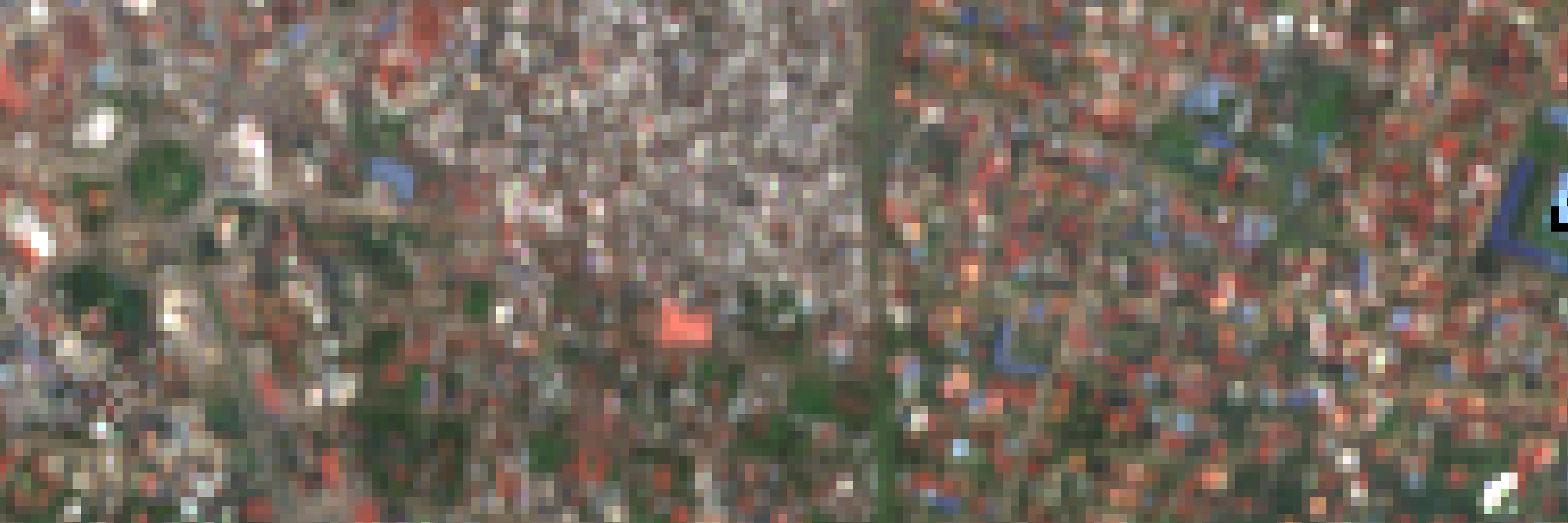} \\
    \rotatebox{90}{\textbf{Prediction}} \includegraphics[width=\imgWidthParam]{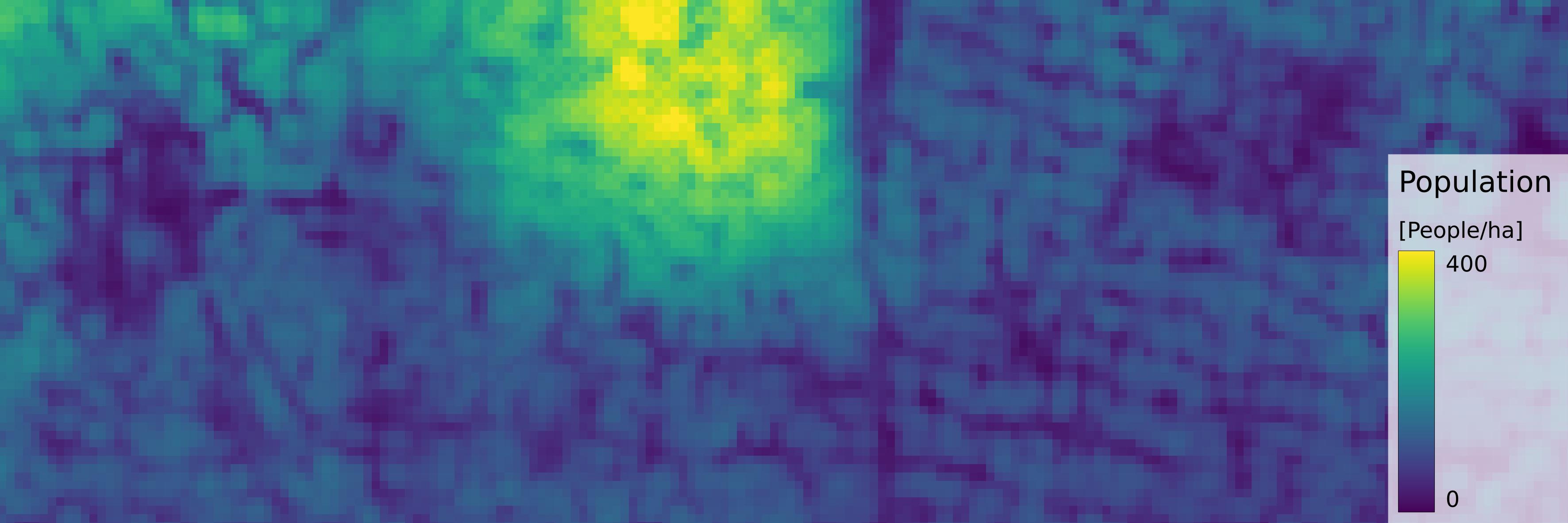} \\
    \rotatebox{90}{\textbf{Occupancy}} \includegraphics[width=\imgWidthParam]{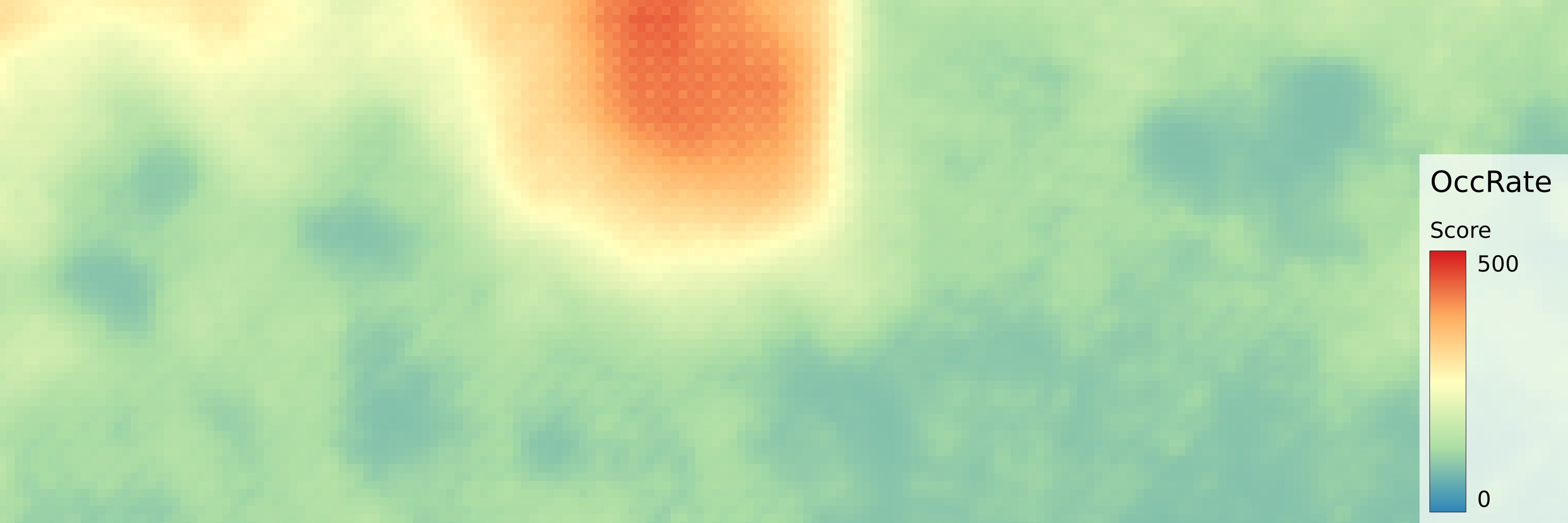} \\
    \rotatebox{90}{\textbf{BuiltUp score}} \includegraphics[width=\imgWidthParam]{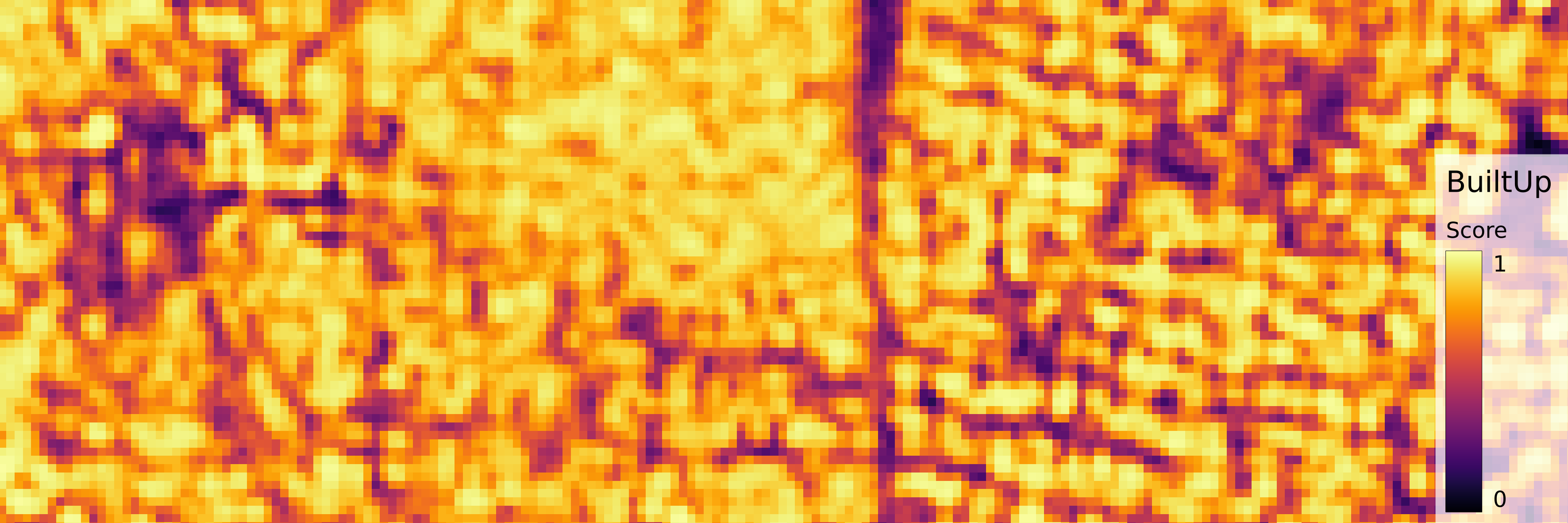} \\
\end{tabular}
}
\captionof{figure}{Comparison of population density estimates in the Goma-Gisenyi border region: The vertical strip is the boundary, the left part lies in the DRC, and the right in Rwanda. Sources: VHR~\citep{googlemaps2023}, Sentinel-2~\citep{sentinel2_composite}.}
\label{fig:performance_vis}
\end{table}

Figure~\ref{fig:performance_vis} displays a part of the Goma and Gisenyi, two cities at the border between the DRC and Rwanda respectively. The example illustrates the two streams of the Bag-of-\textsc{Popcorn} model. In the upper part of the left side lies the very densely populated Mapendo neighborhood. The built-up score is saturated, still, the model manages to recognize the substantially higher occupancy, and consequently output a higher population density. The figure displays also VHR imagery to visually confirm the prediction.
Arguably, the explicit maps of occupancy and built-up area score also afford our model a degree of interpretability, in the sense that physically meaningful intermediate quantities can be separately inspected and their distributions checked for plausibility.

In a further example from Puerto Rico, Figure~\ref{fig:pri_ex}, we showcase the ability of the model to distinguish between high-density (top-left) and low-density (top-right) residential zones. {The bottom-left sample demonstrates that the model correctly identifies the industrial area as uninhabited, while correctly assigning a non-zero population density to the neighboring residential area. }
Thanks to the dedicated, separately trained built-up area extractor, the model is capable of detecting isolated buildings that are likely inhabited (bottom-right). Together, the four examples attest to a rather nuanced representation of population patterns across different settlement structures, especially considering that the estimates are derived only from Sentinel imagery at 10$\,$m GSD.

\begin{table}[htbp]
  \centering
\resizebox{0.45\textwidth}{!}{%
  \begin{tabular}{cc}
    \includegraphics[width=0.45\columnwidth]{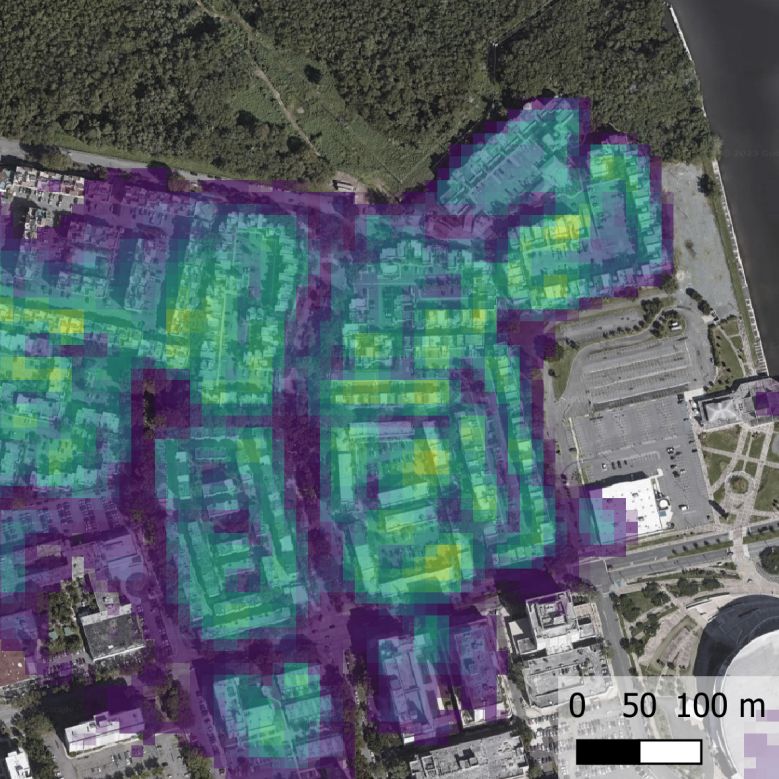} & 
    \includegraphics[width=0.45\columnwidth]{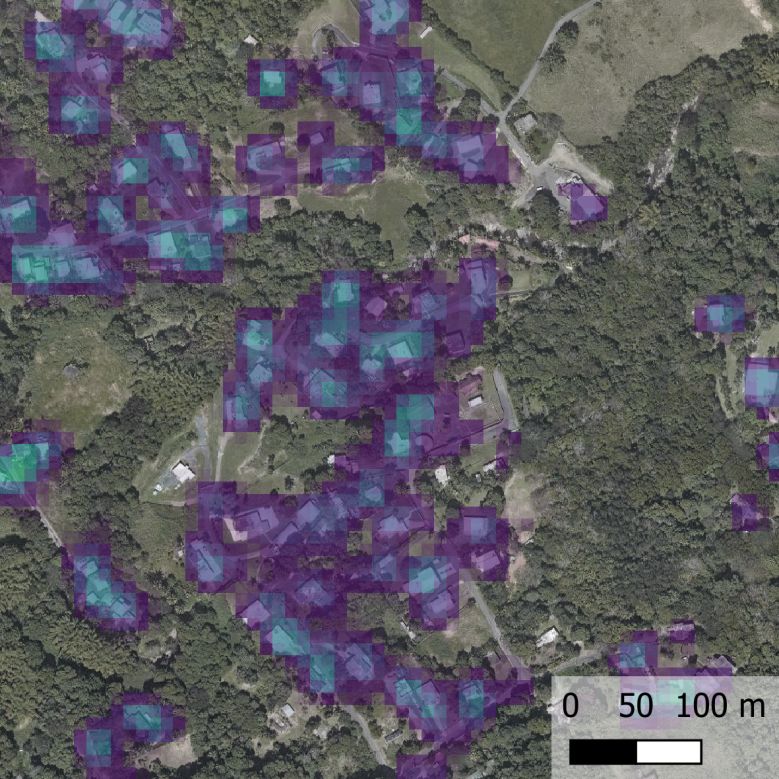} \\
    \includegraphics[width=0.45\columnwidth]{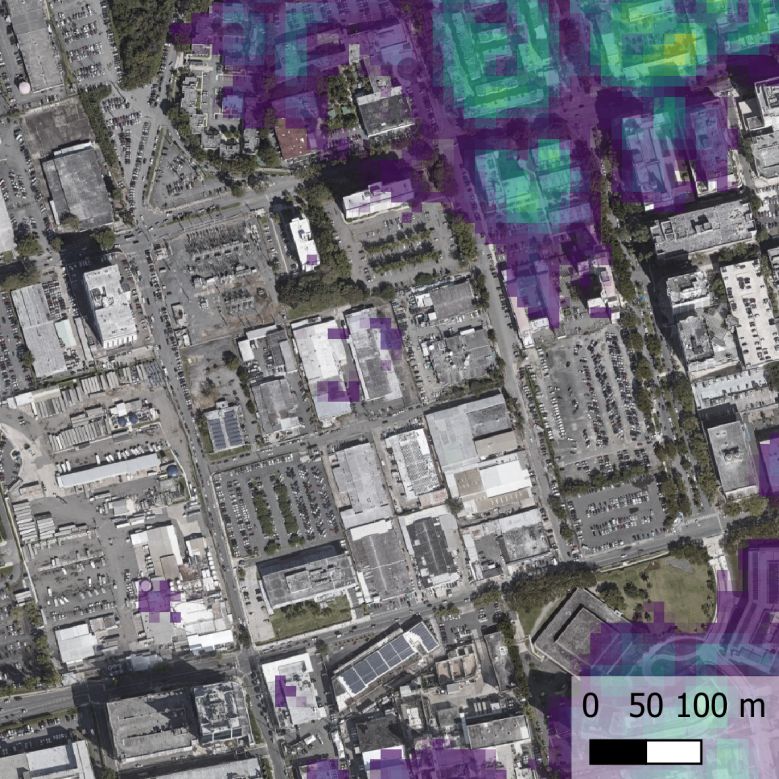} & 
    \includegraphics[width=0.45\columnwidth]{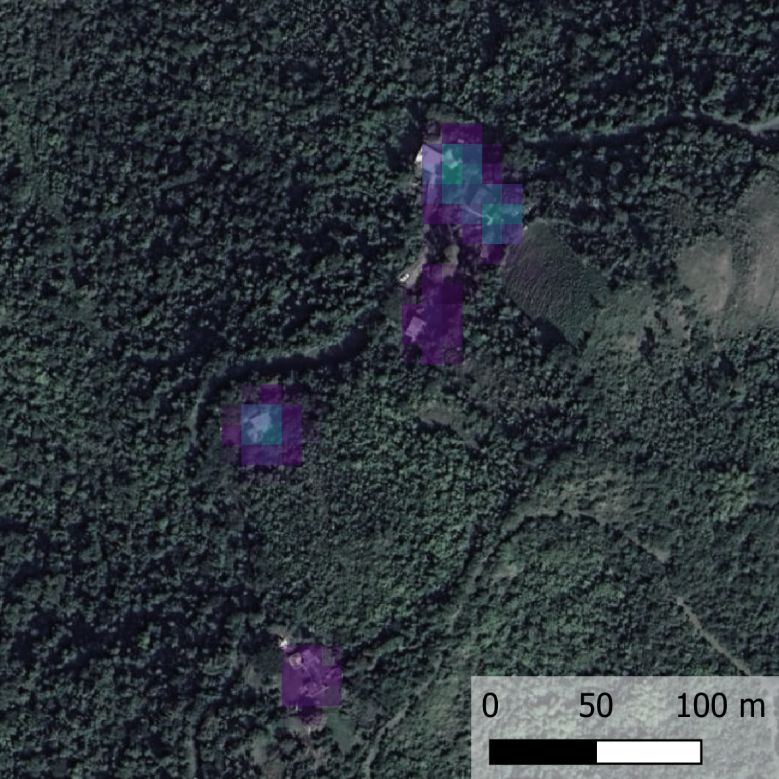} \\
  \end{tabular}
  }
  \captionof{figure}{Population density maps for four distinct locations in Puerto Rico, overlaid on desaturated VHR images of~\cite{googlemaps2023} (used for visualization only, not involved in map generation). Not only vegetation but also industrial warehouses are recognized as uninhabited, whereas isolated buildings in the forest are picked up.}
  \label{fig:pri_ex}
\end{table}

\subsection{Scalability}

To investigate the behavior of our model with increasingly coarse census information, we create six synthetic datasets from Puerto Rico's low-resolution census regions. To that end, we progressively merge the smallest pairs of adjacent regions to simulate census tables with 512, 156, 128, 64, 32, and 16 regions, for the same geographic area and population size.
We retrain the Bag-of-\textsc{Popcorn} model for each of those datasets and compare its performance to the original Puerto Rico model derived from the full 945 census regions. The weight decay regularization for the different (simulated) census scales is set as described in Section~\ref{ssec:implementation}.

    \begin{figure}[tp]
    \centering
    \begin{subfigure}[b]{0.33\textwidth}
        \centering
        \includegraphics[width=\textwidth]{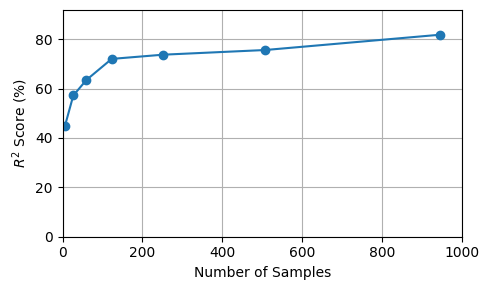}
        \caption{$R^2$}
        \label{fig:r2_performance}
    \end{subfigure} 
    \begin{subfigure}[b]{0.33\textwidth}
        \centering
        \includegraphics[width=\textwidth]{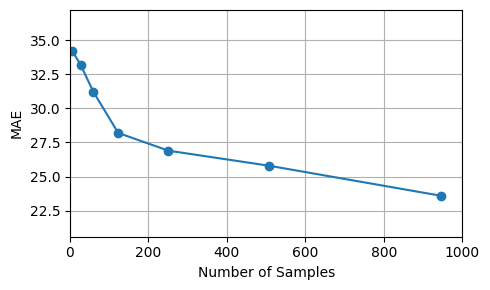}
        \caption{MAE}
        \label{fig:mae_performance}
    \end{subfigure} 
    \begin{subfigure}[b]{0.33\textwidth}
        \centering
        \includegraphics[width=\textwidth]{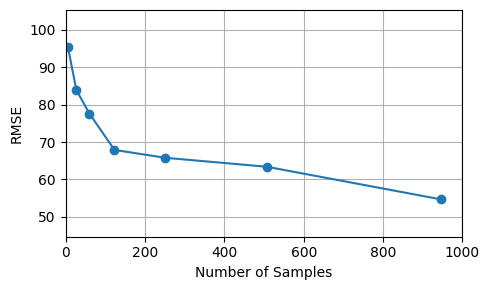}
        \caption{RMSE}
        \label{fig:rmse_performance}
    \end{subfigure}
    \caption{
    {Bag-of-\textsc{Popcorn} performance with progressively coarser and fewer census counts for training, evaluated on Puerto Rico.}}
    \label{fig:scalability_overall}
    \end{figure}

    {
     The resulting performance curves are displayed in Figure~\ref{fig:scalability_overall}. They
    show that the model maintains good performance down to a 8$\times$ coarser granularity of the regions, respectively a
    sample size of only 128. Especially the R\textsuperscript{2} score (regression fitness) stays high. We hypothesize that the robustness of the pre-trained building detector is the main reason why the subsequent disaggregation remains effective under in such challenging low-data conditions: even with as few as 16 samples, \textsc{Popcorn} still reaches an $R^2$ score of 45\%.
    }
    
\subsection{Ablation Studies} \label{ssec:abl}

\subsubsection*{Model Architecture}
We conducted an ablation to illustrate the influence of the separately trained building extractor, and of using it as initialization also for the occupancy branch. The experiment is illustrated in Figure~\ref{tab:ablation}, and with the results subsequently below. The analysis confirms that using separate modules for the built-up score and the occupancy rate, rather than a monolithic population estimator, offers noticeable advantages in the low data regime. The difference is most pronounced in Rwanda, which features the smallest number of census regions and the largest upscaling factor. In contrast, for the larger datasets from Switzerland and Puerto Rico, the performance of both variants is comparable.

Moreover, the study highlights the importance of initializing the occupancy branch with the weights of the pre-trained building extractor, benefiting from the larger underlying training dataset. Again the difference is largest in Rwanda, whereas in Switzerland and Puerto Rico, the number of census samples is sufficient to learn a passable occupancy predictor from scratch. Note that starting from pre-trained weights not only enables the model to capitalize on features learned from a larger training set but also promotes coherence between the built-up area and occupancy branches, which may simplify learning based on small gradient steps.

\begin{table*}[tp]
    \centering
    \begin{minipage}{\textwidth}
        \centering
            \centering
            \caption{Ablations of the model architecture. Cases A and C employ the dual-branch structure with separate branches for the builtUp score and the occupancy rate, c.f.~Figure~\ref{fig:graph_abs}. Cases A and B initialize the trainable weights with those of the pretrained building detector.}
            \includegraphics[width=0.85\textwidth]{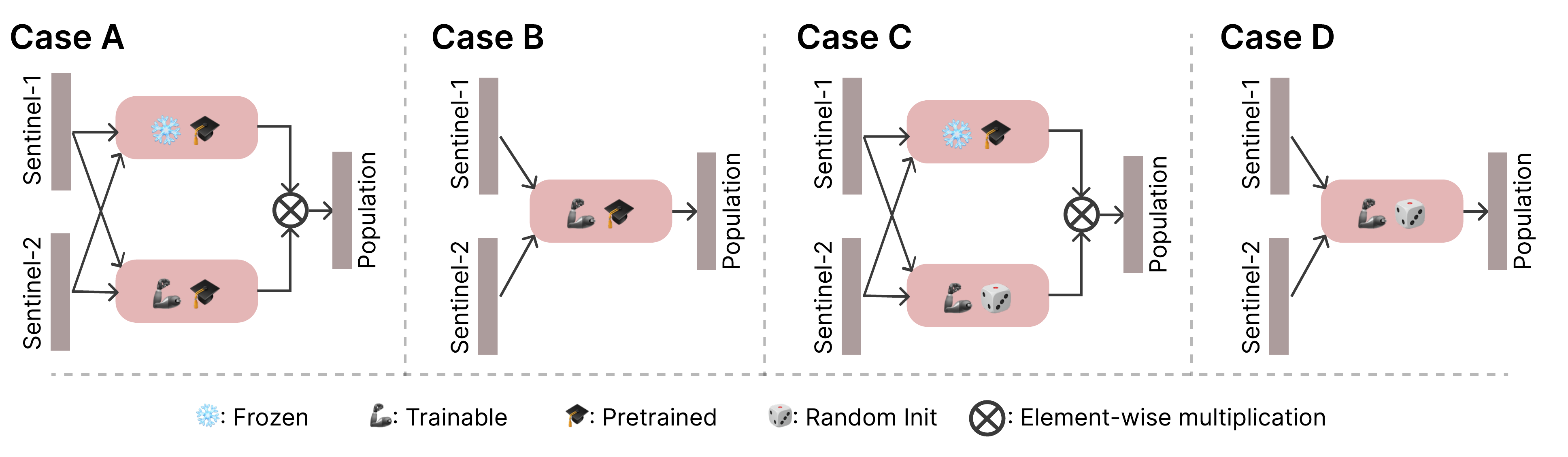}
        \vspace{35pt}
            \resizebox{\textwidth}{!}{
            \begin{tabular}{lcc|ccc|ccc|ccc}
                \toprule
                && & \multicolumn{3}{c|}{Switzerland} & \multicolumn{3}{c|}{Rwanda} & \multicolumn{3}{c}{Puerto Rico}\\
                &Weight Init. & OccRate & $R^2$ [\%] & MAE  & RMSE & $R^2$ [\%] & MAE &RMSE & $R^2$ [\%] & MAE &RMSE \\ \midrule
                Case A        &\cmark & \cmark&  \textbf{59.9}&  1.35&    \textbf{8.4}& \textbf{66 }& \textbf{10.1}& \textbf{20.2}& 81.8 & 23.6 & 54.8\\
                Case B        &\cmark & &  59.2&  \textbf{1.34}&   8.5& 54& 11.6& 23.7& \textbf{82.0}& \textbf{23.3}& \textbf{54.4}\\
                Case C        & & \cmark&  55.0 &  1.47 &  8.9& 46 & 11.3& 25.6& 77.6 & 26.6 & 60.7\\
                Case D        & &  & 52.7& 1.48&  9.1& 48& 12.5& 25.2& 74.6& 28.9& 64.6\\ \bottomrule
            \end{tabular}
            }
            \label{tab:ablation}
    \end{minipage}
    
\end{table*}

\subsubsection*{Bagging}

We evaluated the impact of model ensembling, transitioning from a singular \textsc{Popcorn} model to a Bag-of-\textsc{Popcorn} configuration, using the Kigali dataset, as detailed in Table~\ref{tab:ensembling_abl}.
Five independent instances of the \textsc{Popcorn} model were trained, moreover, we have four seasonal composites of Sentinel-1 and Sentinel-1 images for every location.
The base case is a prediction with a single model and a single seasonal composite (including Sentinel-1 and Sentinel-2). We repeat that experiment with each of the five model instances and each of the four composites and report average performance metrics.
In a second scenario, a single model is applied to each of the four seasonal composites. The predictions are averaged, corresponding to an ensemble with a single model instance, but four different inputs ("test-time augmentation"). Again we repeat the experiment with all five model instances and report average metrics. Note the implicit assumption that population changes between seasons of the same year can be neglected.
The third scenario is the standard ensemble setting where the same input data is fed to five different model instances. Again we run the experiment for each of the four seasons and report average metrics.
Finally, we regard each of the 20 possible combinations (5 instances $\times$ 4 seasons) as a member of one large ensemble and average their results.

Our findings indicate that each ensembling method markedly enhances the model's predictive capability. Test-time augmentation using different seasonal composites contributed more significantly to performance improvement than the random ensemble of model instances. However, the most robust performance was observed in the ``Large Bag" scenario, which integrated both multiple models and seasonal inputs.

\begin{table*}[tp]
    \centering
    \caption{Effect of ensembling evaluated on the Rwanda Dataset, evaluated on Kigali.}
    \resizebox{0.8\textwidth}{!}{%
    \begin{tabular}{ll|ccc|ccc}
        \toprule
          &&   \multicolumn{3}{c|}{Ensemble over} & \multicolumn{3}{c}{}  \\ 
         &&  members & seasons  & \#estimates & $R^2$ [\%]  & MAE   & RMSE \\ \midrule
  \textsc{Popcorn}                         & & &                   & 1  & 55  & 11.2  & 23.2 \\ \midrule
  \multirow{3}{*}{Bag-of-\textsc{Popcorn}} & Small &  & \cmark & 4  & 65  & 10.3  & 20.7 \\
 & Medium & \cmark & & 5  & 60  & 10.9  & 21.9 \\ 
 & Large & \cmark & \cmark & 20 & \textbf{66}  & \textbf{10.1}  & \textbf{20.2} \\ \bottomrule
    \end{tabular}
    }
    \label{tab:ensembling_abl}
\end{table*}

\subsubsection*{Input Modalities}

Finally, to assess the contributions of the two modalities -- Sentinel-1 SAR amplitudes in two polarization, respectively Sentinel-2 RGB-NIR spectra -- we train versions of the model that utilize only one modality.
From the dual-stream architecture depicted in Figure~\ref{fig:DDA}, we selectively remove the backbone corresponding to either Sentinel-1 or Sentinel-2 and pass only the features from the other, active backbone to the prediction heads for built-up score and occupancy.
That procedure disentangles the relative contributions of the two modalities and makes it possible to quantitatively assess how much each of them contributes to the overall model output.

Results are displayed in Table~\ref{tab:modality_ablation}. The experiment demonstrates that the two input data sources contribute complementary information, and leveraging both together gives the best results, across all three test sites.
On the contrary, there is no clear trend that one or the other is a more important data source for population mapping. In Switzerland, Sentinel-1 alone gives better results than Sentinel-2 alone, albeit by a moderate margin.
In Rwanda and Puerto Rico, Sentinel-2 alone nearly matches the performance of the combined model, whereas Sentinel-1 alone leads to a considerable performance drop.

\begin{table*}[tp]
    \centering
    \caption{Contribution of Sentinel-1 SAR and Sentinel-2 optical inputs.}
    \resizebox{0.8\textwidth}{!}{% 
    \begin{tabular}{cc|lll|lll|lll}
        \toprule
       && \multicolumn{3}{c|}{Switzerland}& \multicolumn{3}{c|}{Rwanda}& \multicolumn{3}{c}{Puerto Rico}\\
       S1 & S2 & \( R^2 \) [\%] & MAE  & RMSE & \( R^2 \) [\%] & MAE  & RMSE & \( R^2\) [\%] & MAE  & RMSE  \\ \midrule
       \cmark & \cmark & \textbf{59.9} & \textbf{1.35} & \textbf{8.4} & \textbf{66} & \textbf{10.1} & \textbf{20.2}& \textbf{81.8} & \textbf{23.6} & \textbf{54.8}\\
       \cmark & \xmark             & 54.9 & 1.50 & 8.9& 48 & 12.3 & 25.1& 69.0 & 35.7 & 71.4\\ 
       \xmark & \cmark             & 51.7 & 1.52 & 9.2& 66 & 10.5 & 20.4& 80.9 & 24.1 & 56.1\\ \bottomrule
        \end{tabular} 
    }
    \label{tab:modality_ablation}
\end{table*}

\section{Conclusion} \label{sec:con}

We have introduced Bag-of-\textsc{Popcorn}, a neural network model capable of estimating gridded population counts at 1$\,$ha resolution from Sentinel-1 and Sentinel-2 satellite imagery.
The proposed model has been shown to work well in a range of geographic regions, and greatly simplifies fine-grained population mapping: it requires neither a ground-based micro-census that is difficult to conduct at scale in large parts of the world; nor high-resolution geodata products that are expensive and/or not guaranteed to be updated. The proposed model can be trained solely based on coarse census data and requires only a small amount of training data to yield quite convincing results (e.g., 381 regional counts in our experiments for Rwanda).

We have experimentally demonstrated that the proposed Bag-of-\textsc{Popcorn} model compares favorably to existing population mapping tools. In particular, it excels in locations that lack high-resolution building information. However, we found that even in locations for which high-quality building polygons are available the model often outperforms methods that use those polygons.

The \textsc{Popcorn} framework separates population mapping into two separate streams that estimate built-up areas and building occupancy rates. This strategy makes it possible to exploit training data for built-up area mapping, which is much more plentiful than population counts. By separately training a built-up area detector on publicly available data, which also can serve as initial values for building occupancy estimation, we can compensate for the coarseness of census data in many parts of the world.

\section{Outlook} \label{sec:out}

The Bag-of-\textsc{Popcorn} model can retrieve high-resolution population maps only from free satellite images, and optionally coarse census counts for dasymetric rescaling. Still, several limitations remain that should be addressed in future work. A main bottleneck is scalability. Although we used comparatively small countries for our study, we had to limit model complexity to process them on a GPU with 24 GB of onboard memory. To enable population mapping for larger countries or even entire continents it will be necessary to develop leaner, more memory-efficient variants of our model, or to port it to a high-performance computing system (or a combination of both). Moreover, the lack of reliable and up-to-date census data (even at coarse granularity) in some parts of the world still constitutes a major challenge.
 
Larger domain shifts between training and testing sites remain an unsolved problem. 
If a region exhibits unique image patterns and no census data is available at all, then our methodology is not applicable.
This does not imply that one always needs census data for the target region. E.g., one may be able to train on a not too distant country or region with similar characteristics, leaving out the dasymetric adjustment. A useful edge case of our method is the situation where only a single, country-wide population estimate is available (which ich almost always the case). It turns out that disaggregating such a global population number with satellite-derived built-up scores achieves quite competitive results, as shown in our experiments. For cases with no suitable training regions at all, we speculate that recent remote sensing foundation models~\citep{prexl2023multi,fibaek2024phileo} could offer a way to bridge the domain gap.
Furthermore, we point out that the proposed method -- like any methodology based on building footprints or built-up area detection -- will be misled by abandoned buildings and settlements. Lastly, a significant part of the world population lives in the tropical climate zone, where persistent cloud cover may render optical imaging impossible for months. While we could show that \textsc{Popcorn} is to some degree able to compensate for missing optical imagery with SAR imaging~(Table~\ref{tab:modality_ablation}), there is still room for improvement. One promising direction is to design integrated models that directly work with raw image time series instead of composites and internally handle cloud masking and data fusion with SAR in an optimal, task-specific fashion.

% Technical development
From an engineering point of view, it may be interesting to explore the link to other down-scaling (a.k.a\ "guided super-resolution")  techniques, for instance, guided filtering~\citep{he2012guided} or image diffusion~\citep{metzger2023guided}. Moreover, we speculate that techniques for regression in unbalanced datasets~\citep{yang2021delving} could be useful to more precisely estimate very high population densities. Moreover, one should still consider using the high-resolution building datasets during the training process to better be able to generalize to unseen regions.

% Nightlights, future works
In future works, we aim to explore the integration of nightlight data into our model. This study has successfully employed a shared feature representation for Sentinel-1 and Sentinel-2 at a 10m GSD, but extending the pretraining approach to include the approximately 500m resolution of nightlights presents a formidable challenge. It is commonly known that nightlights are a reliable indicator of economic activity~\cite{martinez2018much}. However, it has also been found that nightlights are not always a reliable estimator for population density, primarily due to the underrepresentation of slums and rural regions~\cite{gibson2021night}, which poses a further challenge that needs to be addressed.

% Real-world Applications
Finally, we believe that timely, spatially explicit, high-resolution population maps have a lot of untapped application potential. We hope that easy-to-use models based on reliable and widely accessible data sources, in the spirit of our Bag-of-\textsc{Popcorn}, will make population mapping accessible to a wider community of interested users and service providers in sectors like urban planning, disaster relief, and public health.

\section{Declaration of Competing Interest}

The authors declare no competing interests.

\section{Author contributions}

KS, DT, RCD, and NM together developed the concept and designed the methodology. NM curated the data, implemented the methodology, and conducted the experiments and analysis. KS, DT, and RCD supervised the study. All authors contributed to writing the manuscript, based on an initial draft by NM.

\section{Acknowledgement}

We thank Luca Dominiak, who helped us with the \textsc{Pomelo} experiments on the Swiss dataset. We further thank Arno Rüegg and Leonard Haas for conducting prestudies related to this work.

\newpage 

\bibliographystyle{elsarticle-harv} 
\bibliography{bibliography}

%%%%%%% APPENDIX %%%%%%%%%%%%%%%%%%

\newpage
\section*{Appendix}

\appendix
\addcontentsline{toc}{section}{Appendices}
\renewcommand{\thesubsection}{\Alph{subsection}}

\subsection{Definition of Seasons} \label{app:season}

For our work, we define the temporal windows for the seasonal composites as listed in Table~\ref{tab:seasons}.

\begin{table}[htb]
    \centering
    \begin{tabular}{l|c|c}
         &  Start &  End\\ \midrule
         Spring & 01-03-2020 & 31-05-2020\\ 
         Summer & 01-06-2020 & 31-08-2020\\
         Autumn & 01-09-2020 & 30-11-2020\\
         Winter & 01-12-2020 & 28-02-2021\\ \bottomrule
    
    \end{tabular}
    \caption{Start and end dates of seasons for image compositing.}
    \label{tab:seasons}
\end{table}

\subsection{Building Extraction Performance} \label{app:builtup}

The building extractor is trained with the same protocol and the same building labels as in~\cite{hafner2022unsupervised}. The labeled portion of the training set contains $\approx$10$^9$ pixels covering 132$\,$000$\,$km$^2$, of which 11\% are labeled as built-up. The additional, unlabeled training images cover $\approx1.8\cdot10^9$ pixels, or  178$\,$000$\,$km$^2$. The validation set from the source domain comprises $2\cdot10^8$ pixels (22$\,$000$\,$km$^2$), while the target domain test set covers 10$^7$ pixels (1$\,$100$\,$km$^2$).

Table~\ref{tab:builtup_ablation} presents a comparative analysis of the built-up area detection results with latent representations of varying size (i.e., channel depth per U-net layer). The \emph{original} configuration is the one recommended in ~\cite{hafner2022unsupervised}, with 64 channels at half-resolution and 128 channels at quarter-resolution. The \emph{slim} configuration is our down-sized variant with 8, respectively channels.
The higher capacity of the original model yields substantially better built-up area detection in the ``Source Domain", where the model's training labels are located, i.e., North America and Australia. But that advantage does not persist in the ``Target Domain" covered only by a consistency loss between Sentinel-1 and Sentinel-2 prediction on unlabeled data, i.e., on all other continents. There, our slim version of the model performs as well. It appears that the extra capacity allows for a more specific (over-)fit to the local patterns, but the domain adaptation scheme is not able to carry over that advantage to unseen places with different settlement structures.

\begin{table}[htb]
    \centering
    \caption{Performance comparison of original and slim model configurations in built-up area detection. Metrics include F1 score and Intersection over Union (IoU) for both source and target domains.}
    \label{tab:builtup_ablation}
    \resizebox{0.475\textwidth}{!}{%
    \begin{tabular}{l|c|cc|cc}
        \multirow{2}{*}{} &  $\#$params & \multicolumn{2}{c|}{Source Domain} & \multicolumn{2}{c}{Target Domain} \\
                    &  &F1 [\%] & IoU [\%] & F1 [\%] & IoU [\%] \\
        \hline
        Original & 1.8M &\textbf{79} & \textbf{65} & 64 & \textbf{48} \\
        Slim & 30k &72 & 57 & \textbf{65} & \textbf{48} \\ \bottomrule
    \end{tabular}
    }
\end{table}

\subsection{Effect of Dasymetric Scaling}

    We compare the raw population numbers estimated by the model to those after dasymetric rescaling  Table~\ref{tab:diaggregation}. As expected, scaling to the census totals improves the mapping accuracy if reliable census information is available, whereas it can hurt the mapping accuracy when done with inaccurate, extrapolated census counts. 

    \begin{table*}[ht]
        \centering
        \begin{tabular}{l|ccc|ccc|ccc}
        \toprule
        
        & \multicolumn{3}{c|}{Switzerland} & \multicolumn{3}{c|}{Rwanda} & \multicolumn{3}{c}{Puerto Rico}\\ 
        & $R^2$ [\%] & MAE & RMSE & $R^2$ [\%] & MAE & RMSE  & $R^2$ [\%] & MAE & RMSE \\
        \midrule
        Raw             & 55.2 &  1.37 & 8.9 & \textbf{66} & \textbf{10.1} & \textbf{20.2} &  76.5 & 25.3 & 62.25\\ 
        Disaggregated   & \textbf{59.9} &  \textbf{1.35} & \textbf{8.4} & 55* & 10.7* & 23.2* &  \textbf{81.8} & \textbf{23.6} & \textbf{54.8} \\
        \bottomrule
        \end{tabular}
        \caption{Effects of dasymetric rescaling. * For Rwanda, the 2020 counts used to scale the raw model estimates were projections based on the 2012 census.}
        \label{tab:diaggregation}
    \end{table*}

\subsection{Distribution of Scale Factors}

When applying dasymetric rescaling, we obtain an individual scale factor per census region that indicates the (multiplicative) deviation between the model estimates (accumulated per region) and the corresponding census counts. We examine the distribution of those scale factors across all coarse census regions (i.e., at the granularity used for training) in Table~\ref{tab:scales_summary}. Besides raw histograms, we also show summary statistics, as well as the census counts and model estimates for the total national population.

    \begin{table*}[ht]
        \centering
        \caption{Distribution of dasymetric scale factors per country.}
        \label{tab:scales_summary}
        \begin{tabular}{lccc}
            \toprule
            & \textbf{Puerto Rico} & \textbf{Rwanda} & \textbf{Switzerland} \\
            \midrule
             & \includegraphics[width=0.24\linewidth]{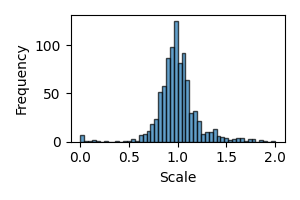} & \includegraphics[width=0.24\linewidth]{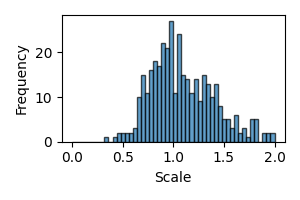} & \includegraphics[width=0.24\linewidth]{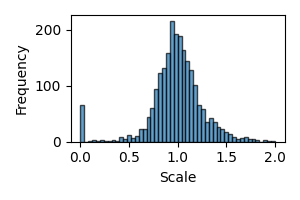} \\
            \midrule 
            \textbf{10$^\text{th}$ Percentile} & 0.81& 0.73& 0.73\\
            \textbf{Median} & 0.98 & 1.05& 0.99\\
            \textbf{90$^\text{th}$ Percentile} & 1.29 & 1.61& 1.32\\
            \textbf{Estimation (national)} & 3$\,$271$\,$710 & 11$\,$465$\,$801 & 8$\,$071$\,$153 \\
            \textbf{Coarse GT (national)} & 3$\,$285$\,$874 & 12$\,$355$\,$930 & 8$\,$740$\,$066 \\
            \bottomrule
        \end{tabular}
    \end{table*}

    In all three cases, the scale factors are centered around 1, meaning that the mean/median bias of the \textsc{Popcorn} model is low.
    For Switzerland and Puerto Rico, the histograms also have a single, clear peak, while for Rwanda the scales are more broadly scattered over a range from about 0.5 to 1.5. 
    We point out that the coarse census counts were used during training, hence the fairly good agreement only shows that our model was able to correctly fit them. The histograms provide no indication of how well the population distribution \emph{within} a region was predicted, for this please refer to the experiments described in Section~\ref{sec:res}.

\subsection{Generalization} \label{ssec:trans}

We have evaluated how well the Bag-of-\textsc{Popcorn} model generalizes by training it on data from Uganda and then applying it to Kigali, Rwanda. We present the resulting performance in Table~\ref{tab:uganda_vs_others}. As expected there is a noticeable drop in performance, with an $R^2$ score of 44\% compared to 66\% for the model trained for Rwanda. Still, even the Bag-of-\textsc{Popcorn} model trained on a different, neighboring country surpasses all medium-resolution baselines that are specialized to Rwanda, c.f.\ Table~\ref{tab:rwa}. In absolute terms, there is only a moderate increase in mapping error from 10.1 to 11.6 people/ha in MAE.
\begin{table}[htp]
    \centering
    \caption{Transferring a Bag-of-\textsc{Popcorn} model to a new geographic location. Two model instances trained on Rwanda, respectively Uganda, are evaluated on the same target region, Kigali (Rwanda).}
    \label{tab:uganda_vs_others}
    \begin{tabular}{cccc}
        \toprule
        \multicolumn{1}{c}{Training set} & $R^2$ [\%] & MAE  & RMSE \\ 
        \midrule
        Rwanda  & 66 & 10.1 & 20.2 \\
        Uganda & 44 & 11.6 & 26.1 \\
        \bottomrule
    \end{tabular}
\end{table}

\begin{figure*} [tbp]
    \centering
    \includegraphics[width=\textwidth]{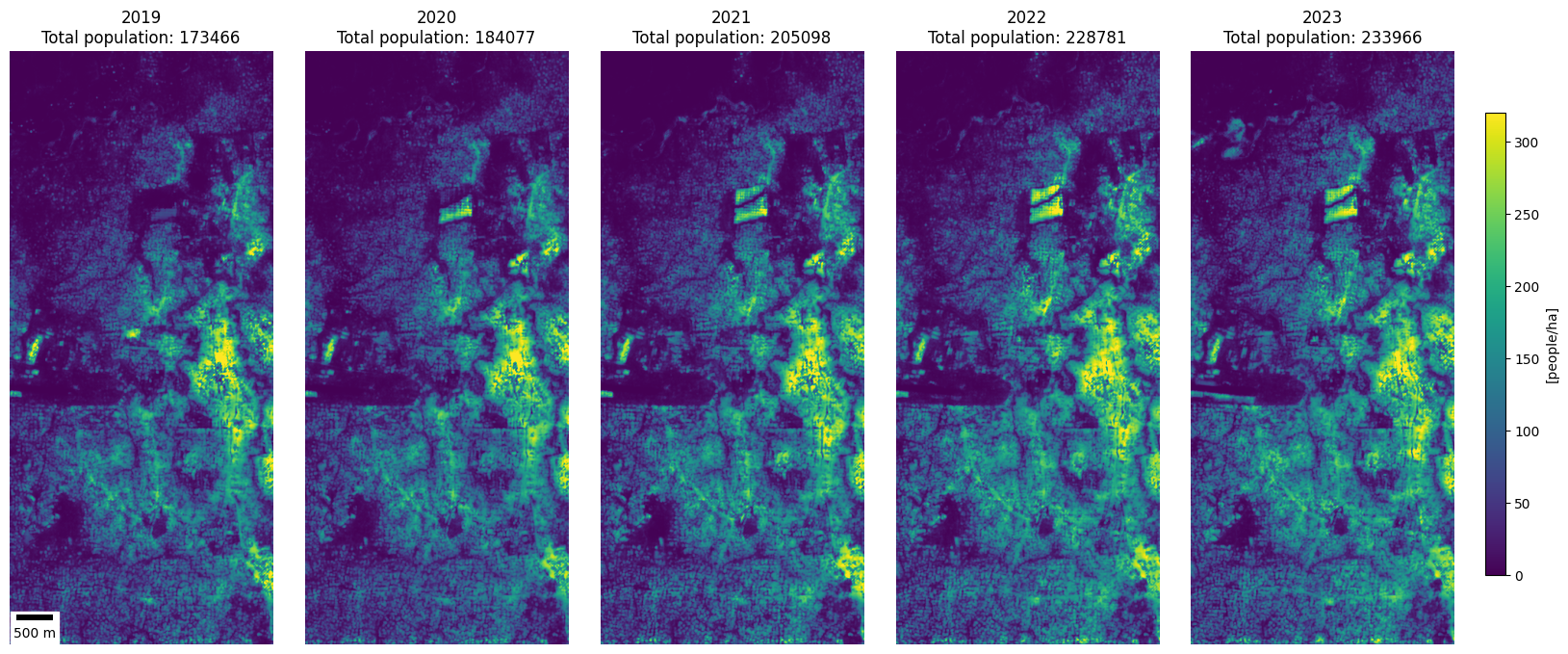}
    \caption{Estimated population time-series for Bunia, DRC, generated by independent, repeated mapping with the Bag-of-\textsc{Popcorn} model. Notably, it captures the progressive formation of the Kigonze refugee camp north of the city center, but also the gradual densification of the urban fringe west of the center.}
    \label{fig:time_series}
\end{figure*}

To showcase the potential of our model for monitoring population dynamics, we applied the instance trained in Rwanda to a time series of images showing Bunia (in neighboring DRC), one of the fastest-growing African cities with estimated 7\% annual growth~\citep{africancities2021}.

Figure~\ref{fig:time_series} chronicles the estimated evolution of the city's population from 2019 to 2023.
Conspicuously,
the model seems to track the locally concentrated growth of a refugee camp in the upper half of the depicted region, and diffuse densification, especially in the southwestern periphery.

We hope these quantitative results provide an incentive for future researchers to build on top of our system and if applicable quantitatively evaluate them.

% Time series 
If future research of can validate those findings quantitatively, image-based tracking tracking of population dynamics could be a viable alternative to conventional census projection models. Such a method would not only bridge the periods between census rounds more accurately than generic, exponential growth models but also enhance our understanding of demographic trends, including population growth and migration patterns.

Moreover, the short revisit times of contemporary Earth observation satellites would raise the possibility of continuously tracking population numbers in near-real-time. This could have transformative implications, enabling real-time monitoring and response to population changes. However, further research is required to ensure consistent and comparable estimates over short time scales, despite inevitable fluctuations in observed reflectance and backscatter values.

\end{document}